\documentclass[
    numbers=noenddot,
    parskip=half-,
    fontsize=12pt,
    paper=a4,
    oneside,
    titlepage,
    bibliography=totoc,
    chapterprefix=false,
]{scrbook}

\usepackage{graphicx}
\usepackage[onehalfspacing]{setspace}
\usepackage{wrapfig}
\usepackage{subcaption}
\usepackage{float}

\usepackage{amsmath}
\usepackage{amsfonts}
\usepackage{mathtools}
\usepackage{afterpage}

\usepackage[ruled,linesnumbered]{algorithm2e}

\usepackage{tocbasic}
\usepackage{booktabs}
\usepackage{multicol}
\usepackage{multirow}

\usepackage[skip=5pt,font=footnotesize,margin=2cm]{caption}

\usepackage[]{scrlayer-scrpage}

\usepackage{hyperref}
\hypersetup{%
  colorlinks,
  citecolor=[rgb]{0,0.5,0},
  linkcolor=blue,
  urlcolor=cyan
}

\usepackage[citestyle=alphabetic, bibstyle=alphabetic, sorting=nyt, backend=biber, language=english, backref=true, maxcitenames=2]{biblatex}

\usepackage[autostyle,english=american,german=quotes]{csquotes}
\usepackage[titletoc]{appendix}
\usepackage{listings}
\usepackage[frozencache=true,cachedir=minted-cache,newfloat]{minted}
\usepackage{caption}
\usepackage{xcolor}

\newenvironment{code}{\captionsetup{type=listing}}{}
\SetupFloatingEnvironment{listing}{name=Code}
\definecolor{code_bg}{HTML}{F2F2F2}
\removefromtoclist[float]{lol}

\definecolor{commentcolor}{rgb}{0.5,0.5,0.5}
\definecolor{keywordcolor}{rgb}{0,0,1}
\definecolor{stringcolor}{rgb}{0.58,0,0.82}

\graphicspath{{images/}}

\title{Investigating Sparsity in \\Recurrent Neural Networks}

\author{Harshil Jagadishbhai Darji}

\date{July 1, 2021}

\newcommand{\thesisType}{Masterarbeit}

\makeatletter
\let\thetitle\@title
\let\theauthor\@author
\let\thedate\@date
\makeatother

\begin{document}

\frontmatter
\begin{titlepage}
    \centering
    \begin{onehalfspace}
    	
        	\includegraphics[width=7cm]{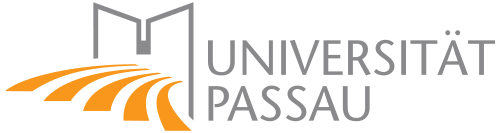}\\
        	\vspace{1.0cm}
        	\large {\bfseries Lehrstuhl f\"ur Data Science }\\

        	\vspace{2.5cm}

            \begin{doublespace}
            	{\textsf{\Huge{\thetitle}}}
            \end{doublespace}

        	\vspace{2cm}

            \Large{Masterarbeit von}\\

        	\vspace{1cm}

        	{\bfseries \large{\theauthor}}

        	\vfill

        	{\large
        		\begin{tabular}[l]{cc}
        			\textsc{1.~Pr\"ufer} & \textsc{2.~Pr\"ufer} \\
        			Prof.~Dr.~Michael Granitzer& Prof.~Dr.~Harald Kosch
        		\end{tabular}
        	}

        	\vspace{1.5cm}

        	\parbox{\linewidth}{\hrule\strut}

            \vfill

	    \thedate
    \end{onehalfspace}
\end{titlepage}

{
  \hypersetup{linkcolor=black}
  \tableofcontents
}
\newpage

\chapter*{Abstract}
In the past few years, neural networks have evolved from simple Feedforward Neural Networks to more complex neural networks, such as Convolutional Neural Networks (CNNs) and Recurrent Neural Networks (RNNs). Where CNNs are a perfect fit for tasks where the sequence is not important such as image recognition, RNNs are useful when order is important such as machine translation. An increasing number of layers in a neural network is one way to improve its performance, but it also increases its complexity making it much more time and power-consuming to train. One way to tackle this problem is to introduce sparsity in the architecture of the neural network. Pruning is one of the many methods to make a neural network architecture sparse by clipping out weights below a certain threshold while keeping the performance near to the original. Another way is to generate arbitrary structures using random graphs and embed them between an input and output layer of an Artificial Neural Network (ANN). Many researchers in past years have focused on pruning mainly CNNs, while hardly any research is done for the same in RNNs. The same also holds in creating sparse architectures for RNNs by generating and embedding arbitrary structures. Therefore, this thesis focuses on investigating the effects of the before-mentioned two techniques on the performance of RNNs. We first describe the pruning of RNNs, its impact on the performance of RNNs, and the number of training epochs required to regain accuracy after the pruning is performed. Next, we continue with the creation and training of Sparse Recurrent Neural Networks (Sparse-RNNs) and identify the relation between the performance and the graph properties of its underlying arbitrary structure. We perform these experiments on RNN with Tanh nonlinearity (RNN-Tanh), RNN with ReLU nonlinearity (RNN-ReLU), GRU, and LSTM. Finally, we analyze and discuss the results achieved from both the experiments.
\newpage

\chapter*{Acknowledgments}
Many people helped and encouraged me during my Master's studies, and I would like to express my gratitude to all of them. However, there are some people whom I would like to thank individually as their vital contribution helped me get this far.

First, I would like to thank Julian Stier, my thesis advisor, for his constant guidance and encouragement to not only focus on my preliminary research questions but also to explore other possibilities to improve and expand my research interests. He was always quick in providing feedback on my work that allowed me to complete my work promptly.

Secondly, I would like to thank my thesis supervisors, Prof. Dr. Michael Granitzer and Prof. Dr. Harald Kosch. Prof. Granitzer's comments on my initial work helped me expand my thesis work. Their comments during the initial presentation of my work helped me think of original and efficient ways to improve my implementations.

During my studies in Passau, I became friends with people who not only made this city a second home, but they were always there whenever I needed them. One of the important things we all learned from this experience is that a friendship goes beyond any borders.

Finally, but most importantly, I would like to thank my parents and my sister for their continuous support and blessings. They always made sure I get the best education, no matter what the situation is. They have always believed in me, and their combined efforts and sacrifices made it possible for me to continue my Master's education in Germany.
\newpage

\thispagestyle{empty}
\cleardoublepage
{
  \hypersetup{linkcolor=black}
  \listoffigures
}
\newpage

\thispagestyle{empty}
\cleardoublepage
{
  \hypersetup{linkcolor=black}
  \listoftables
}
\newpage

\thispagestyle{empty}
\cleardoublepage
{
  \hypersetup{linkcolor=black}
  \listofalgorithms
}
\newpage

\mainmatter

\chapter{Introduction}\label{chap:introduction}

A neural network, also known as an Artificial Neural Network (ANN), is an algorithm that mimics the way a human brain operates. It took millions of years of evolution for the human brain to achieve the level of intelligence we observe today. This intelligence helps us quickly interpret and evaluate what we see in our surroundings. For example, consider the following image of hand-written digits from the MNIST dataset \cite{mnist}:

\begin{figure}[h]
	\centering
	\includegraphics[width=0.5\linewidth]{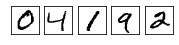}
	\caption[Hand-written digits]%
	{\textbf{Hand-written digits}: Each hand-written digit is a $28\times28$ pixel grayscale image. The entire MNIST dataset contains a total of $70000$ such images that split into a training set of $60000$ images and a test set of $10000$ images.}
	\label{fig:digits}
\end{figure}

One can quickly recognize these digits as $04192$, thanks to the network of billions of neurons available in our brain that makes it easy to identify visual patterns. Past experiences and memories stored in our brain make this process so simple that it happens subconsciously most of the time. This simple process becomes much more difficult when we try to write computer programs to identify similar patterns due to the lack of precise rules and hundreds of exceptions and varieties of a single pattern.

Neural networks tackle this problem by inferring rules from a large set of related training data. For example, to train a neural network to efficiently recognize hand-written digits shown in figure \ref{fig:digits}, a dataset large enough to include different styles of hand-written digits is needed.

Neural networks and their variations such as Convolutional Neural Networks (CNNs), Recurrent Neural Networks (RNNs) are proven to be useful in many different application areas such as image recognition \cite{image_recon}, machine translation \cite{russian}, recommender systems \cite{recommender}. Despite this state-of-the-art performance, neural networks are computationally complex and memory-intensive due to their deep structures.

Sparse Neural Networks are a viable option to make neural networks less complex and make them memory efficient. Such sparsity in neural networks can be induced by pruning weights, utilizing skip connections, or generating random architectures based on graphs. Sparsity in a neural network can reduce that network's size by many times with little to no drop in performance \cite{sparse_nn, dey, liu}. Therefore, in this thesis, we implement two different ways to achieve sparsity in Recurrent Neural Networks and analyze its performance impact.


\section{Motivation}\label{section:motivation}

As stated before, deep neural networks are likely to have an increased performance but at the cost of higher complexity and increased fast memory requirements. One way to mitigate this issue while maintaining the performance is to introduce sparsity into a network's connections \cite{deep_res}. For example, Mao et al. in \cite{mao} report a higher-compression ratio with coarse-grained pruning without loss of accuracy while saving about twice the memory references. Similarly, Sun et al. in \cite{sparse_face} reported that the sparse ConvNet model, which only has 12\% of the original parameters, still performs similarly to the baseline model.

These are just a few examples where Sparse Neural Networks have proven advantageous in time, energy, and memory savings. Sparsity in traditional neural networks and CNNs is studied widely by many researchers but is not explored much in Recurrent Neural Networks that are difficult to train due to their nonlinear iterative nature. Sparse structures in traditional neural networks have shown a promise of training potential \cite{sparse_nn} which, if applied to Recurrent Neural Networks, can also make training RNNs less difficult by reducing their network size while retaining their performance.


\newpage
\section{Research questions}\label{section:research_questions}

The primary research goal of this master thesis is to investigate the effects of sparse structure on the performance of the Recurrent Neural Networks and its variants, namely Long Short-Term Memory and Gated Recurrent Unit. To do so, we have defined the following set of correlated research questions:

\begin{enumerate}
	\item\label{rq:q1} What is the effect of weights pruning on a recurrent network's accuracy?
	    
	\item\label{rq:q2} What percentage of weights pruning is permissible without triggering a significant reduction in the performance?
	    
	\item\label{rq:q3} After pruning a certain percent of weights, if we see a significant reduction in the accuracy,  how many re-training epochs can regain accuracy?
	    
	\item\label{rq:q4} How does a randomly structured recurrent network's performance correlate with the graph properties of its internal structure?
	
	\item\label{rq:q5} Is it possible to predict a randomly structured recurrent network's performance using the graph properties of its base random graph?
\end{enumerate}

We will answer all these questions for four different recurrent networks, namely RNN with Tanh nonlinearity (RNN-Tanh), RNN with ReLU nonlinearity (RNN-ReLU), Long Short-Term Memory (LSTM), and Gated Recurrent Unit (GRU), by following the approach explained in section \ref{section:proposed_approach}.


\newpage
\section{Proposed approach}\label{section:proposed_approach}

To investigate the effects of sparse structures in recurrent networks, we propose the following three experiments:
\begin{enumerate}
	\item \textbf{Investigating the effects of weights pruning on recurrent networks}: \\
	    In this experiment, we prune input-to-hidden and hidden-to-hidden weights, both simultaneously and individually.
	    
	    We start by training a recurrent network on a dataset for a certain number of epochs. Once this training is complete, we follow the process shown in the following flowchart:
	    
	    \begin{figure}[h]
            \centering
            \includegraphics[width=0.5\linewidth]{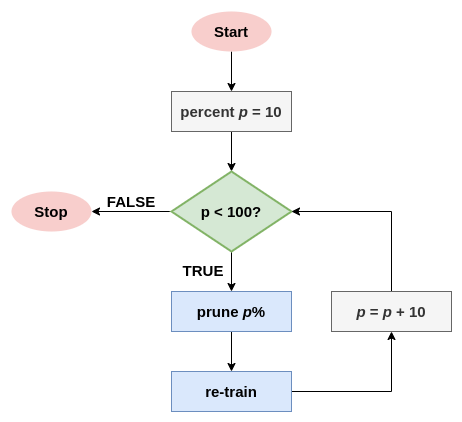}
            \caption[Flowchart for the pruning experiment]{The basic flowchart depicting the process of pruning and re-training a trained recurrent network.}
            \label{fig:flowchart_pruning}
        \end{figure}
        
    	 To outline this flowchart, once the initial training is complete, retrieve the learned weights of the trained model, prune the lower $p\%$ of weights (where $p \in \mathbb{Z}:p \in [1, 100]$), and re-train this pruned model to check for the number of epochs required to regain accuracy.
    	 
    	 This experiment will help answer research questions \ref{rq:q1}, \ref{rq:q2}, and \ref{rq:q3}.
    	
	\item \textbf{Analyzing the performance of recurrent networks with randomly structured recurrent networks}: \\
	    We start by generating randomly structured neural networks by following the technique described by Stier et al. in \cite{julian} that generate Sparse Neural Networks.

        After generating Sparse Neural Networks, we introduce recurrent connections to generate Sparse RNNs. These recurrent connections are necessary to make each subsequent run dependent on the previous run.

        Once we have Sparse RNNs, we train them on a dataset and compute the correlation between its accuracy and its internal structure's graph properties.
        
        This experiment will help answer the \ref{rq:q4}th research question.
	    
    \item \textbf{Performance prediction of a randomly structured neural network:}: \\
        Once we have trained Sparse RNNs, we train three different regressor algorithms with graph properties as features and corresponding accuracy as the target. Since this is a regression problem, we finally report an R-squared value for each RNN variant and each regressor algorithm.
        
        This experiment will help answer the \ref{rq:q5}th research question.
\end{enumerate}

All these three approaches are described in detail in the later chapters.

\chapter{Related works}\label{chap:literature}

A few researchers have worked on pruning to induce sparsity in Recurrent Neural Networks in the past couple of decades. Apart from that, to our knowledge, only one paper explains the process of generating Sparse Neural Networks from randomly generated graphs. This section will summarize these researches to have an idea about what has been done so far.

In September 1994, Lee et al. in \cite{lee_pruning} published one of the first pruning techniques in terms of Recurrent Neural Networks in which authors apply pruning to trained recurrent neural nets. They follow the train, prune, and retrain approach where they train an extensive network on grammar dataset, apply pruning on the trained network, and retrain on the same training set until either satisfactory performance is achieved or the network no longer converge. Since the authors use the incremental training approach, there is no need for the network to use the entire training set, meaning the network can achieve satisfactory achievement using only a part of the training set. After each pruning and retraining cycle, the network would require even less data than the initial network. Based on their experiments' results, the authors claim that ``our pruning/retraining algorithm is an effective tool for improving the generalization performance". According to the authors, this improved performance is due to the reduced size of the network.

In June 2017, Han et al. propose a Recurrent Self-Organising Neural Network (RSONN) in \cite{han_pruning}. To self-organize the RNNs' structure, they use Adaptive Growing and Pruning Algorithm (AGPA). This algorithm works by adding or pruning the hidden neurons based on their competitiveness during the training phase. Authors verify their approach on various benchmark datasets to compare against existing approaches. Based on the experiment results, the authors claim that the proposed RSONN effectively simplifies the network structure and performs better than some exiting methods. Authors compare their performance against various neural network models (such as Fuzzy Neural Network (FNN, \cite{han-why}), Exponential Smoothing Recurrent Neural Network (ESRNN, \cite{esrnn}), and Self-Organizing Radial Basis Function (SORBF, \cite{han-again-why})) and report better performance than others with fewer hidden neurons and less computational complexity. According to the authors, ``the proposed RSONN achieved better testing RMSE and running times than the other algorithms".

In April 2017, a group of researchers from Baidu research published \cite{sparse-rnn} in which they apply pruning to weights during the initial training of the network. At the end of the training, according to the authors, ``the parameters of the network are sparse while accuracy is still close to the original dense neural network". The authors claim ``the network size is reduced by $8\times$ and the time required to train the model remains constant". As explained in the paper, each recurrent layer has a constant number of hidden units, while we plan to have a different number of hidden units at each recurrent layer. All the experiments are also done on a private dataset, which prevents others from replicating the results, which is necessary to compare this technique of introducing sparsity with other approaches.

In November 2017, another group of researchers from Baidu research published \cite{narang-sparse} in which they propose a new pruning technique, block pruning. This technique can be used to zero out a block of weights during the training phase of a recurrent network, resulting in a Block-Sparse RNN. In addition to this, authors also use group lasso regularization to check if it can induce block sparsity. Authors report that ``block pruning and group lasso regularization with pruning are successful in creating block-sparse RNNs".  Their experiments report a nearly $10\times$ reduction in model size with a loss of 9\% to 17\% accuracy in vanilla RNN and GRU with block-sparsity of 4x4 blocks. Authors claim their approach is ``agnostic to the optimization" and it ``does not require any hyper-parameter retuning".

In September 2019, researchers from the University of Passau published \cite{julian} in which they aimed to predict the performance of Convolutional Neural Networks (CNNs) using its structural properties. Authors build Sparse Neural Networks (ANNs) by embedding Directed Acyclic Graphs (DAG) obtained through Random Graph Generators into Artificial Neural Networks. Using this approach, they create a dataset of 10000 such graphs, split them into the train-test set, and train it on the MNIST (\cite{mnist}) dataset. We will follow a similar approach to obtain arbitrarily structured Sparse Recurrent Neural Networks (Sparse-RNNs). Based on its performance, we will examine the importance of specific structural properties of the internal structure based on its impact on the performance of Sparse-RNNs.  This approach can also help in Neural Architectural Search (NAS) for RNNs using performance prediction as a tool.

In November 2019, Zhang et al. proposed a new one-shot pruning approach for Recurrent Neural Networks in \cite{zhang} obtained from the recurrent Jacobian spectrum. According to the authors, this technique works by ``forcing the network to preserve weights that propagate information through its temporal depths". Authors verified their approach with a GRU network on sequential MNIST (\cite{mnist}), Wikitext (\cite{wikitext}), and Billion words (\cite{billion_words}). Authors claim that ``At 95\% sparsity, our network achieves better results than fully dense networks, randomly pruned networks, SNIP (\cite{one_shot}) pruned networks, and Foresight (\cite{foresight}) pruned networks".

In December 2019, Wen et al. in \cite{wen} explain a different way of applying pruning to speedup Recurrent Neural Networks while avoiding the Lasso-based pruning methods. They introduced two sets of binary random variables to generate sparse masks for the weight matrix that, according to the authors, ``act as gates to the neurons". As explained in the paper, ``the presence of the matrix entry $w_{ij}$ depends on both the presence of the $i$-th input unit and the $j$-th output unit, while the value of $w_{ij}$ indicates the strength of the connection if $w_{ij} \neq 0$". The optimization of these variables is then performed by minimizing the $L_0$ norm of the weight matrix. This approach to language modeling and machine reading comprehension works comparatively to the state-of-the-art pruning approaches. Authors claim ``nearly 20$\times$ practical speedup during inference was achieved without losing performance for the language model on the Penn Treebank dataset". 

Now, after briefly acknowledging the research work done in terms of sparsity in RNNs, in the next chapter, we briefly explain the theoretical background of different approaches and algorithms that we will later use as part of our experiments.
\chapter{Technical background}\label{chap:background}

Before describing pruning and random structures, we must know how RNNs and their different variants, namely LSTM and GRU, operate. Therefore, in this section, we provide a detailed explanation of RNNs, LSTM, and GRU.

To properly understand how RNNs operate, we must understand how a perceptron, a building block of neural networks, works.


\section{Perceptron}\label{section:perceptron}

A perceptron, also known as a Single Layer Perceptron (SLP), is a single layer neural network developed by F. Rosenblatt in \cite{Rosenblatt}, inspired by \cite{mcculloch}. It is a linear classifier that accepts multiple binary inputs and returns a single binary output.

\begin{figure}[h]
	\centering
	\includegraphics[width=0.7\linewidth]{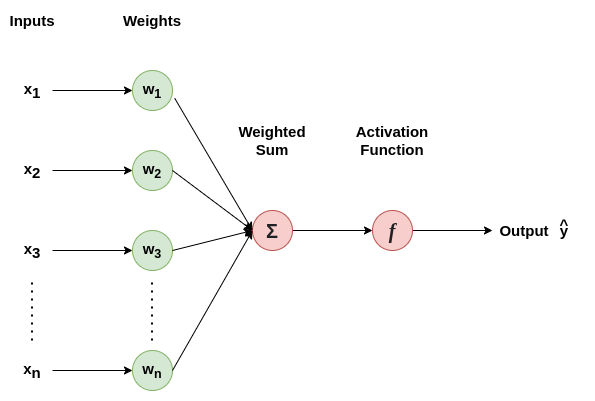}
	\caption[Single Layer Perceptron]%
	{\textbf{Single Layer Perceptron} with binary inputs $x_1, x_2, x_3, ..., x_n$ and its corresponding weights $w_1, w_2, w_3, ..., w_n$.}
	\label{fig:perceptron}
\end{figure}

As shown in the above figure, a perceptron consists of four main parts, inputs, weights, weighted sum, and an activation function. A perceptron works by following these simple steps:

\begin{enumerate}
    \item Multiply each input $x$ with its corresponding weight $w$.
    \item Get a value for the weighted sum by adding all the multiplied values together.
        \begin{equation}
        \label{eqn:weighted_sum}
            weighted\; sum = \sum_{i=1}^{n}w_i.x_i
        \end{equation}
    \item Apply this weighted sum to a activation function to generate the output $\hat{y}$.
\end{enumerate}

An activation function is a significant part of a perceptron. It transforms the input of the node into the output for that node. It ensures the output value is mapped between (0, 1) or (-1, 1). Rectified Linear Unit (ReLU), Hyperbolic Tangent (Tanh) are two of the popular nonlinear activation functions explained later in the following section.

\subsection{Nonlinearity (Nonlinear Activation function)}\label{subsection:nonlinearity}

A nonlinearity, as the name suggests, is used when it is not possible to produce an output for any unit using a linear function. Concerning neural networks, three of the most widely used nonlinearities are, ReLU, Sigmoid, Tanh \cite{nonlin}.

\subsubsection{ReLU}\label{subsubsection:relu}

A \textit{Rectified Linear Unit} is a nonlinear activation function mathematically defined as:

\begin{equation}
    \label{eqn:relu}
    y = max(0, x)
\end{equation}

As per the above equation, for a given input $x$, if the value of $x$ is less than $0$, it returns $0$, otherwise $x$.

The following figure shows the line plot of the ReLU activation function.

\begin{figure}[h]
    \centering
    \includegraphics[width=0.5\linewidth]{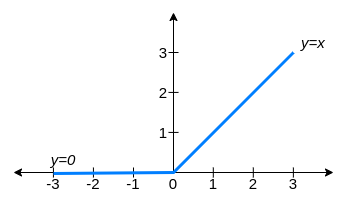}
    \caption[Rectified Linear Unit]{Line plot of ReLU activation function}
    \label{fig:relu}
\end{figure}

For any given positive input, the derivative of ReLU simply returns 1. This eliminates the need to perform computationally expensive exponentials and divisions, as required by other activation functions such as the Sigmoid activation function.

\subsubsection{Sigmoid}\label{subsubsection:sigmoid}

The \textit{Sigmoid} activation function, unlike ReLU, is mainly used in \textit{Feedforward Neural Networks}. This activation function returns a value between $0$ and $1$. The following figure shows the line plot of the Sigmoid activation function.

\begin{figure}[h]
    \centering
    \includegraphics[width=0.7\linewidth]{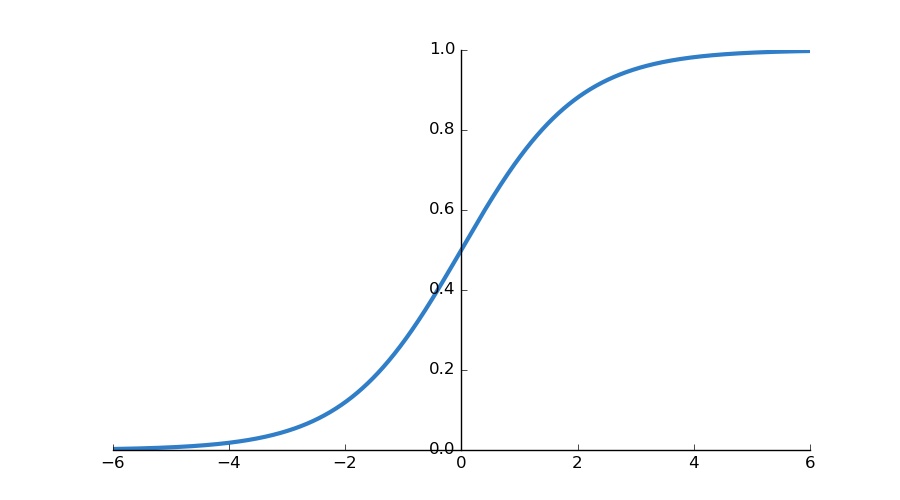}
    \caption[Sigmoid activation function]{Line plot of Sigmoid activation function \cite{sigmoid}}
    \label{fig:sigmoid}
\end{figure}

For a given input $x$, the Sigmoid activation is mathematically written as:

\begin{equation}
    \label{eqn:sigmoid}
    S(x) = \frac{1}{1 + e^{-x}}
\end{equation}

As mention by Nwankpa et al. in \cite{activation}, this activation function has many drawbacks, including gradient saturation and slow convergence. Some of these shortcomings are possible to avoid using other forms of activation functions such as \textit{hyperbolic tangent}.

\subsubsection{Tanh}\label{subsubsection:tanh}

Tanh, short for \textit{hyperbolic tangent} is an activation function with a value range between $-1$ to $1$, making it a zero-centered activation function. The following figure shows the line plot of the Tanh activation function.

\begin{figure}[h]
    \centering
    \includegraphics[width=0.6\linewidth]{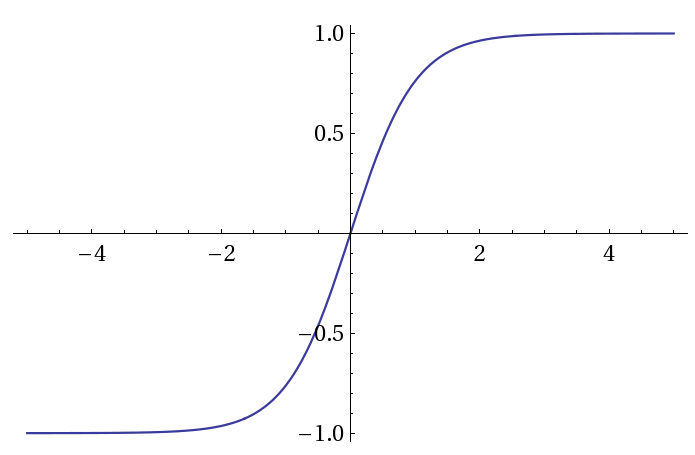}
    \caption[Tanh activation function]{Line plot of Tanh activation function \cite{tanh}}
    \label{fig:tanh}
\end{figure}

For a given input $x$, the Tanh activation is mathematically written as:

\begin{equation}
    \label{eqn:tanh}
    tanh(x) = \frac{e^{x} - e^{-x}}{e^{x} + e^{-x}}
\end{equation}

Although compared to the Sigmoid activation function, Tanh has better performance during training \cite{tanh_1, tanh_2}, it suffers from the vanishing gradient problem.

The output from a nonlinear activation function is not a perfect match to the actual target. For this reason, it is necessary to optimize weight values such that the difference between the actual target and the final output is the smallest, which is done by a process called gradient descent.

\subsection{Batch Gradient Descent}\label{subsection:bgd}

A perceptron has fixed input and output, meaning we can only modify and improve weights to minimize errors. An error function $E$ returns the deviation of predicted outcome $\hat{y}$ from the actual one $y$ as sum of squared errors:

\begin{equation}
    \label{eq:error_func}
    E(w) = \frac{1}{2} \sum_{i=1}^{n}(\hat{y}^{(i)} - y^{(i)})^2
\end{equation}

Weights are then updated using this error function as:

\begin{equation}
    \label{eq:weight_update}
    w \coloneqq w - \eta \nabla E(\text{w})
\end{equation}

where $\eta$ is the learning rate and $\nabla E(\text{w})$ is partial derivative of the cost function, computed for each weight in the weight vector as:

\begin{equation}
    \label{eq:part_der}
    \nabla E(\text{w}) = \frac{\partial E(\text{w})}{\partial w_j}
\end{equation}

By substituting the value of $E(\text{w})$ from equation \ref{eq:error_func} in above equation, we can derive $\nabla E(\text{w})$, as shown by \cite{perc_eq}, as:

\begin{align}
    \nabla E(\text{w}) &= \frac{\partial}{\partial w_j}\frac{1}{2} \sum_{i=1}^{n}(\hat{y}^{(i)} - y^{(i)})^2 \nonumber \\
                       &= \frac{1}{2} \sum_{i=1}^{n}\frac{\partial}{\partial w_j}(\hat{y}^{(i)} - y^{(i)})^2 \nonumber \\
                       &= \frac{1}{2} \sum_{i=1}^{n}2(\hat{y}^{(i)} - y^{(i)})\frac{\partial}{\partial w_j}(\hat{y}^{(i)} - y^{(i)}) \nonumber \\
                       &= \sum_{i=1}^{n}(\hat{y}^{(i)} - y^{(i)}) \frac{\partial}{\partial w_j}(\hat{y}^{(i)} - \sum_{j}w_{j}.x_{j}^{(i)}) \nonumber \\
                       &= \sum_{i=1}^{n}(\hat{y}^{(i)} - y^{(i)})(-x_{j}^{(i)}) \label{eq:part_derived}
\end{align}

By substituting the value from equations \ref{eq:part_derived} in \ref{eq:weight_update}, we can re-write the weight update formula as:

\begin{equation}
    w \coloneqq w + \eta \sum_{i=1}^{n}(\hat{y}^{(i)} - y^{(i)})(x_{j}^{(i)})
\end{equation}

This approach for updating weights is known as Batch Gradient Descent because each sample in the training set is considered at each step of weights update. Repetition of these training and weights update steps is necessary to obtain convergence.

An SLP has no hidden layers. By adding one or more hidden layers, we can generate a Multi-Layer Perceptron (MLP), also known as an Artificial Neural Network (ANN) or simply, a Neural Network.


\newpage
\section{Artificial Neural Network}\label{section:ann}

An Artificial Neural Network is a network of neurons that tries to capture the essential features of the given inputs to infer rules needed to complete a given task, such as image recognition or machine translation. It achieves this by a series of one or more hidden layers, as shown in the below figure:

\begin{figure}[h]
	\centering
	\includegraphics[width=0.7\linewidth]{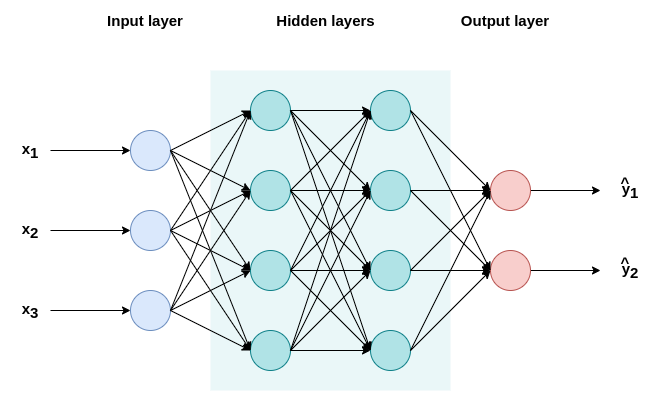}
	\caption[Artificial Neural Network]%
	{\textbf{Artificial Neural Network} with two hidden layers, $h_1$ and $h_2$, each consisting of a plethora of neurons.}
	\label{fig:neural_network}
\end{figure}

The neural network shown in the above figure is a Feedforward Neural Network (FFN), where the output of one layer is an input for the next layer. Each neuron of one layer is directly connected to all other neurons of the next layer. Due to this dense connectivity between neurons of each layer, such layers are called dense layers.

As Goodfellow et al. described in \cite{goodfellow}, the goal of a Feedforward Neural Network is to approximate some function $f^*$. According to the authors, It does so by defining a mapping $y = f(x;\theta)$ that maps the given input $x$ to a corresponding classification category $y$. The feedforward network learns the value of $\theta$ that returns the best function approximation.

The input layers accept the data from the outside world and transfer it directly to the first hidden layer without performing any computations.

Hidden layers are helpful when the linear separation of the data is not possible. Each neuron in a hidden layer is a perceptron that accepts inputs, computes a weighted sum (eq. \ref{eqn:weighted_sum}), applies the weighted sum to an activation function, and passes it towards the next layer.

The output layer of a neural network returns the final results based on the input it receives from its previous layer. The number of neurons in an output layer must match the expected outputs of its respective classification problem.  For example, if a neural network's task is to recognize the hand-written digits shown in figure \ref{fig:digits}, then the output layer of this neural network must have ten neurons where each neuron corresponds to a number from $0$ to $9$.

As with perceptron, neural networks also aim at minimizing the error. Due to the availability of hidden layers, the error is propagated backward using a chain rule, as explained in the following section.

\subsection{Backpropagation}\label{subsection:backprop}

Although the idea of backpropagation was initially pitched in the 1970s, it became famous by the work of Rumelhart et al. in \cite{backprop86} published in 1986. That paper showed the importance of this algorithm by demonstrating that it works better than other earlier approaches to learning. Given an error function for a neural network, the backpropagation algorithm computes the gradient of this error function. This computation happens backward, from the final layer to the first layer.

To properly understand how chain rule applies in error backpropagation, consider a simple neural network shown in below figure:

\begin{figure}[h]
	\centering
	\includegraphics[width=0.7\linewidth]{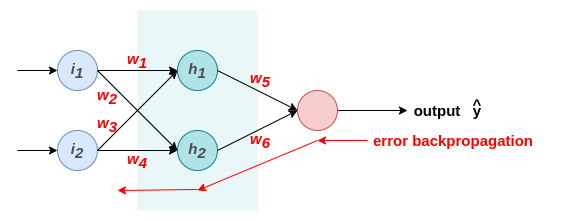}
	\caption[Backpropagation in a neural network]%
	{\textbf{Backpropagation} in a neural network with one hidden layer consisting of two hidden neurons $h_1$ and $h_2$, two input neurons $i_1$ and $i_2$, and one output.}
	\label{fig:back_prop}
\end{figure}

After getting output from the forward pass, we calculate the error using equation \ref{eq:error_func}. By using the partial derivative of this error, the following equation updates all the weights of the neural network:

\begin{equation}
    \label{eq:backprop_wu}
    w_j \coloneqq w_j - \eta \frac{\partial E}{\partial w_j}
\end{equation}

As stated before, we start error backpropagation from the final layer, i.e., the output layer of our neural network shown in figure \ref{fig:back_prop}. Consider the following figure that is visualizing the output layer of the neural network.

\begin{figure}[h]
	\centering
	\includegraphics[width=0.4\linewidth]{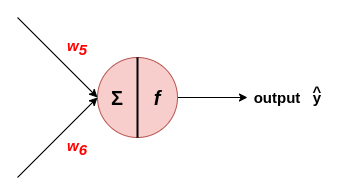}
	\caption[Backpropagating error from output layer]%
	{Backpropagating error from output layer to modify weights $w_5$ and $w_6$.}
	\label{fig:back_prop_1}
\end{figure}

To update $w_5$ and $w_6$, we must first compute $\frac{\partial E}{\partial w_5}$ and $\frac{\partial E}{\partial w_6}$. However, error $E$ is the difference between target $y$ and predicted output $\hat{y}$. Furthermore, the predicted output $\hat{y}$ is the result of applying the weighted sum to an activation function. Here, the weighted sum is calculated as:

\begin{equation}
    weighted\; sum\; (z) = \hat{y}_{h1} w_5 + \hat{y}_{h2} w_6
\end{equation}

where $\hat{y}_{h1}$ is output of the hidden neuron $h_1$ and $\hat{y}_{h2}$ is output of the hidden neuron $h_2$.

Based on this, we can derive the following formula that computes the partial derivative of $E$ with respect to $w_5$ and $w_6$ as:

\begin{equation}
    \frac{\partial E}{\partial w_5} = \frac{\partial E}{\partial \hat{y}} \frac{\partial \hat{y}}{\partial z} \frac{\partial z}{\partial w_5}
\end{equation}

\begin{equation}
    \frac{\partial E}{\partial w_6} = \frac{\partial E}{\partial \hat{y}} \frac{\partial \hat{y}}{\partial z} \frac{\partial z}{\partial w_6}
\end{equation}

We follow the similar process to compute the partial derivative of error $E$ with respect to $w_1$, $w_2$, $w_3$, and $w_4$.

Consider the following figure that is visualizing the hidden neuron $h_1$ of the neural network:
\begin{figure}[h]
	\centering
	\includegraphics[width=0.4\linewidth]{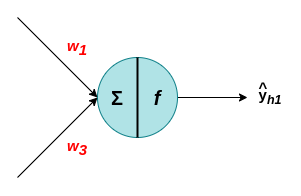}
	\caption[Backpropagating error from $h_1$]%
	{Backpropagating error from $h_1$ to modify weights $w_1$ and $w_3$.}
	\label{fig:back_prop_2}
\end{figure}

To update $w_1$ and $w_3$, we first compute $\frac{\partial E}{\partial w_1}$ and $\frac{\partial E}{\partial w_3}$ as:

\begin{equation}
    \frac{\partial E}{\partial w_1} = \frac{\partial E}{\partial \hat{y}_{h1}} \frac{\partial \hat{y}_{h1}}{\partial z_{h1}} \frac{\partial z_{h1}}{\partial w_1}
\end{equation}

\begin{equation}
    \frac{\partial E}{\partial w_3} = \frac{\partial E}{\partial \hat{y}_{h1}} \frac{\partial \hat{y}_{h1}}{\partial z_{h1}} \frac{\partial z_{h1}}{\partial w_3}
\end{equation}

Here, $\hat{y}_{h1}$ is output of the hidden neuron $h_1$ and $z_{h1}$ is the weighted sum used to compute $\hat{y}_{h1}$, calculated as:
\begin{equation}
    z_{h1} = i_1 w_1 + i_2 w_3
\end{equation}

Now, consider the following figure that is visualizing the hidden neuron $h_2$ of the neural network:
\begin{figure}[h]
	\centering
	\includegraphics[width=0.4\linewidth]{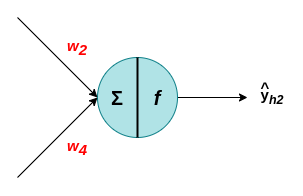}
	\caption[Backpropagating error from $h_2$]%
	{Backpropagating error from $h_2$ to modify weights $w_2$ and $w_4$.}
	\label{fig:back_prop_3}
\end{figure}

To update $w_2$ and $w_4$, we first compute $\frac{\partial E}{\partial w_2}$ and $\frac{\partial E}{\partial w_2}$ as:

\begin{equation}
    \frac{\partial E}{\partial w_2} = \frac{\partial E}{\partial \hat{y}_{h2}} \frac{\partial \hat{y}_{h2}}{\partial z_{h2}} \frac{\partial z_{h2}}{\partial w_2}
\end{equation}

\begin{equation}
    \frac{\partial E}{\partial w_4} = \frac{\partial E}{\partial \hat{y}_{h2}} \frac{\partial \hat{y}_{h2}}{\partial z_{h2}} \frac{\partial z_{h2}}{\partial w_4}
\end{equation}

Here, $\hat{y}_{h2}$ is output of the hidden neuron $h_2$ and $z_{h2}$ is the weighted sum used to compute $\hat{y}_{h2}$, calculated as:
\begin{equation}
    z_{h2} = i_1 w_2 + i_2 w_4
\end{equation}

Once we have the partial derivatives of error $E$ with respect to weights, we can use equation \ref{eq:backprop_wu} to update each weight, and then we train the neural network with updated weights. This process of weights update and re-training is repeated until the neural network converges.

Each layer in a traditional neural network is densely connected, making it a complex architecture. Also, in such a neural network, information travels only in the forward direction, i.e., there are no feedback loops available where the input to a function also depends on the output. However, there exist other variations of neural networks that are either computationally less complex (i.e., Convolutional Neural Networks) or support feedback loops (i.e., Recurrent Neural Networks).

A Convolutional Neural Network (CNN) ensures less computational complexity by forcing neurons of one layer to share weights. This sharing of weights reduces the number of learnable parameters, allowing for a better generalization. This process is based on the neocognitron network, published by K. Fukushima in \cite{neocognitronbc}. Results of \cite{7382560, 7822567} show that, in the areas of image recognition and classification, CNNs are very useful. However, similar to Feedforward Neural Networks, CNNs also do not have feedback loops. For this reason, Recurrent Neural Networks are better than CNNs while working on tasks where sequence order is essential such as machine translation or music composition.


\newpage
\section{Recurrent Neural Network}\label{section:rnn}

\begin{wrapfigure}{r}{6.4cm}
    \centering
    \includegraphics[width=1.0\linewidth]{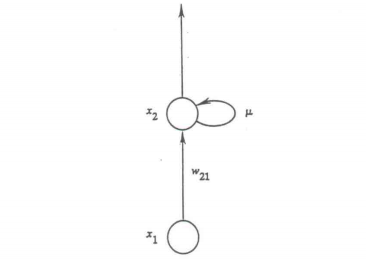}
    \caption[A simple Recurrent Network]{A simple Recurrent Network as depicted in \cite{jordan}}
    \label{fig:jordan_rnn}
\end{wrapfigure} 

As stated before, one of the drawbacks of the standard neural network model is the lack of feedback loops. In 1986, M. Jordan published \cite{jordan}, in which he described Recurrent Networks as networks that have a connection from a unit to itself, i.e., a recurrent connection. This recurrent connection act as a feedback loop that makes it possible to use the previous outputs as inputs. This property of Recurrent Neural Networks makes them suitable to work with sequential information where all the inputs depend on each other.

As I. Sutskever describes in \cite{ilya}, a Recurrent Neural Network uses hidden states to incorporate new observations using an intricate nonlinear function. Such a Recurrent Neural Network can effortlessly be understated when it is unfolded in time, as shown in the figure:

\begin{figure}[h]
	\centering
	\includegraphics[width=0.9\linewidth]{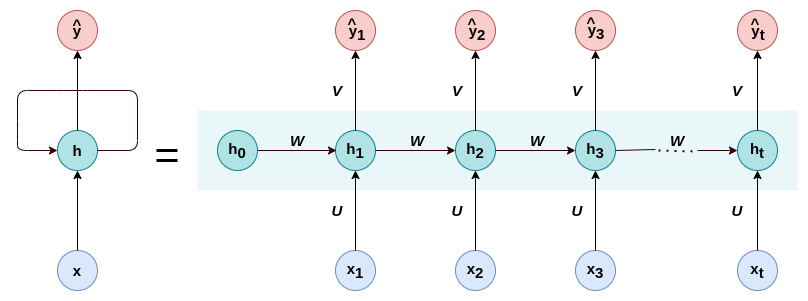}
	\caption[A Recurrent Neural Network unfolded in time]%
	{A simple \textbf{Recurrent Neural Network} unfolded in time with $t$ sequences. Here, $U$ represents \textit{input-to-hidden weights}, $W$ represents \textit{hidden-to-hidden weights}, and $V$ represents \textit{hidden-to-output weights}.}
	\label{fig:rnn}
\end{figure}

The depicted Recurrent Neural Network accepts inputs $x_1, x_2, x_3, ..., x_t$, and outputs $\hat{y}_1, \hat{y}_2, \hat{y}_3, ..., \hat{y}_t$. The hidden states $h_0, h_1, h_2, h_3, ..., h_t$ are high-dimensional vectors, connected to each other to create recurrence. Deep extensions of a basic RNN can be constructed by stacking multiple recurrent hidden states on top of each other as shown in \cite{deeprnn}. Bias vectors $b_h$ and $b_o$, although are not shown in the above figure, are optional, but are useful to shift the activation function.

The RNN uses the following algorithm to compute $h_t$ and $\hat{y}_t$:

\begin{algorithm}
  \caption[Standard RNN algorithm]%
  {A standard RNN algorithm}
  \label{alg:rnn}
    \For{$t$ \textbf{from} $1$ \textbf{to} $T$}{
        \DontPrintSemicolon
        $z_{t} \gets U x_{t} + W h_{t-1} + b_{h}$ \tcp*{$b_{h}$ is optional}
        $h_{t} \gets e$($z_{t}$) \\
        $o_{t} \gets V h_{t} + b_{o}$ \tcp*{$b_{0}$ is optional}
        $\hat{y}_{t} \gets g$($o_{t}$)
    }
\end{algorithm}

where $e$($\cdot$) and $g$($\cdot$) are the hidden and output nonlinearities of the RNN. This computation of $h_t$ is visualized in the following figure:

\begin{figure}[h]
	\centering
	\includegraphics[width=0.6\linewidth]{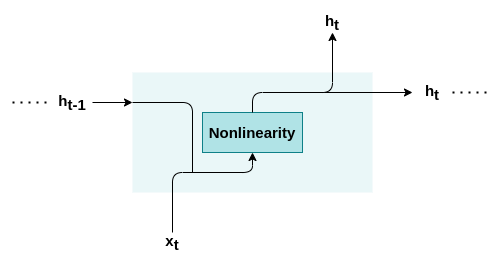}
	\caption[Internal structure of a single hidden unit in a standard RNN]%
	{The internal structure of a single hidden unit in a standard RNN visualizing the computation of $h_t$ using an input $x_t$, and the hidden state value of the previous unit $h_{t-1}$.}
	\label{fig:rnn_unit}
\end{figure}

The nonlinearity is required to produce a nonlinear decision boundary. Out of all three nonlinearities explained in section \ref{subsection:nonlinearity}, for the scope of this thesis, we only focus on ReLU and Tanh due to their added advantages over the Sigmoid activation function.

Similar to error backpropagation in traditional neural networks (as explained in section \ref{subsection:backprop}), RNNs also backpropagate error to update weights and minimize the absolute error. However, in recurrent networks, since the output of one time-step depends on previous ones, the error also backpropagates from time-step $t$ through the entire network to the first time-step. This is known as Backpropagation Through Time (BPTT \cite{bptt-1}).

\subsection{Backpropagation Through Time}\label{subsection:bptt}

To understand backpropagation through time, consider the following simple RNN with three time-steps visualizing error backpropagation from time-step 3:

\begin{figure}[h]
    \centering
    \includegraphics[width=0.7\linewidth]{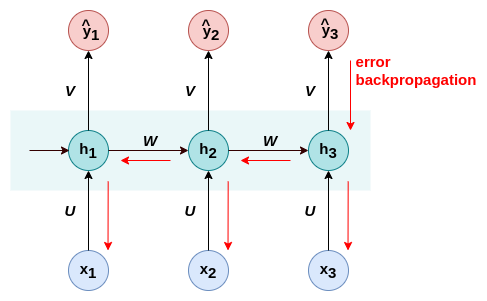}
    \caption[Backpropagation Through Time]{A standard RNN visualizing BPTT from time-step 3. Here, $U$ represents \textit{input-to-hidden weights}, $W$ represents \textit{hidden-to-hidden weights}, and $V$ represents \textit{hidden-to-output weights}.}
    \label{fig:bptt}
\end{figure}

Error at time-step 3 $E_3$ is the difference between target output $y_3$  and predicted output $\hat{y}_3$. This error $E_3$ backpropagates from time-step 3 to time-step 1, as shown in the above figure. 

To start, using the algorithm \ref{alg:rnn}, we can write following equation for output $\hat{y}_3$:

\begin{equation}
    \hat{y}_3 = g(o_3)
\end{equation}

where $o_3$ is product of $V$ and $h_3$ given as:

\begin{equation}
    o_3 = V h_3
\end{equation}

where $V$ represents \textit{hidden-to-output} weights, $e$ is output nonlinearity, and $h_3$ is calculated as:

\begin{equation}
    h_3 = e(z_3)
\end{equation}

where, $z_3$ is the weighted sum computed as:

\begin{equation}
    \label{eq:z3}
    z_3 = Ux_3 + Wh_2
\end{equation}

where $U$ is \textit{input-to-hidden} weights, $W$ is \textit{hidden-to-hidden} weights, and $e$ is hidden nonlinearity.

As we can see in the above equations, to compute output $\hat{y}_3$, we need a total of three different weight vectors $V$, $W$, and $U$. Since our error $E_3$ is dependent on output $\hat{y}_3$, we must calculate the partial derivative of the error $E_3$ with respect to all three weight vectors, i.e., we need to compute $\frac{\partial E_3}{\partial V}$, $\frac{\partial E_3}{\partial W}$, and $\frac{\partial E_3}{\partial U}$. To compute these gradients, we follow the same chain rule as we did in backpropagation (section \ref{subsection:backprop}).

Calculating $\frac{\partial E_3}{\partial V}$ is easy as it only depends on $\hat{y}_3$, $o_3$. Therefore, by following the simple backpropagation, we can calculate $\frac{\partial E_3}{\partial V}$ as:

\begin{equation}
    \frac{\partial E_3}{\partial V} = \frac{\partial E_3}{\partial \hat{y}_3} \frac{\partial \hat{y}_3}{\partial o_3} \frac{\partial o_3}{\partial V}
\end{equation}

Calculation of $\frac{\partial E_3}{\partial W}$ is where calculation gets more complicated. To see why it is complicated from $\frac{\partial E_3}{\partial V}$, we first apply the chain rule as:

\begin{equation}
    \label{eq:wrt_W}
    \frac{\partial E_3}{\partial W} = \frac{\partial E_3}{\partial \hat{y}_3} \frac{\partial \hat{y}_3}{\partial h_3} \frac{\partial h_3}{\partial W}
\end{equation}

However, as we can see in equation \ref{eq:z3}, $z_3$ depends on $h_2$, which depends on $h_1$. For this reason, we cannot treat $h_2$ as constant; instead, we need to sum up the contributions of $h_2$ and $h_1$ as:

\begin{align}
    \frac{\partial h_3}{\partial W} &= \frac{\partial h_3}{\partial W} + \frac{\partial h_3}{\partial h_2} \frac{\partial h_2}{\partial W} + \frac{\partial h_3}{\partial h_2}\frac{\partial h_2}{\partial h_1}\frac{\partial h_1}{\partial W} \nonumber \\
                                    &= \frac{\partial h_3}{\partial h_3} \frac{\partial h_3}{\partial W} + \frac{\partial h_3}{\partial h_2} \frac{\partial h_2}{\partial W} + \frac{\partial h_3}{\partial h_1} \frac{\partial h_1}{\partial W} \nonumber \\
                                    &= \sum_{t=1}^{3} \frac{\partial h_3}{\partial h_t} \frac{\partial h_t}{\partial W} \label{eq:sum_h}
\end{align}

By substituting value from equation \ref{eq:sum_h} in equation \ref{eq:wrt_W}, we get

\begin{equation}
    \frac{\partial E_3}{\partial W} = \frac{\partial E_3}{\partial \hat{y}_3} \frac{\partial \hat{y}_3}{\partial h_3} \left[ \sum_{t=1}^{3} \frac{\partial h_3}{\partial h_t} \frac{\partial h_t}{\partial W} \right]
\end{equation}

We follow the similar process to calculate $\frac{\partial E_3}{\partial U}$ as:

\begin{equation}
    \frac{\partial E_3}{\partial U} = \frac{\partial E_3}{\partial \hat{y}_3} \frac{\partial \hat{y}_3}{\partial h_3} \left[ \sum_{t=1}^{3} \frac{\partial h_3}{\partial h_t} \frac{\partial h_t}{\partial U} \right]
\end{equation}

Similar to the backpropagation algorithm, we use these partial derivatives of error with respect to $V$, $W$, and $U$ to update the corresponding weights.

In standard RNNs, each hidden unit only performs the specified nonlinear activation (as depicted in figure \ref{fig:rnn_unit}). However, there exist other variations of RNNs, such as Long Short-Term Memory, Gated Recurrent Unit, that do more than just performing a single computation per hidden unit.


\newpage
\section{Long Short-Term Memory}\label{section:lstm}

Two of the problems with this algorithm are of exploding gradients\footnote{Exploding gradients is a problem when large error gradients accumulate, resulting in substantial updates to model weights making the model unstable.} or vanishing gradient\footnote{The vanishing gradient is a problem that prevents making changes in the weight values due to vanishingly small gradients.}. Two mitigate these problems, in 1997, Hochreiter et al. proposed a new recurrent architecture, termed Long Short-Term Memory (LSTM) in \cite{lstm}.

As opposed to the standard RNN's internal structure (figure \ref{fig:rnn_unit}), LSTM has a very different and much more complex internal structure made up of three gates (i.e., an \textit{input gate $i_t$}, an \textit{output gate $o_t$}, and a \textit{forget gate $f_t$}) to regulate the flow of information \cite{lstm_gates}. This is visualized in the following figure:

\begin{figure}[h]
	\centering
	\includegraphics[width=0.8\linewidth]{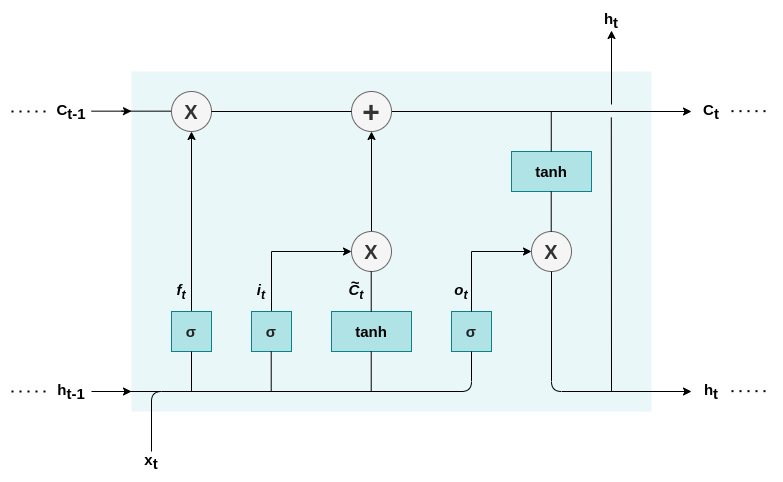}
	\caption[Internal structure of a single hidden unit in an LSTM]%
	{The internal structure of a single hidden unit in an LSTM visualizing the computation of $h_t$ and $C_t$ using an input $x_t$, the hidden state value of the previous unit $h_{t-1}$, and the cell state value of the previous unit $C_{t-1}$.}
	\label{fig:lstm_unit}
\end{figure}

Given an input $x_t$, previous hidden state $h_{t-1}$, and previous cell state $C_{t-1}$, the current hidden state $h_t$ and cell state $C_t$ can mathematically be calculated as follow:

\begin{enumerate}
    \item The first step is to compute the forget gate that informs the cell state about which past information to keep       and which one to forget.
        \begin{equation}
            \label{eqn:forget_gate}
            f_t = \sigma(W_f \cdot [h_{t-1}, x_t] + b_f)
        \end{equation}
        For each value in the cell state $C_{t-1}$, it returns a value between $0$ and $1$.
        
    \item The second step is to determine what new information needs to be stored in the cell state.
        \begin{equation}
            \label{eqn:input_gate}
            i_t = \sigma(W_i \cdot [h_{t-1}, x_t] + b_i)
        \end{equation}
        \begin{equation}
            \widetilde{C} _t = tanh(W_C \cdot [h_{t-1}, x_t] + b_C)
        \end{equation}
        The addition of these two values then will be used to update the previous cell state.
        
    \item The third step is to update the previous cell state as:
        \begin{equation}
            C_t = f_t * C_{t-1} + i_t * \widetilde{C}_t
        \end{equation}
        
    \item The final step is to compute the value for the output gate, that then will be used to update the hidden state       as:
        \begin{equation}
            \label{eqn:output_gate}
            o_t = \sigma(W_o \cdot [h_{t-1, x_t}] + b_o)
        \end{equation}
        \begin{equation}
            h_t = o_t * tanh(C_t)
        \end{equation}
    
    These updated state values, $h_t$, and $C_t$ are then forwarded to be used in the next unit.
\end{enumerate}

One of the benefits of LSTM over a standard RNN is the capability of learning long-term dependencies. Because of this added benefit, LSTMs have shown exceptional performance in many application areas including time series forecasting \cite{lstm_time}, drug design \cite{lstm_drug}, and music composition \cite{lstm_music}.

Although LSTMs have a better performance than standard RNNs, their execution is slower due to the availability of a higher number of computations. Therefore, another variation of RNNs, the Gated Recurrent Unit, is faster than an LSTM but has a better performance than a standard RNN.


\newpage
\section{Gated Recurrent Unit}\label{section:gru}

The Gated Recurrent Unit (GRU) was introduced by Cho et al. in \cite{gru}. Similar to Long Short-Term Memory, Gated Recurrent Unit also aims to solve the vanishing gradient problem.

The internal structure of a GRU is very different from that of a standard RNN unit but is quite similar to that of an LSTM unit, except with only two gates (i.e., an \textit{update gate} $z_t$, and a \textit{reset gate} $r_t$) than three. These two gates control the flow of information that flows into and out of the memory \cite{gru_gates}. This internal structure is visualized in the following figure:

\begin{figure}[h]
	\centering
	\includegraphics[width=0.8\linewidth]{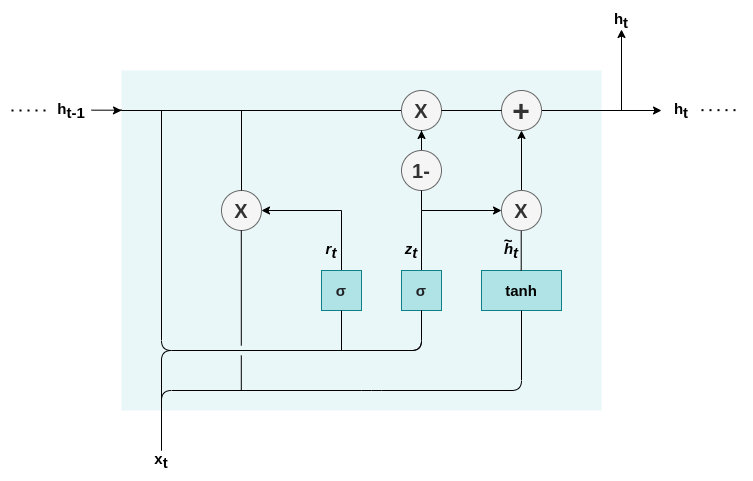}
	\caption[Internal structure of a single hidden unit in a GRU]%
	{The internal structure of a single hidden unit in a GRU visualizing the computation of $h_t$ using an input $x_t$, and the hidden state value of the previous unit $h_{t-1}$.}
	\label{fig:gru_unit}
\end{figure}

Given an input $x_t$, and previous hidden state $h_{t-1}$, the current hidden state $h_t$ can mathematically be calculated as follow:

\begin{enumerate}
    \item The first step is to compute the update gate value as:
        \begin{equation}
            \label{eqn:update_gate}
            z_t = \sigma(W_z \cdot [h_{t-1}, x_t] + b_z)
        \end{equation}
        This gate helps to determine what information from the past to carry forward.
        
    \item The second step is to compute the value of the reset gate as:
        \begin{equation}
            \label{eqn:reset_gate}
            r_t = \sigma(W_r \cdot [h_{t-1}, x_t] + b_r)
        \end{equation}
        This gate decides how much of the past information to omit.
        
    \item The final step is to update the hidden state using the values of the update gate and the reset gate obtained        from equations \ref{eqn:update_gate} and \ref{eqn:reset_gate}, respectively.
        \begin{equation}
            \widetilde{h}_t = tanh(W_h \cdot [r_t * h_{t-1}, x_t] + b_h)
        \end{equation}
        \begin{equation}
            \label{eqn:update_hidden}
            h_t = (1 - z_t) * h_{t-1} + z_t * \widetilde{h}_t
        \end{equation}
        
    This updated hidden state value $h_t$ is then forwarded to be used in the next unit.
\end{enumerate}

Although LSTMs mostly outperform GRUs in many scenarios, GRUs have the upper hand when the dataset is small \cite{gru_lstm}. Similar to LSTM, GRUs are also capable of learning long-term dependencies. Because of this, GRUs have shown outstanding performance in different application areas, including speech recognition \cite{gru_speech} and water level prediction \cite{gru_water_level}.

In the scope of this thesis, we only focus on working with RNN with Tanh nonlinearity (RNN-Tanh), RNN with ReLU nonlinearity (RNN-ReLU), Long Short-Term Memory (LSTM), and Gated Recurrent Unit (GRU) as these four are the most popularly practiced recurrent networks.


\newpage
\section{Pruning}\label{section:pruning}

It is well known that the use of more extensive neural networks is favorable to achieve better performance, but at the same time, they are more expensive to use, takes more time to run, and in most cases, require expensive hardware. Therefore, in the past couple of decades, many researchers have worked on various model compression techniques to train large neural networks more efficiently while maintaining their accuracy. Pruning is one such technique that removes the top $k$ ranked weight parameters of a neural network. The weight parameters removed first are usually the ones with the most negligible impact on the model's final output.

In \cite{blalock}, Blalock et al. studied 81 different papers on pruning to identify important (hyper-)parameters and methods that give the best results with pruning compared to others. In that paper, the authors state that ``pruning methods vary primarily in their choices regarding sparsity structure, scoring, scheduling, and fine-tuning". Here, the sparse structure results from performing unstructured pruning in which individual weight parameters are pruned, and fine-tuning is the process of retraining the pruned neural network to identify the amount of pruning that can be applied without significantly reducing the performance.

There are different pruning techniques such as Magnitude-based pruning, Error-based pruning, Entropy-based pruning, Evolutionary pruning. In the scope of this thesis, we only focus on Magnitude-based pruning as it is proven to give satisfactory results despite being an easy technique.

\subsection{Magnitude-based pruning}

This is the simplest way of pruning that considers the magnitude of weight parameters as pruning criteria, meaning weight parameters that are least important are zeroed out before retraining the model to perform fine-tuning. Li et al. in \cite{li} state the OLMP (\textbf{O}ptimization based \textbf{L}ayer-wise \textbf{M}agnitude-based \textbf{P}runing) can reduce the size of AlexNet by 82\% without any loss inaccuracy. We also apply layer-wise pruning, where we first calculate threshold based on the percent of pruning to apply and zero-out values below this threshold using a binary mask.


\newpage
\section{Graph Theory}\label{section:graphs}

Graph theory is the study of graphs, a network of nodes connected with edges. In \cite{carlson}, Carson states that the graph theory originated from Leonhard Euler's solution to the famous Königsberg bridge problem. This section briefly explores random graphs, their graph properties, and two random graphs with small-world properties, namely Watts–Strogatz and Barabási–Albert.

An undirected graph $G$ is given as a set of $(V, E)$, where $V$ is the set of nodes, and $E$, where $E \subseteq E_{comp} \coloneqq \{\{v, w\} | v, w \in V, v \neq w\}$ is the set of edges. A complete graph, $G_{comp}$, is a graph where each node is connected to every other node in the same network, which is different from a connected graph, where it is possible to get from one node to any other node in the same network via a series of edges.

Graphs, in mathematics, are usually used to model real-world structures, but if the structure is very complex and cannot model it in all details, random graphs are used.

\subsection{Random Graphs}\label{subsection:randomgraphs}

Random graphs, as the name states, have randomly distributed edges according to some probability measure. Based on which adequate probability measure is used, random graphs can be given as:

\begin{enumerate}
    \item \textbf{Uniform random graph}: For the given set of nodes $V$, edges $M$, and a probability space $\Omega$ where all graphs with nodes $V$ and $M$ edges has the uniform distribution, the probability is given as:
    \begin{equation}
        P(G) = \left( M_{comp} \atop M \right)^{-1}
    \end{equation}
    
    \item \textbf{Binomial random graph}: Given the probability $0 \leq p \leq 1$ that there exists an edge between two given nodes $v, w \in V$ and a probability space $\Omega$ where all graphs with nodes $V$ and edges $M$ has the binomial distribution, the probability is given as:
    \begin{equation}
        P(G) = p^M \cdot (1-p)^{M_{comp}-M}
    \end{equation}
\end{enumerate}

\subsection{Graph properties}\label{subsection:properties}

This section explores graph properties that we later use to find a correlation with performance. Along with simple properties such as the number of layers, the number of nodes, and the number of edges, we also use other graph properties, as given below:

\subsubsection{Diameter}
A graph's diameter is the maximal distance between any pair of nodes in a given graph, excluding any detours or loops. We can use the following equation to find the graph's diameter:
\begin{equation}
    \delta = \max_{ij}\{s(i, j)\}
\end{equation}
where $s(i,j)$ is the shortest path between two nodes $i$ and $j$.

\subsubsection{Density}
The density of a graph is the ratio of edges present to all possible edges. Given an undirected graph, the density of this graph is calculated as:
\begin{equation}
    d = \frac{2m}{n(n-1)}
\end{equation}
For a directed graph, the density is calculated as:
\begin{equation}
    d = \frac{m}{n(n-1)}
\end{equation}
where $n$ is the number of nodes, and $m$ is the number if edges.

\subsubsection{Average Shortest Path Length}
The average shortest path length is the mean number of steps on all potential pairs of nodes' shortest paths and is calculated as:
\begin{equation}
    a = \sum_{i,j \in V}\frac{d(i, j)}{n(n-1)}
\end{equation}
where $V$ is the set of nodes, $d(i,j)$ is the shortest path from node $i$ to $j$, and $n$ is the number of nodes.

\subsubsection{Eccentricity}
The eccentricity of a given node $V$ in a connected graph $G$ is the maximum distance from node $V$ to all other nodes in that graph. In contrast, for a disconnected graph, all nodes have an infinite eccentricity.

The maximum a graph can have is the graph diameter, while the minimum eccentricity is the graph radius.

\subsubsection{Degree}
The degree of a node is the number of nodes adjacent to that node.

\subsubsection{Closeness}
Closeness, often known as closeness centrality, indicates how close a node is to all other nodes in a given graph and is calculated as:
\begin{equation}
    C(u) = \frac{n-1}{\sum_{i=1}^{n=1}d(i, j)}
\end{equation}
where $d(i,j)$ is the shortest path between nodes $i$ and $j$, $n$ is the number of nodes in the graph.

\subsubsection{Node betweenness}
Node betweenness, also known as the betweenness centrality of node $u$, is the sum of the fraction of all shortest path pairs that pass through $u$ and is calculated as:
\begin{equation}
    c_B(u) = \sum_{i,j \in V}\frac{\sigma(i,j|u)}{\sigma(i,j)}
\end{equation}
Here, $V$ is the set of nodes, $\sigma(i,j)$ is the number of shortest paths between nodes $i$ and $j$, and $\sigma(i,j|u)$ is the number of shortest paths through node $u$.

\subsubsection{Edge betweenness}
In contrast to node betweenness, Edge betweenness is the number of shortest paths that pass through an edge in a given graph. It is calculated as:
\begin{equation}
    c_B(e) = \sum_{i,j \in V}\frac{\sigma(i,j|e)}{\sigma(i,j)}
\end{equation}
Here, $V$ is the set of nodes, $\sigma(i,j)$ is the number of shortest paths between nodes $i$ and $j$, and $\sigma(i,j|e)$ is the number of shortest paths through an edge $e$.

\subsection{Small World Networks}
In a Small World Network, the mean shortest-path distance between any two given nodes increases sufficiently slowly as a function of the number of nodes in the network \cite{porter}.

A Small World Network have the following three properties:
\begin{enumerate}
    \item \textbf{Higher order of the graph}: The order of a graph is simply the number of nodes in that graph, in other words, the cardinality of its node-set. In a Small World Network, the order of the graph is much higher than its average vertex degree.
    
    \item \textbf{Small characteristic path length}: As stated by F. Schreiber in \cite{schreiber}, the characteristic path length of a given graph is the average number of edges in the shortest paths between all pairs of nodes. For a given graph to be considered a Small World Network, it must have a small characteristic path length.
    
    \item \textbf{High clustering coefficient}: A clustering coefficient is a measure of the degree to which nodes in a graph are clustered together. A given graph must have a higher clustering coefficient to be considered a Small World Network.
\end{enumerate}

Watts–Strogatz is one of the popular Small World Networks, that is explained in the following section.

\subsubsection{Watts–Strogatz model}\label{subsubsection:wsmodel}
Watts–Strogatz (WS) are regular graphs with degree $k$. In 1998, Watts and Strogatz \cite{watts} proposed the Watts-Strogatz model to counteract the incomplete regularity and randomness in social networks.

The construction of a WS model begins with a lattice structure\footnote{A lattice graph is a graph embedded in a Euclidean space $\mathbb{R}$ that has a regular tiling form  \cite{lattice}.} that has a high clustering coefficient ($C$) but also a large characteristic path length ($L$). We can rewire some edges in this lattice structure to reduce $L$.

For each node $i \in V$, for each edge $\{i, j\}$ from node $i$ to one of the $r$ right neighbor $j$, the edge is left unchanged with probability $1-p$, and with probability $p$, the edge is replaced by a new edge $\{i, j\prime\}$ where $j\prime\in V \setminus (N(v) \cup \{v\})$ is randomly chosen.

For a small probability $p$, a sharp drop in characteristic path length ($L$) can be observed with no change in clustering coefficient ($C$).

\subsection{Scale Free Networks}\label{subsection:sfn}
In contrast to Small World Networks, where graphs have a small path length due to local clustering, Scale Free Networks have a skewed degree distribution \cite{aarstad}. Probability distribution of node degrees in such graphs is given as:
\begin{equation}
    P(d(v) = k) \sim k^{-\gamma}
\end{equation}
where $d(v)$ is the degree of node $v \in V$ and $\gamma$ is the power law factor.

As seen in the above equation, this probability distribution has power law where the probability of a node having a degree $k$ varies as the power law factor $\gamma$ varies.

In a Scale Free Network, the system grows with time, and new nodes and edges are added based on the preferential attachment. Such networks are observed in different science and technology areas such as the internet (nodes are HTML pages and edges are hyperlinks), electronic mail (nodes are email addresses and edges are sent emails between two users), and scientific literature (nodes are publications and edges are citations).

\subsubsection{Barabási–Albert model}\label{subsubsection:bamodel}
Barabási–Albert model is one of many Scale Free Networks that follows preferential attachment with power-law growth for a given system.

The construction of a Barabási–Albert model begins with $N_0 > 1$ nodes and $m < N_0$ edges where $N_0 = |V_0|$, cardinality of set of nodes at $t = 0$ without any edges.

At time $t = 1$, a new node is added to $V_0$ with $m$ edges from this new node to $m$ random different nodes.

At time $t > 1$, a new node is added to $V_{t-1}$ with $m$ edges from this new node to $m$ random different nodes, where the preferential attachment gives the probability of choosing node $i$ as:

\begin{equation}
    P(i) = c \cdot d(i)
\end{equation}
\centerline{where, $c = \frac{1}{\sum_{j \in V_{t-1}}d(j)}$}

In any given Barabási–Albert model, for any given time $t$, there are $N_0 + t = N_t$ nodes and $m \cdot t$ edges available.
\chapter{Dataset}\label{chap:dataset}

For our experiments, we make use of a dataset composed of Reber sequences. A Reber sequence is a grammar string made of finite states, in simple words, formed by using a confined set of characters. In the research paper that proposed the LSTM \cite{lstm}, the authors use embedded Reber grammar due to its short time lags. For our purpose, we use a simple version of this embedded Reber grammar, called Reber grammar (\cite{reber}).

The following figure shows the flow diagram to generate Reber grammar sequences:

\begin{figure}[h]
    \centering
    \includegraphics[width=0.9\linewidth]{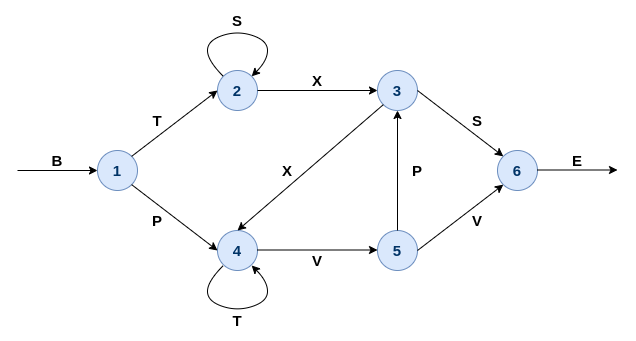}
    \caption[Flow diagram to generate Reber grammar sequences]{Flow diagram to generate Reber grammar sequences}
    \label{fig:reber}
\end{figure}

As shown in the above graph, a sequence is a true Reber sequence if
\begin{enumerate}
    \item it contains only these characters: \textit{B, E, P, S, T, V, X}
    \item it starts with a $B$ and ends with an $E$, \textit{always!}
    \item it strictly follows the transition diagram shown in figure \ref{fig:reber}.
\end{enumerate}

After starting at $B$, we move from one node to the next until we reach $E$. If we have more than one path to choose from, we can randomly select one of them. Two of the allowed characters, $S$ and $T$, have a self-loop at nodes $2$ and $4$, respectively, meaning we can have more than two consecutive $S$s and $T$s. For example, the following table shows some of the true and false Reber sequences from our dataset:

\begin{table}[h]
	\centering
	\begin{tabular}{|c|c|}
	    \hline
		\textbf{True Reber sequences} & \textbf{False Reber sequences} \\
		\hline
		BPTVPXTTVVE & BTTVPXTVPSE \\
		BTSXXTTTTVVE & BPSXXTTTVPSE \\
		BTSSXXTTVVE & BPSSSXXTVVE \\
		BTXXVPXTVVE & BTTTVPXTVVE \\
		BPTTTTTVPSE & BPSSXXTTVVE \\
		\hline
	\end{tabular}
	\caption[A few examples of true and false Reber sequences]{A few examples of true and false Reber sequences.}
	\label{tab:reber}
\end{table}

Since, to our knowledge, there is no public dataset of Reber sequences available, we created our own dataset of $25000$ true and false Reber sequences combined. We follow the same flow, as shown in figure \ref{fig:reber}, to generate true Reber sequences. Then, by slightly altering the same procedure, we generate false Reber sequences.

We then split this dataset into a train-test set, details of which are shown in the following bar chart:

\begin{figure}[h]
    \centering
    \includegraphics[width=1.0\linewidth]{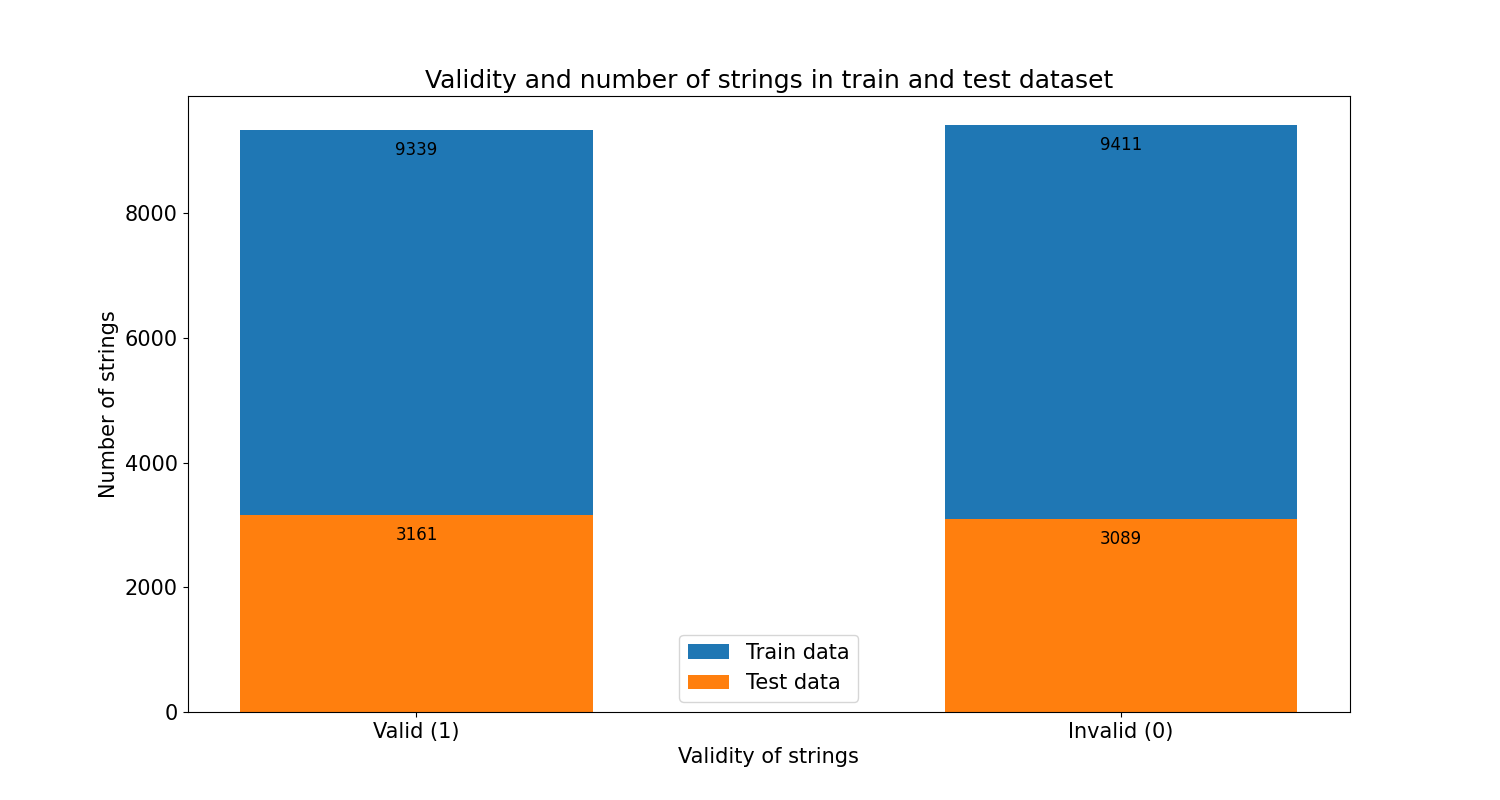}
    \caption[Validity and number of strings in train-test dataset]{Number of valid (true) and invalid (false) Reber sequences in our train-test set}
    \label{fig:string_count}
\end{figure}

Our dataset has $12500$ true Reber sequences and $12500$ false, thus totaling $25000$. Although the true Reber sequences only contain allowed characters (i.e., \textit{B, E, P, S, T, V, X}), the false sequences may contain any character from A to Z.

Out of $25000$ total sequences, our train-test set contains $18750$ and $6250$ sequences, respectively. Furthermore, this train-test individually include the following number of true and false Reber sequences:

\begin{table}[h]
	\centering
	\begin{tabular}{|c|c|c|}
	    \hline
		 & \textbf{Training set} & \textbf{Test set} \\
		\hline
		\textbf{Valid (true)} & 9339 & 3161 \\
		\textbf{Invalid (false)} & 9411 & 3089 \\
		\hline
	\end{tabular}
	\caption[The number of true and false Reber sequences in our train-test set]{The number of true and false Reber sequences in our train-test set.}
	\label{tab:train-test_strings}
\end{table}
Next, we visualize the string length distribution that shows not only the minimum and maximum string length but also the number of strings for each length value.

The following plot displays the string length distribution of our entire dataset:

\begin{figure}[h]
    \centering
    \includegraphics[width=0.9\linewidth]{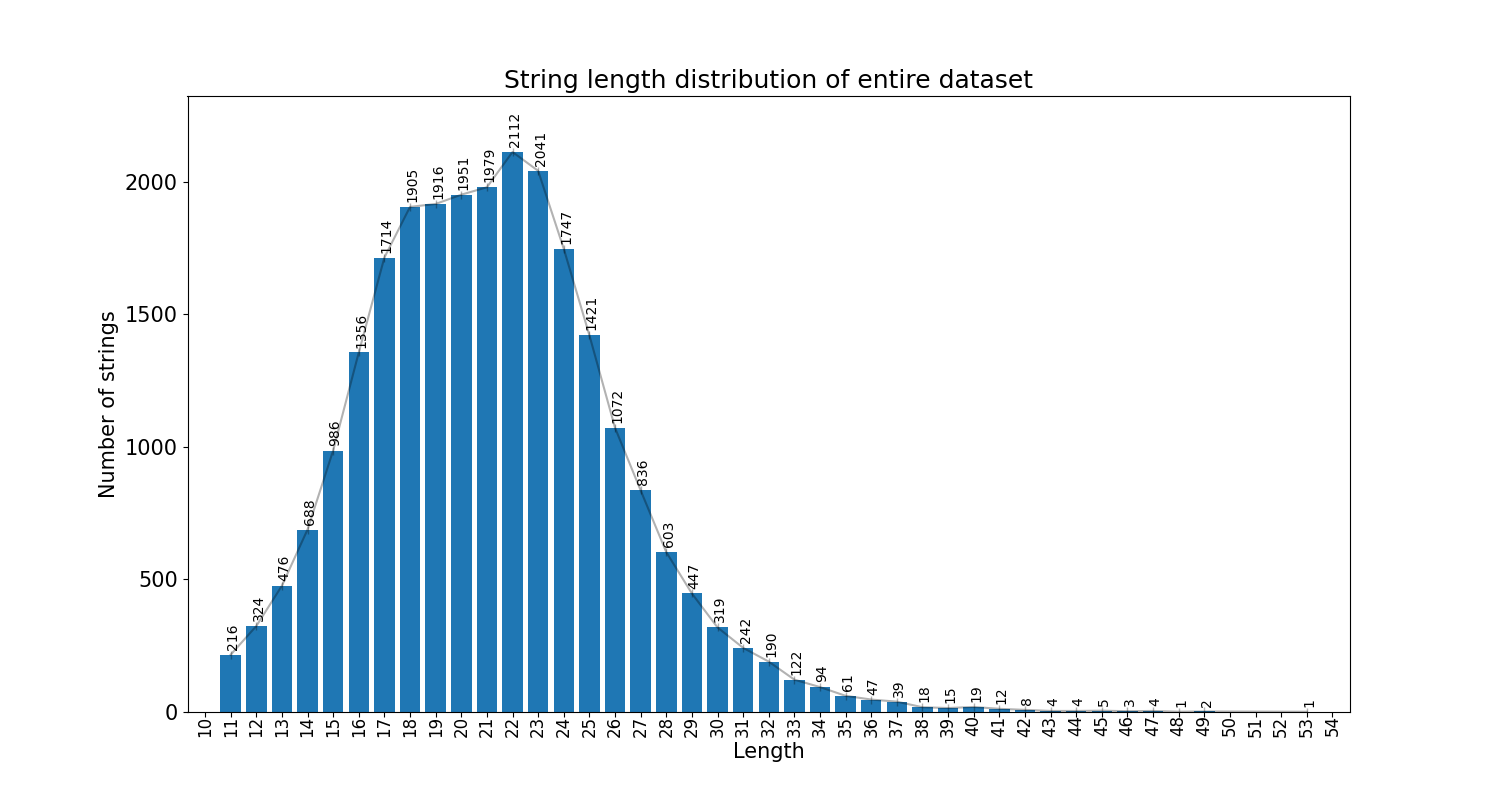}
    \caption[String length distribution of entire dataset]{String length distribution of entire dataset}
    \label{fig:string_len}
\end{figure}

As we can see in the above plot, in our entire dataset, the minimum string length is $11$ with $216$ sequences, and the maximum string length is $53$ with $1$ sequence. The highest number of strings, $2112$ strings, in our entire dataset is of length $22$.

The following plot displays the string length distribution of our train dataset:

\begin{figure}[h]
    \centering
    \includegraphics[width=0.9\linewidth]{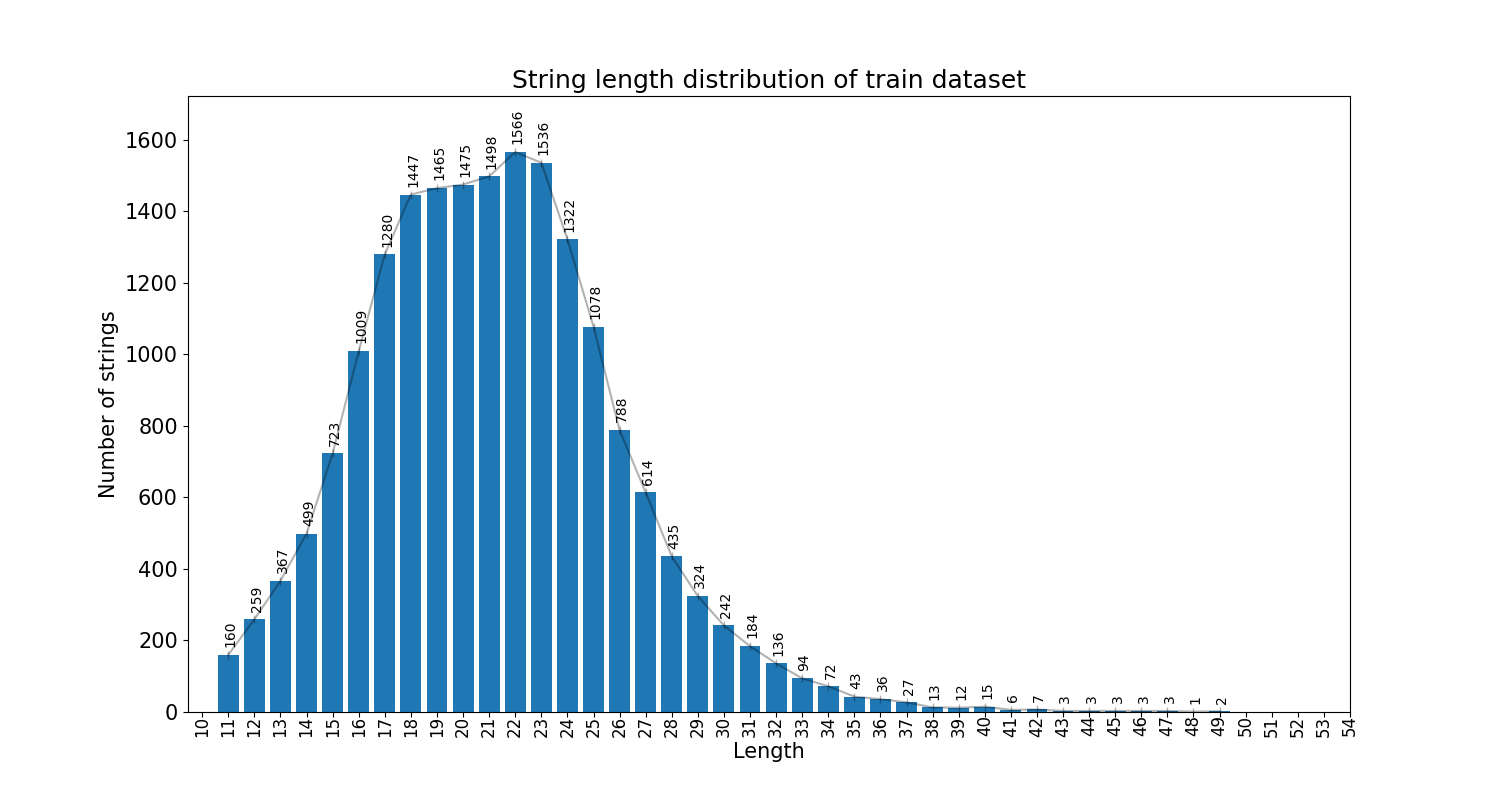}
    \caption[String length distribution of train dataset]{String length distribution of train dataset}
    \label{fig:train_string_len}
\end{figure}

As we can see in the above plot, in our train dataset, the minimum string length is $11$ with $160$ sequences, and the maximum string length is $49$ with $2$ sequence. The highest number of strings, $1566$ strings, in our train dataset is of length $22$.

The following plot displays the string length distribution of our test dataset:

\begin{figure}[h]
    \centering
    \includegraphics[width=0.9\linewidth]{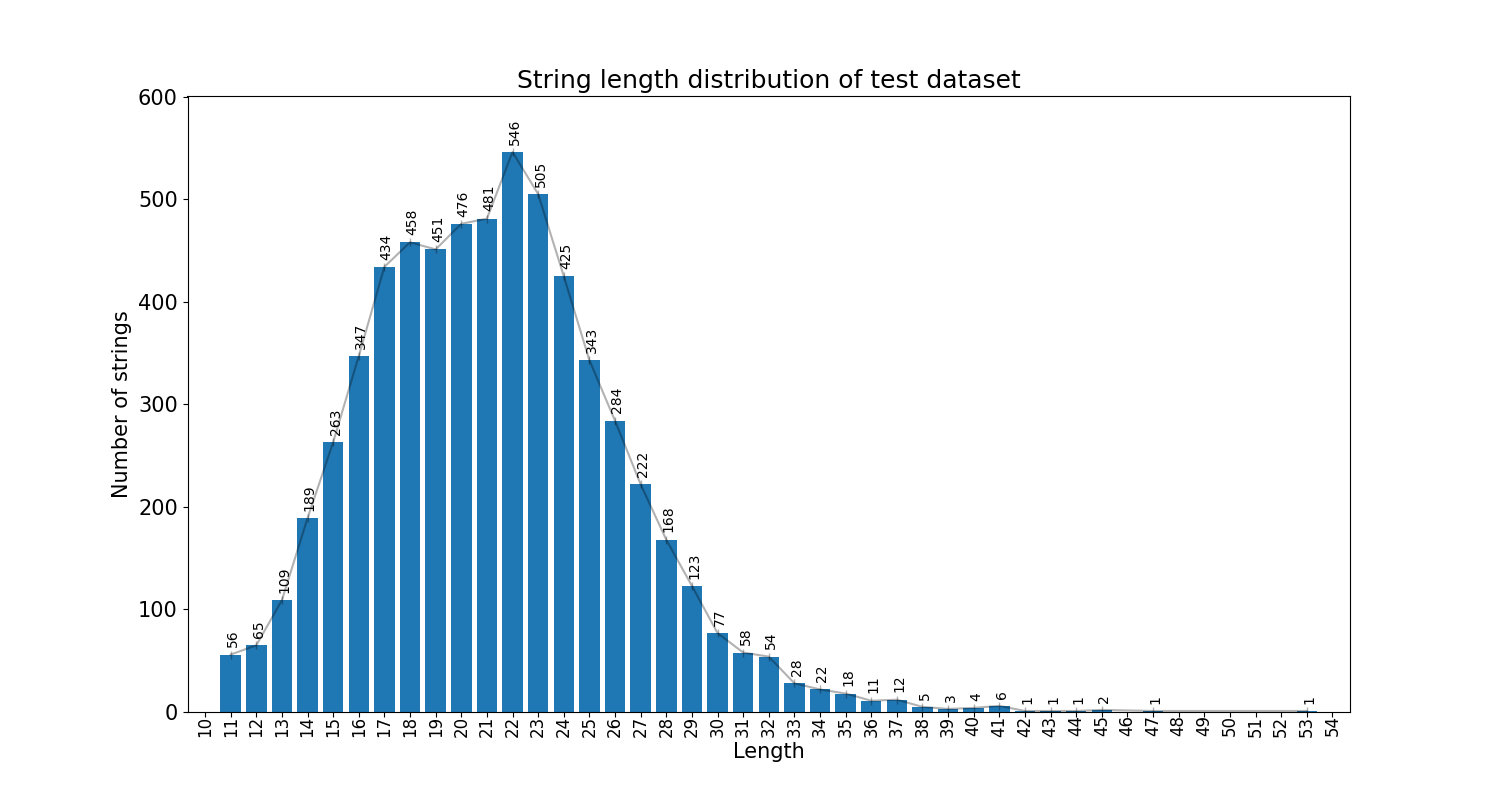}
    \caption[String length distribution of test dataset]{String length distribution of test dataset}
    \label{fig:test_string_len}
\end{figure}

As we can see in the above plot, in our test dataset, the minimum string length is $11$ with $56$ sequences, and the maximum string length is $53$ with $1$ sequence. The highest number of strings, $546$ strings, in our test dataset is of length $22$.
\chapter{Sparse Recurrent Neural Networks}\label{chap:experiments}
This chapter explains the experiments described in section \ref{section:proposed_approach}, but before that, we first describe our base model, against which we will compare the performance of our pruning experiment.


\section{Base model}\label{section:training_rnn}
Apart from the standard input and output layer, our base model is a regular RNN model followed by a linear layer and stacked upon an embedding layer, as shown in the following figure:

\begin{figure}[h]
	\centering
	\includegraphics[width=0.85\linewidth]{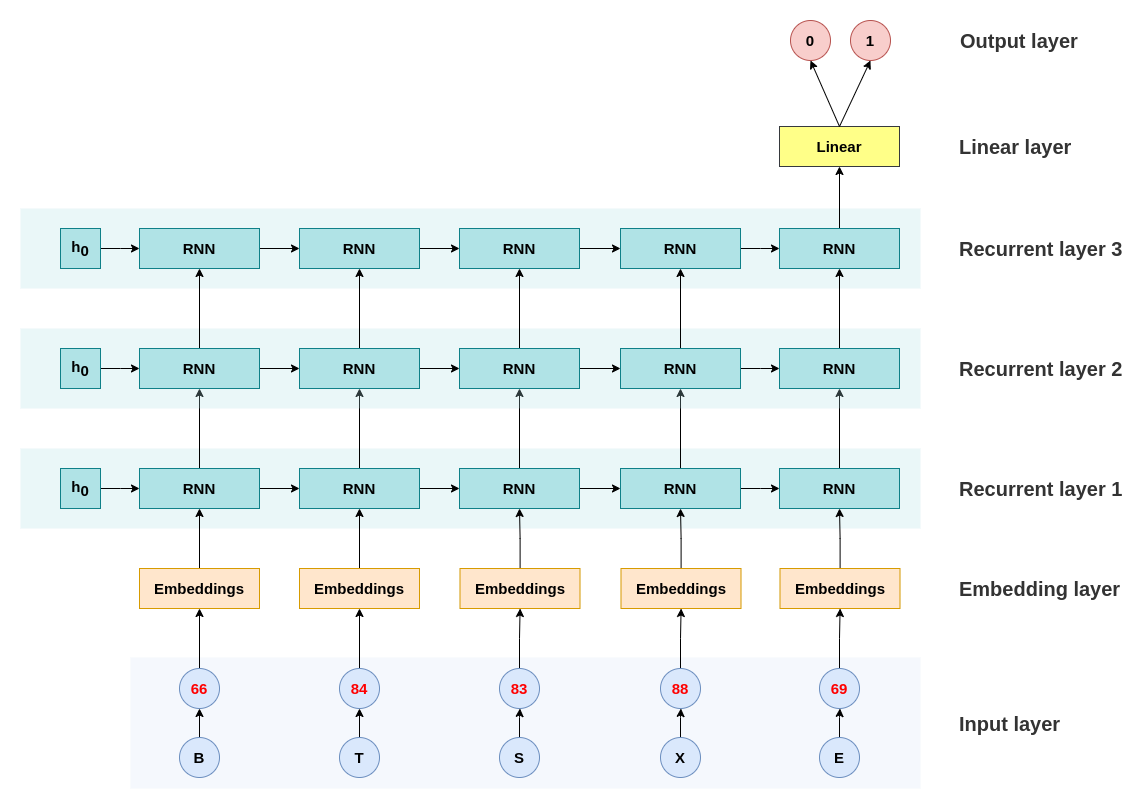}
	\caption[Training recurrent networks]%
	{Visualizing training of a recurrent network with 3 recurrent layers, each layer consisting of 50 neurons.}
	\label{fig:training_rnn}
\end{figure}

Our base model starts with an input layer that feeds data to the model. Since our data is in text format, we convert each character into its corresponding ASCII value. This conversion is visualized in figure \ref{fig:training_rnn} with a valid example Reber sequence, \textbf{BTSXE}. Furthermore, we are working with batches, so all the sequences in a batch must be of equal length. Therefore, we apply zero-padding to ensure all the sequences in a given batch are of the same length.

\begin{figure}[h]
	\centering
	\includegraphics[width=0.85\linewidth]{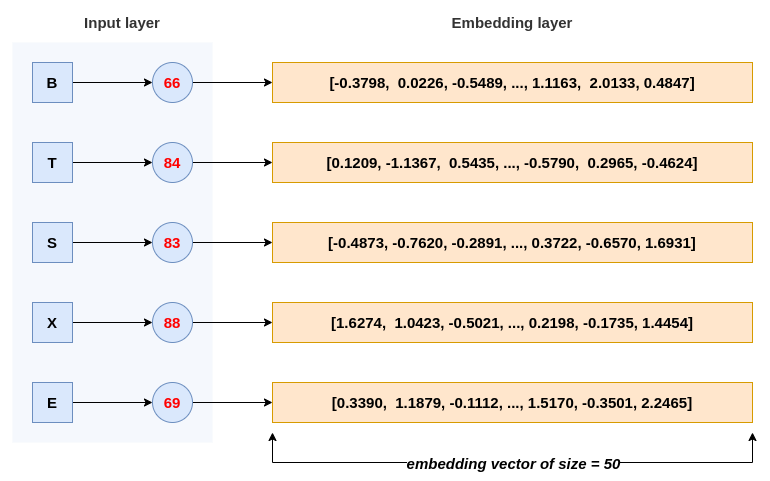}
	\caption[Generating embeddings of example input sequence]%
	{Generating embeddings of example Reber sequence after converting each character of this sequence into its corresponding ASCII value.}
	\label{fig:input-to-embedding}
\end{figure}

After the input layer, the second layer is an embedding layer that transforms each input character into a fixed-length vector as visualized in figure \ref{fig:input-to-embedding}. In our case, this embedding layer is of shape $(128, 50)$, where $128$ is the number of possible ASCII characters, and $50$ is the size of each embedding vector known as embedding dimensions. The value of an embedding dimension must be equal to the number of neurons in the next layer.

After the embedding layer, the subsequent three layers are recurrent, each layer containing $50$ neurons. These layers start with an initial hidden state of shape \textit{(batch size, hidden size)}, comprising all zeros. In LSTM, these layers begin with one initial hidden state and one initial cell state, both of similar shape and comprising of all zeros. Each neuron in recurrent layers is an individual RNN unit with either Tanh or ReLU nonlinearity in case of vanilla recurrent network, LSTM unit in case of LSTM recurrent network and GRU unit in case of GRU recurrent network as visualized in the following figure:

\begin{figure}[h]
	\centering
	\includegraphics[width=1.0\linewidth]{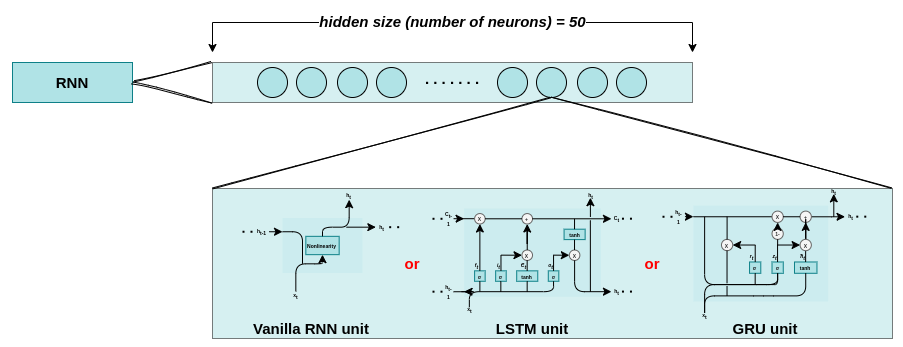}
	\caption[Visualization of a single recurrent layer]%
	{Visualizing a single recurrent layer from our model (fig. \ref{fig:training_rnn}). Each neuron is either an individual RNN unit (fig. \ref{fig:rnn_unit}) or an LSTM unit (fig. \ref{fig:lstm_unit}), or a GRU unit (fig. \ref{fig:gru_unit}).}
	\label{fig:recurrent_layer}
\end{figure}

These recurrent layers are followed by a linear layer that takes input from the third recurrent layer and applies the following linear transformation on the input:
\begin{equation}
\label{eq:linear}
    \hat{y} = xW^{T}+b
\end{equation}
where $\hat{y}$ is the output, $W$ represents weight matrices of shape $(output\_size=2, input\_size=50)$, $x$ is the input, and $b$ is bias of shape $(output\_size=2)$.

\begin{figure}[h]
	\centering
	\includegraphics[width=0.3\linewidth]{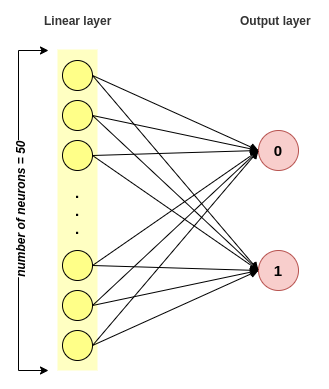}
	\caption[Visualization of linear layer's connection to output layer]%
	{\textbf{Visualizing of linear layer's connection to output layer.} Each of the 50 neurons from the linear layer is densely connected to both the output layer neurons.}
	\label{fig:output_layer}
\end{figure}

Since the third recurrent layer's hidden size is $50$, the linear layer's input is of size $50$, and output is of size two because our target has only two classes, $\{0, 1\}$, as shown in figure \ref{fig:output_layer}.

For our example Reber sequence, the final output is in the form of a float array of size two, for instance, $[-0.8169, 1.3229]$. To transform it into $0$ or $1$, take the index of maximum value from this array. After getting an output, the model calculates the cross-entropy loss and optimizes it via backpropagation through time to a minimum value to achieve better performance.

We train this model on our training dataset of $18750$ sequences for $50$ epochs, and to check its performance, we evaluate it on the test dataset of $6250$ sequences. This training and evaluation employ the hyper-parameters shown in the following table.

\begin{table}[h]
	\centering
	\begin{tabular}{|c|c|}
	    \hline
		\textbf{Hyper-parameter} & \textbf{Value} \\
		\hline
		Learning rate & 0.001 \\
		Loss & Cross-Entropy Loss \\
		Optimizer & Adam \\
		Batch size & 32 \\
		\hline
	\end{tabular}
	\caption[Hyper-parameters used for training and evaluating the base model]{Hyper-parameters used for training and evaluating the base model.}
	\label{tab:hype_param_base}
\end{table}

Since we work with four different RNN variants, namely RNN with Tanh nonlinearity, RNN with ReLU nonlinearity, LSTM, and GRU, we also perform this training four times.

As we have already established, training a recurrent network is computationally expensive and time-consuming. Therefore all four models in our experiment are trained on a Tesla K80 GPU.


\newpage
\section{Pruned Recurrent Neural Networks}\label{section:pruned_rnn}

One of the ways to introduce sparsity in recurrent networks is to prune weight below a certain threshold. As shown in figure \ref{fig:rnn}, we have three different types of weights; input-to-hidden weights, hidden-to-hidden weights, and hidden-to-output weights. In our experiment, we individually and simultaneously prune input-to-hidden and hidden-to-hidden weights as envisioned in the following figure:

\begin{figure}[h]
  \begin{subfigure}{0.33\textwidth}
    \includegraphics[width=\linewidth]{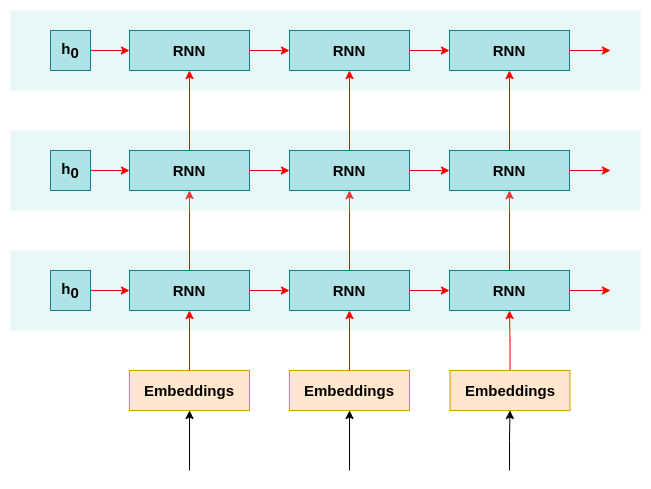}
    \caption{} \label{fig:both}
  \end{subfigure}%
  \begin{subfigure}{0.33\textwidth}
    \includegraphics[width=\linewidth]{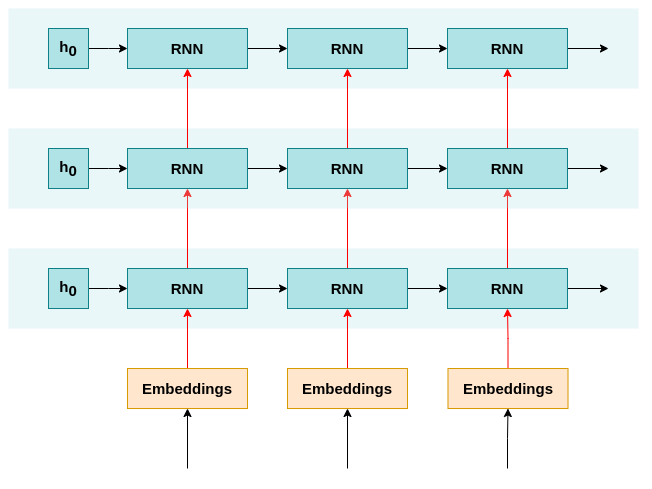}
    \caption{} \label{fig:i2h}
  \end{subfigure}%
  \begin{subfigure}{0.33\textwidth}
    \includegraphics[width=\linewidth]{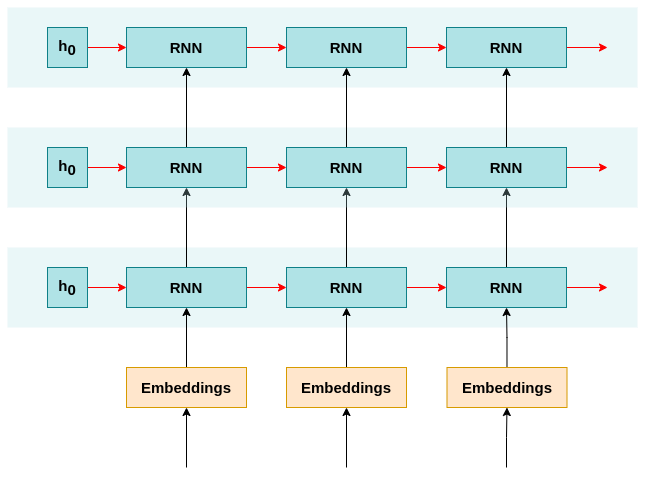}
    \caption{} \label{fig:h2h}
  \end{subfigure}

\caption[Simultaneous and individual pruning of i2h and h2h weights]{Simultaneously and individually pruning input-to-hidden and hidden-to-hidden weights. The red right arrow (\textcolor{red}{$\longrightarrow$}) resembles the pruning of hidden-to-hidden weight, while the red up-arrow (\textcolor{red}{$\big\uparrow$}) resemble the pruning of input-to-hidden weight. Fig. \ref{fig:both} illustrates pruning both, input-to-hidden and hidden-to-hidden weights, fig. \ref{fig:i2h} illustrates pruning only input-to-hidden weights, and fig. \ref{fig:h2h} illustrates pruning only hidden-to-hidden weights.} \label{fig:pruning}
\end{figure}

The threshold based on which we prune weights is calculated based on the percent of weights to prune. Therefore, for example, to prune $p=10\%$ of weights for a given layer, the threshold is the 10th percentile of all the \textbf{absolute} weights in that layer (code \ref{code:threshold}). In our experiment, as shown in figure \ref{fig:flowchart_pruning}, we go from percent $p=10$ to $100$ while incrementing $p$ by $10$ after each round.

Using this threshold value, we create a binary tensor called a mask, where $1$ corresponds to the \textbf{absolute} weight value above the threshold, and $0$ corresponds to the \textbf{absolute} weight value below the threshold (code \ref{code:mask}) in the corresponding weight matrix. Element-wise multiplication of this binary mask with the corresponding weight matrix zero-outs the absolute weight values below the threshold value.

To understand this binary mask generation and mask multiplication, consider the following example weight matrix of shape $(3, 3)$ with float values between $-1$ and $1$:

\[
\begin{bmatrix}
    -0.685 & 0.530 & -0.464 \\
    -0.534 & 0.828 &  0.045 \\
    -0.123 & 0.629 & -0.014
\end{bmatrix}
\]

This weight matrix corresponds to a linear layer with 3 neurons densely connected to the next linear layer with 3 neurons as shown in the below figure:

\begin{figure}[h]
	\centering
	\includegraphics[width=0.3\linewidth]{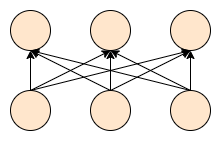}
	\caption[Two densely connected linear layers]%
	{Two densely connected linear layers, each layer containing 3 neurons. This visualization corresponds to the example weight matrix shown above.}
	\label{fig:three_neuron}
\end{figure}

To prune $10\%$ percent of weight values from the above matrix, we calculate the threshold by finding the 10th percentile of this matrix's \textbf{absolute} values, which is $0.0387$. Using this threshold, we create a binary mask of shape $(3, 3)$, where values at positions that correspond to the below threshold \textbf{absolute} values in the weight matrix are kept $0$ while other values are kept $1$ as shown below:

\[
\begin{bmatrix}
    -0.685 & 0.530 & -0.464 \\
    -0.534 & 0.828 &  0.045 \\
    -0.123 & 0.629 & \textcolor{red}{-0.014}
\end{bmatrix}
\longrightarrow
\begin{bmatrix}
    1 & 1 & 1 \\
    1 & 1 &  1 \\
    1 & 1 & \textcolor{red}{0}
\end{bmatrix}
\]

By performing element-wise multiplication of this binary mask with our weight matrix, we get a pruned weight matrix as shown below:

\[
\begin{bmatrix}
    -0.685 & 0.530 & -0.464 \\
    -0.534 & 0.828 &  0.045 \\
    -0.123 & 0.629 & \textcolor{red}{-0.014}
\end{bmatrix}
\circ
\begin{bmatrix}
    1 & 1 & 1 \\
    1 & 1 &  1 \\
    1 & 1 & \textcolor{red}{0}
\end{bmatrix}
=
\begin{bmatrix}
    -0.685 & 0.530 & -0.464 \\
    -0.534 & 0.828 &  0.045 \\
    -0.123 & 0.629 & \textcolor{red}{0}
\end{bmatrix}
\]

The value $0$ in this weight matrix represents the missing connections between the third neuron of the first and second linear layer. After applying the mask, the simple neural network shown in the figure \ref{fig:three_neuron} looks like below:

\begin{figure}[h]
	\centering
	\includegraphics[width=0.5\linewidth]{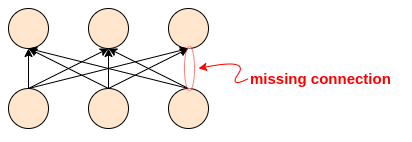}
	\caption[Two densely connected linear layers after pruning]%
	{Two densely connected linear layers containing 3 neurons show the absence of connection between each layer's third neuron after pruning. This visualization corresponds to the pruned weight matrix.}
	\label{fig:pruned_three_neuron}
\end{figure}

We follow a similar mask generation approach and element-wise multiplication for each corresponding weight matrix in each recurrent layer of our trained model. As shown above, this pruning process returns a pruned weight matrix for each corresponding weight matrix. After pruning, we evaluate pruned model's performance on the test dataset to see how such pruning affects an already trained model's overall accuracy.

Once we have the pruned model's performance, the next step is to retrain the same pruned model to estimate the number of epochs required to regain the accuracy. We use the same pruned weight matrices throughout the entire retraining course. This helps us understand how long it takes for a recurrent network to relearn and update weight matrices to make up for the lost connections.

These pruning and retraining steps are repeated three times per RNN variant (i.e., RNN-Tanh, RNN-ReLU, LSTM, and GRU), first time for pruning both, input-to-hidden and hidden-to-hidden weights, the second time for pruning only input-to-hidden weights, and the third time for pruning only hidden-to-hidden weights. 


\newpage
\section{Randomly Structured Recurrent Neural Networks}\label{section:random_rnn}

The pruning method to induce sparsity in a recurrent network works by canceling connections between neurons of two separate layers. While in this section, we explain another method to induce sparsity by generating random structures that are already sparse. This method was introduced for Artificial Neural Network by Stier et al. in \cite{julian}, which we will revise to make it suitable for our goal.

The first step of this experiment is to generate random graphs irrespective of how they might correspond to a neural network. We generate $100$ connected Watts–Strogatz (\ref{subsubsection:wsmodel}) and Barabási–Albert (\ref{subsubsection:bamodel}) graphs using the graph generators provided by the networkx\footnote{NetworkX is an open-source Python library that empowers users to create and manipulate various graph structures.}. Each of these graphs contains nodes between $(10, 51)$, meaning the corresponding neural architecture will have $(10, 51)$ neurons.

The networkx graph generators return undirected graphs. Therefore, the next is to make them directed. Converting a random graph into a Directed Acyclic Graph is vital to make it easier to compute layer indexing.

After converting a random graph into a DAG, the next step is to compute layer indexing of all the nodes in this DAG. This helps us identify which node of this DAG belongs to which layer in the resulting neural network. The recursive algorithm to compute layer indexing of a node (similar to the one used by Stier et al. in \cite{julian}) is as below:

\begin{algorithm}
  \DontPrintSemicolon
  \caption[Compute layer indexing of nodes]%
  {Recursive function to compute layer indexing of nodes}
  \label{alg:layer_index}
  
    \KwIn{Graph $G$, Set of nodes $V$, Set of edges $E$}    
    \KwOut{Layer index mapping for each node $v \in V$}
  
    \SetKwFunction{func}{LayerIndexOf}
    \SetKwProg{Fn}{Def}{:}{}
    
    \Fn{\func{$v$}}{
        $l \gets [-1]$ \tcp*{List $l$, initialized with an element $-1$}
        
        \ForEach{$s \in V | E_{s \rightarrow v}$}{
            $l$.append(LayerIndexOf($s$))
        }
        
        \textbf{return} (max($l$) + 1)
    }
    
    $indexes \gets dict()$ \tcp*{Initializing an empty dictionary, $indexes$}
    \ForEach{$v \in V$}{
        $indexes[v] \gets $ LayerIndexOf($v$)
    }
\end{algorithm}

This recursive function outputs layer indexing of all nodes in our DAG, based on which we generate our final neural architecture, which then we convert into a recurrent network by introducing hidden-to-hidden connections, i.e., recurrent connections.

Once we have our randomly structured recurrent network, the final step is to embed it between an embedding layer and a linear layer. The embedding layer transforms data from the input layer into a fixed-length vector and feeds them to our recurrent layer. The linear layer takes the last recurrent layer's output, applies a linear transformation shown in equation \ref{eq:linear}, and generates final outputs.

\subsection{From Random Graph to recurrent network: An example}

To begin, consider the following undirected graph containing 5 nodes:

\begin{figure}[h]
	\centering
	\includegraphics[width=0.45\linewidth]{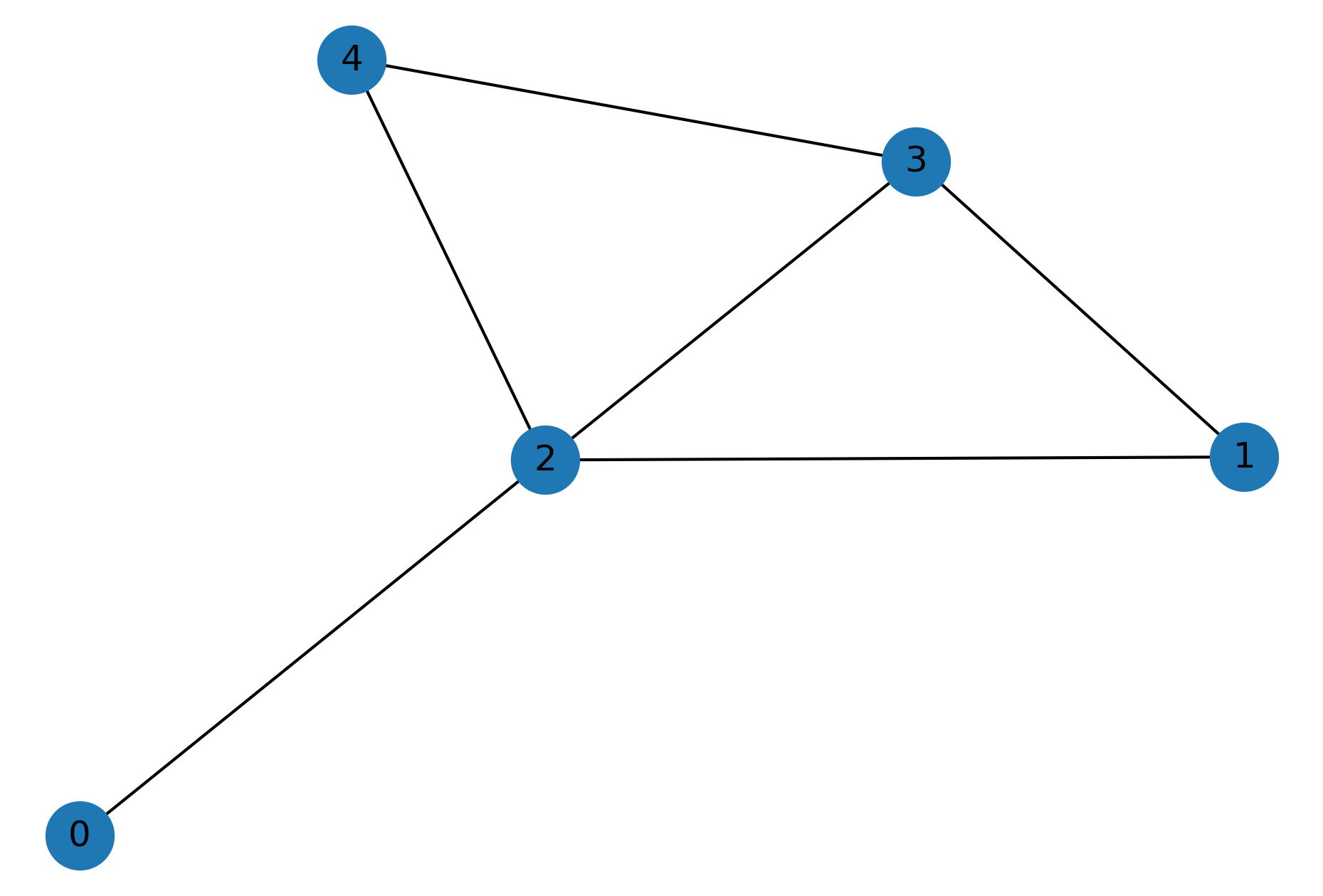}
	\caption[Undirected graph containing $5$ nodes]%
	{An undirected graph containing $5$ nodes.}
	\label{fig:undirected}
\end{figure}

The next step is to convert this graph into a DAG. This conversion introduces directed edges from one node to another, as shown below:

\begin{figure}[h]
	\centering
	\includegraphics[width=0.45\linewidth]{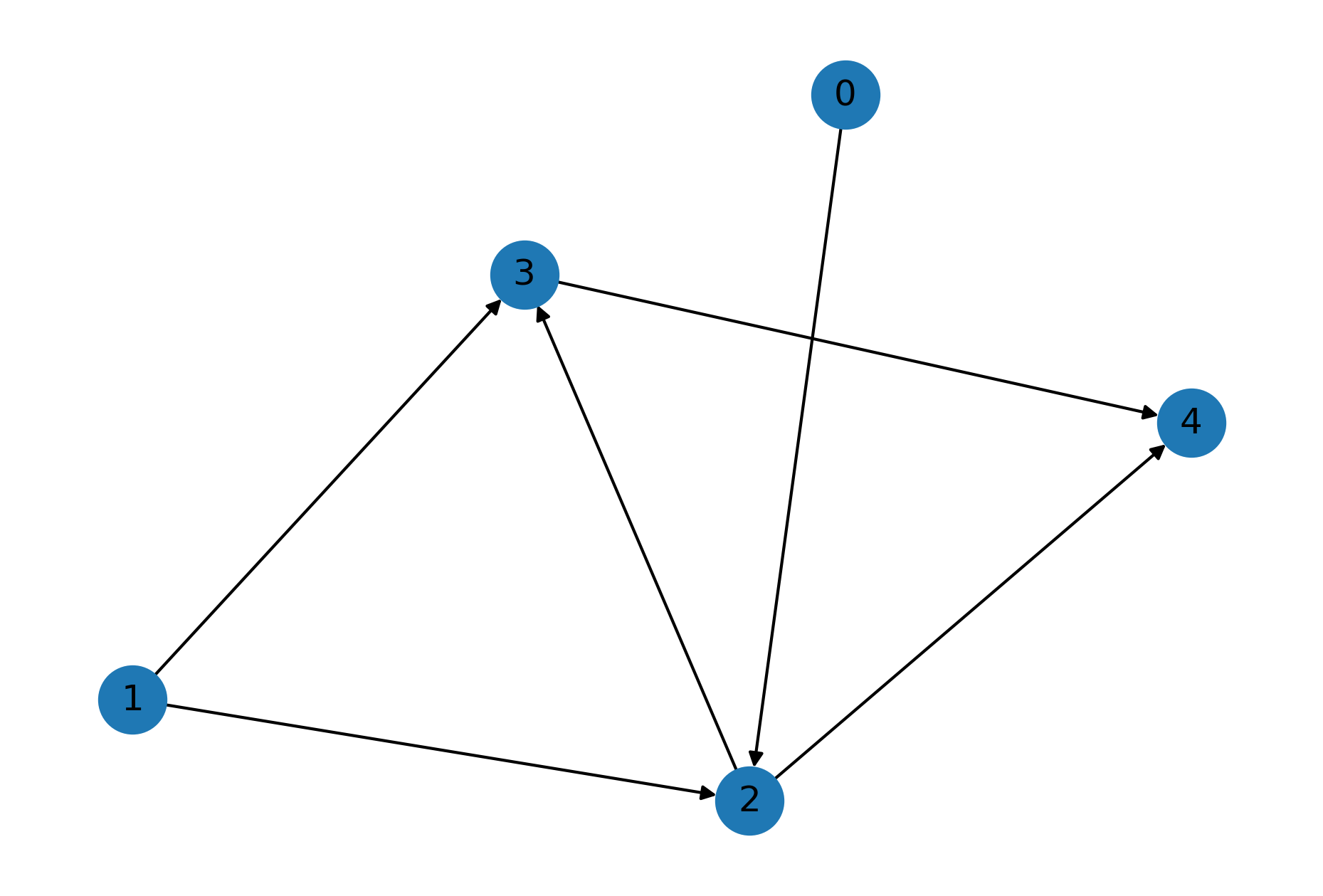}
	\caption[Directed graph containing $5$ nodes]%
	{A directed graph containing $5$ nodes.}
	\label{fig:directed}
\end{figure}

The next step is to compute layer indexing of all $5$ nodes in this DAG using algorithm \ref{alg:layer_index}. The following table shows the values of important variables in this algorithm and the final layer indexing:

\begin{table}[h]
	\centering
	\begin{tabular}{|c|c|c|c|}
	    \hline
		\textbf{Node $v$} & \textbf{Set of $s, \forall s \in V | E_{s \rightarrow v}$} & List $l$ & Layer index of node $v$\\
		\hline
		0 & \{\} & [-1] & 0\\
		1 & \{\} & [-1] & 0\\
		2 & \{0, 1\} & [-1, 0, 0] & 1\\
		3 & \{1, 2\} & [-1, 0, 1] & 2\\
		4 & \{2, 3\} & [-1, 1, 2] & 3\\
		\hline
	\end{tabular}
	\caption[Computing layer indexing of example DAG]{We first identify a set of all the parent nodes for a given node $v$, then append layer indexing of these parent nodes to the list $l$, and finally return $max(l) + 1$ as layer index of node $v$.}
	\label{tab:layer_index}
\end{table}

The next step is to create a neural architecture using this layer indexing:

\begin{figure}[h]
	\centering
	\includegraphics[width=0.25\linewidth]{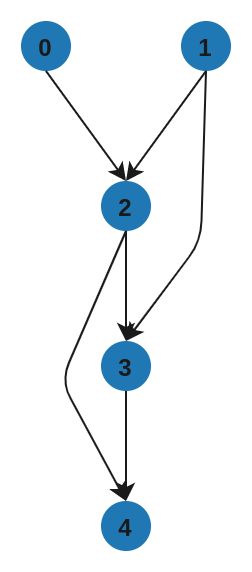}
	\caption[Neural architecture generated using the layer indexing]%
	{A neural architecture of $4$ layers, generated using the layer indexing of nodes shown in table \ref{tab:layer_index}.}
	\label{fig:neural_arc}
\end{figure}

Once we have the neural architecture, as shown in figure \ref{fig:neural_arc}, the next step is to introduce recurrent connections. This will give us our randomly structured recurrent network.

Attaching this randomly structured recurrent network between an embedding layer and a linear layer, as shown in the below figure, gives us our final model:

\begin{figure}[h]
	\centering
	\includegraphics[width=0.8\linewidth]{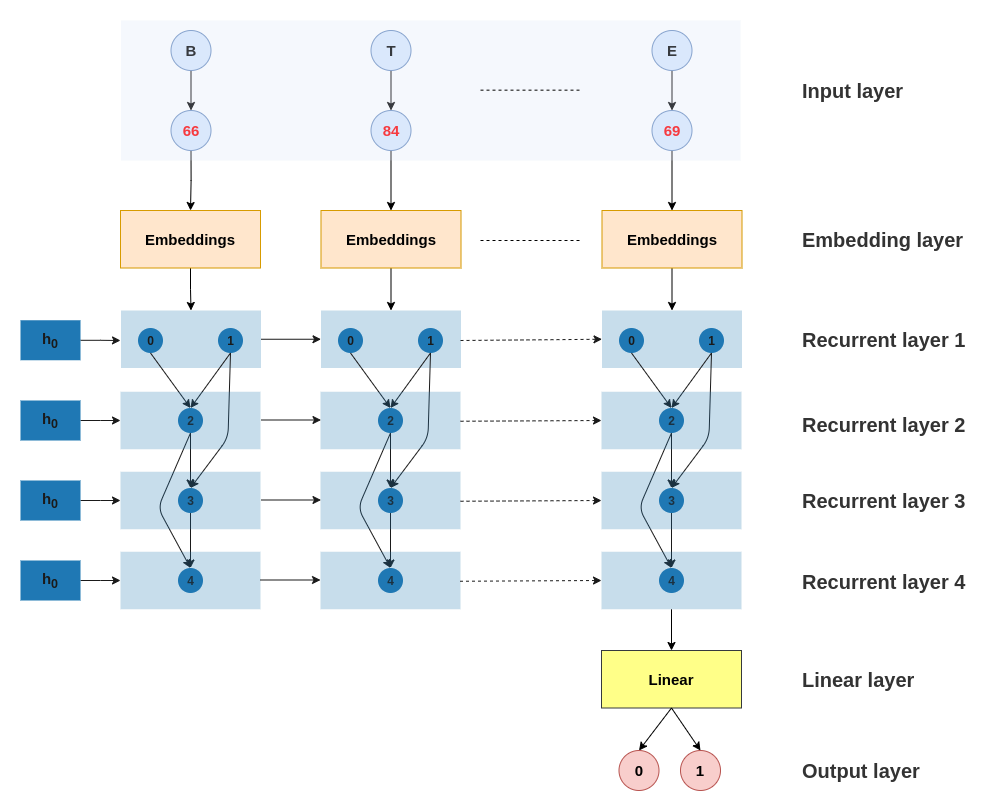}
	\caption[Complete model with recurrent architecture generated from a random graph]%
	{A complete trainable model with an input and output layer, an embedding layer, a linear layer, and a recurrent architecture generated from a random graph.}
	\label{fig:recurrent_arc}
\end{figure}

This is the final trainable model that we train on our training dataset for $15$ epochs and evaluate on the test dataset.

Similarly, we generate $200$ random graphs with nodes between $(10, 51)$ for each RNN variant (i.e., RNN-Tanh, RNN-ReLU, LSTM, and GRU), convert them into a recurrent network, and create the full model for each randomly structured recurrent network, as shown in figure \ref{fig:recurrent_arc}. We train them on our train dataset for $15$ epochs with hyper-parameters shown in table \ref{tab:hype_param_base} except with batch size of $128$. Finally, we evaluate them on our test dataset and store the performance that is useful in the next part of the experiment.

\subsection{Identifying important graph properties}

While training random structure recurrent networks, we also record specific graph properties (mentioned in section \ref{subsection:properties}) of its base random graph. After training and evaluation are complete, we store its accuracy, creating a small dataset of $200$ rows.

Since the graph properties in our small dataset are in different ranges, we apply MinMax scaler on our dataset to scale each graph property between a range $(0, 1)$.

Next, we split this dataset into a train-test with a $0.9/0.1$ ratio. We train three different regressor algorithms: Bayesian ridge regressor, Random Forest regressor, and Ada Boost regressor, on the train set with graph properties as features and its corresponding accuracy as the target.

Then, we evaluate each regressor algorithm on the test set and report an R-squared value to understand how our data fit each regressor model.

This chapter described our experiments with Sparse RNN in detail. After getting familiarized with our experiments, we present our experiments' results and their corresponding visualizations in the next chapter.
\chapter{Experiment results}\label{chap:results}

In this chapter, we present our results from experiments demonstrated previously in chapter \ref{chap:experiments}. We begin with presenting base model performance, which we then compare with pruning results. Afterward, we show results of randomly structured recurrent networks and finally, results explaining the importance of graph properties of a base random graph to its corresponding recurrent network.


\section{Base model performance}\label{section:base_perf}

Our first plot visualizes the base model performance of RNN with Tanh nonlinearity. As we can see, except for a brief period of time initially, our model performs steadily at $>90\%$ accuracy throughout the entire training phase.

\begin{figure}[h]
	\centering
	\includegraphics[width=0.82\linewidth]{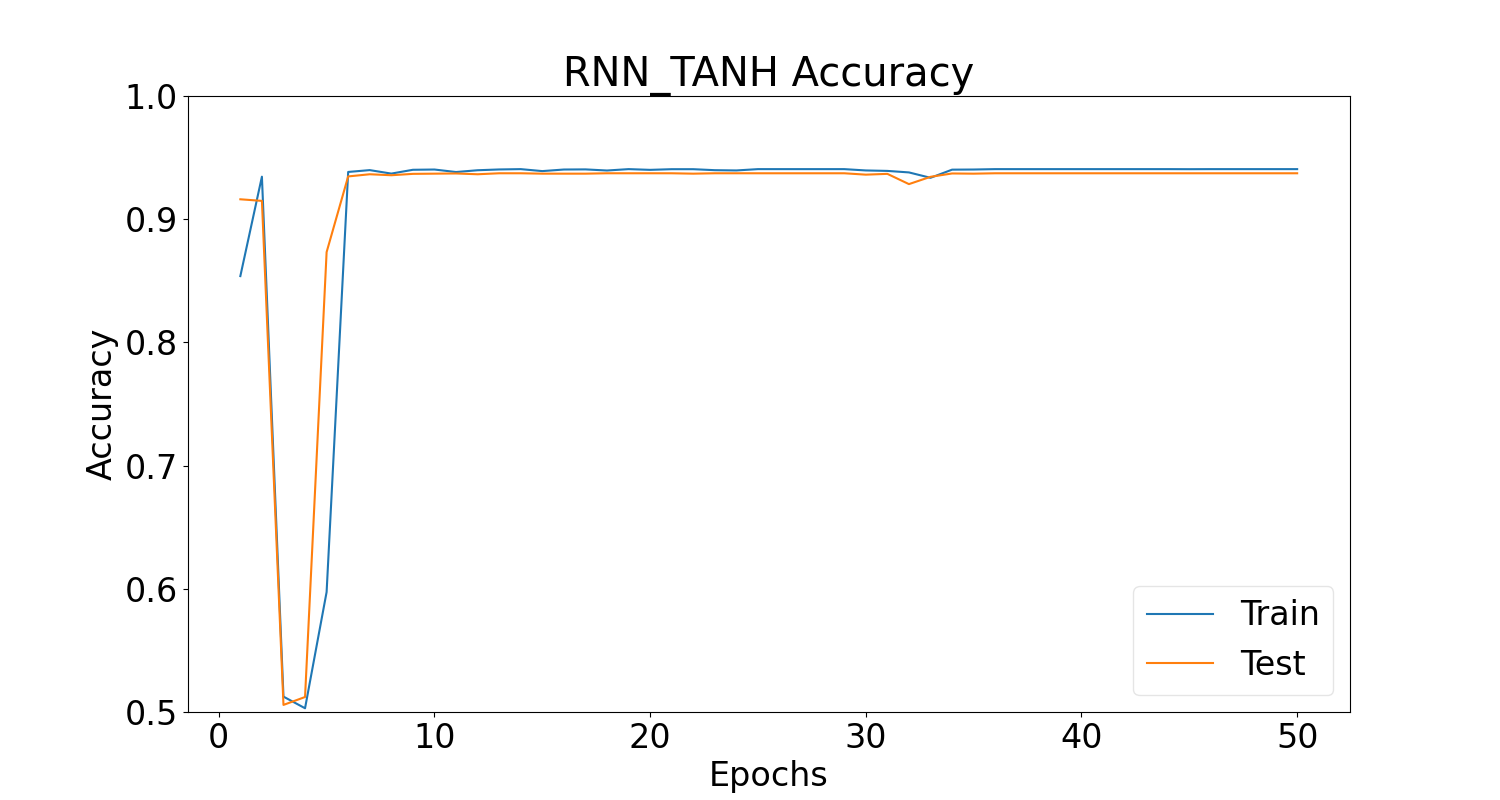}
	\caption[RNN\_Tanh base model performance]%
	{Base model performance of RNN with Tanh nonlinearity. This model is trained for 50 epochs with hyper-parameters shown in table \ref{tab:hype_param_base}.}
	\label{fig:rnn_tanh_bm}
\end{figure}

Next is the base model performance of RNN with ReLU nonlinearity. As we can see in the following graph, our model is consistently above $96\%$ accuracy throughout most of the training and evaluation phase.

\begin{figure}[h]
	\centering
	\includegraphics[width=0.82\linewidth]{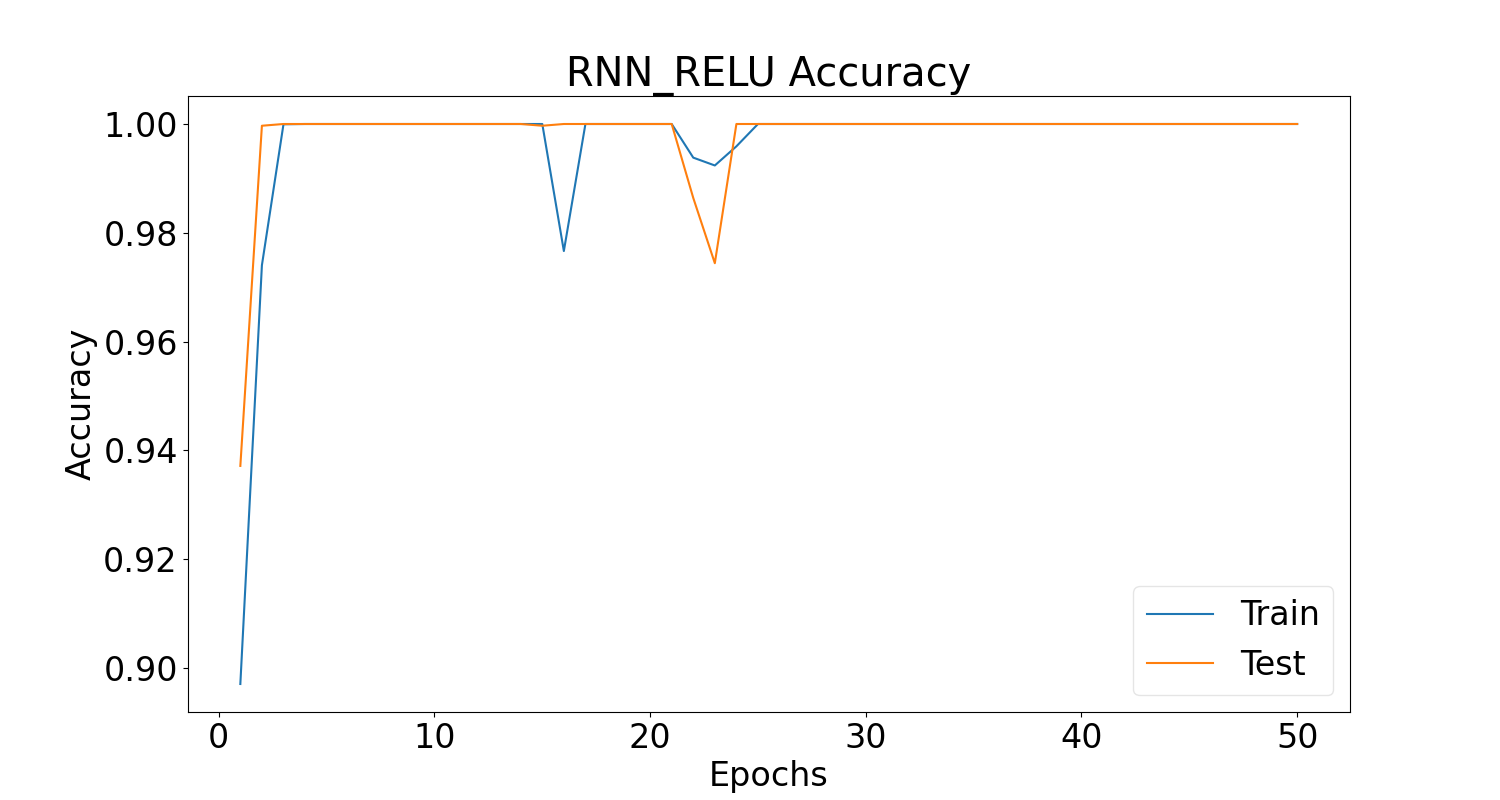}
	\caption[RNN\_ReLU base model performance]%
	{Base model performance of RNN with ReLU nonlinearity. This model is trained for 50 epochs with hyper-parameters shown in table \ref{tab:hype_param_base}.}
	\label{fig:rnn_relu_bm}
\end{figure}

Our base LSTM model's performance is consistent at $100\%$ after beginning at around $93\%$ in the first epoch, as shown in the following figure:

\begin{figure}[h]
	\centering
	\includegraphics[width=0.82\linewidth]{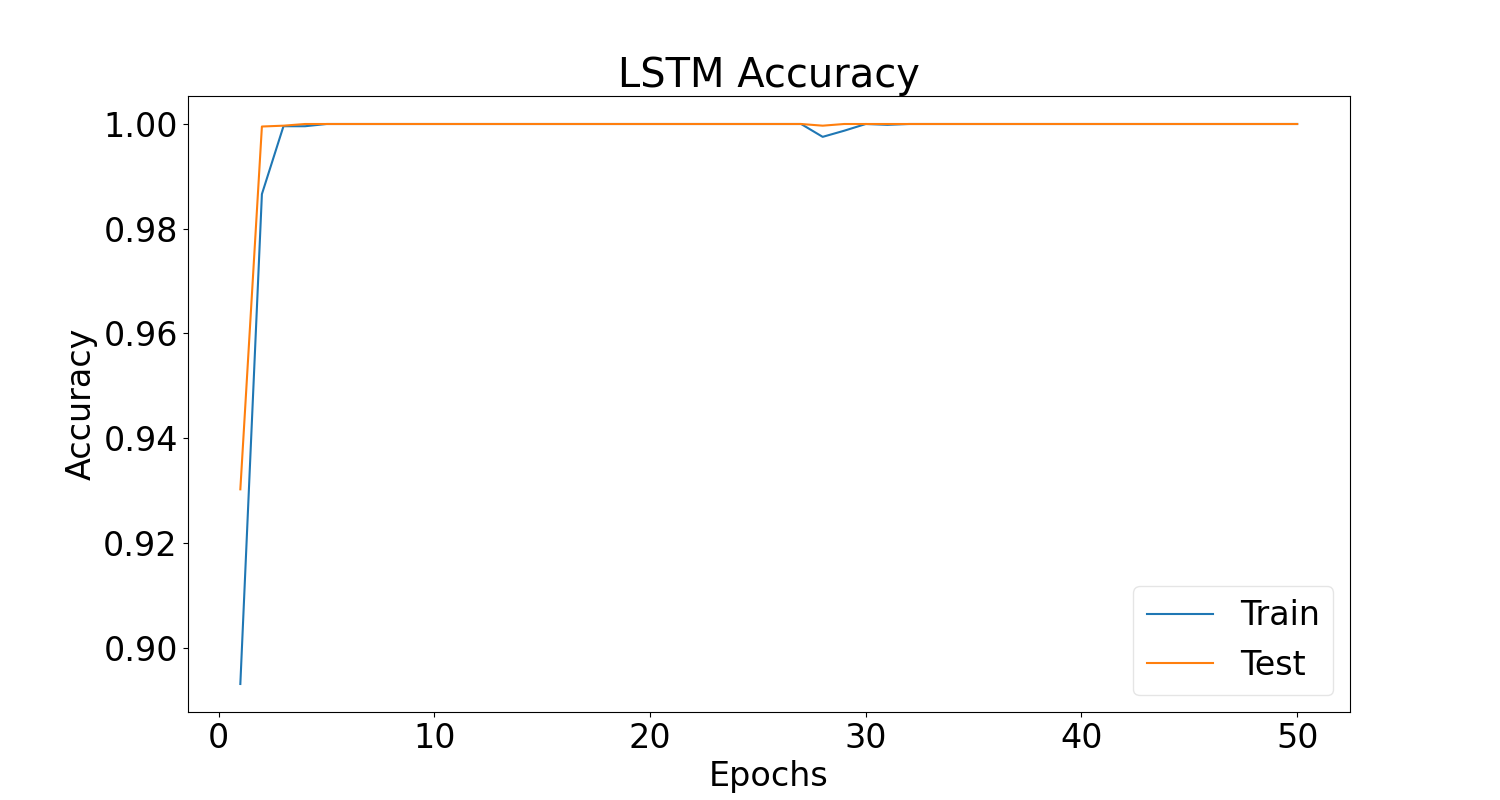}
	\caption[LSTM base model performance]%
	{Base model performance of LSTM. This model is trained for 50 epochs with hyper-parameters shown in table \ref{tab:hype_param_base}.}
	\label{fig:lstm_bm}
\end{figure}

Similar to LSTM, our base GRU model also performs consistently at $100\%$ after beginning at around $93\%$ in the first epoch, as shown in the following figure:

\begin{figure}[h]
	\centering
	\includegraphics[width=0.82\linewidth]{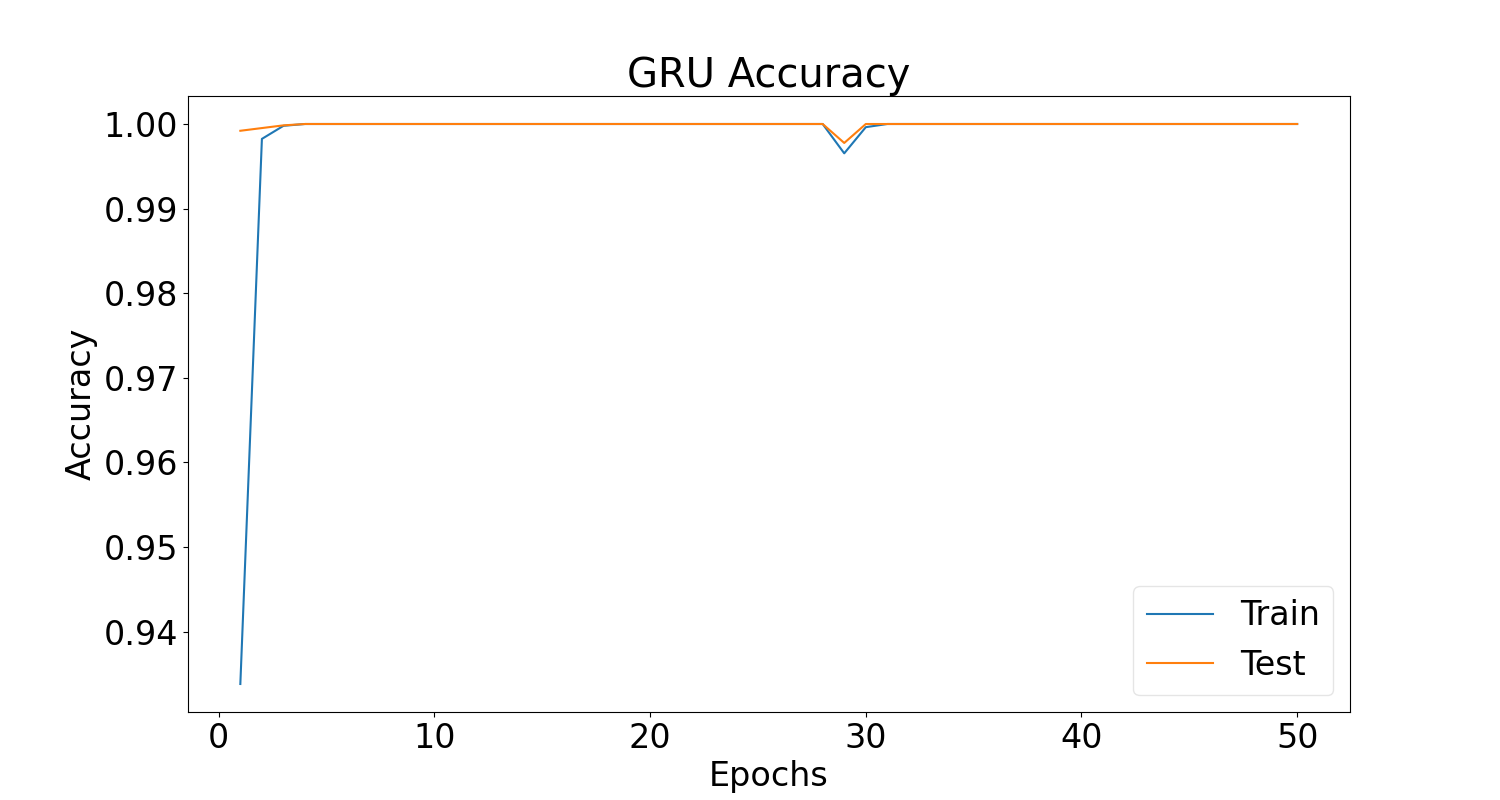}
	\caption[GRU base model performance]%
	{Base model performance of GRU. This model is trained for 50 epochs with hyper-parameters shown in table \ref{tab:hype_param_base}.}
	\label{fig:gru_bm}
\end{figure}

Next, we look at the performance of our base model after pruning it.


\section{Base model performance after pruning}

We begin with displaying the results of pruning both input-to-hidden and hidden-to-hidden weights for each RNN variant, then we proceed with the results of pruning only input-to-hidden weights, and later, the results of pruning only hidden-to-hidden weights.

In each section, we also present the visualizations showing the number of epochs required to regain accuracy after pruning each RNN variant's models.

\subsection[Pruning both, input-to-hidden and hidden-to-hidden weights]{Pruning input-to-hidden and hidden-to-hidden weights simultaneously}

\begin{figure}[h]
	\centering
	\includegraphics[width=0.8\linewidth]{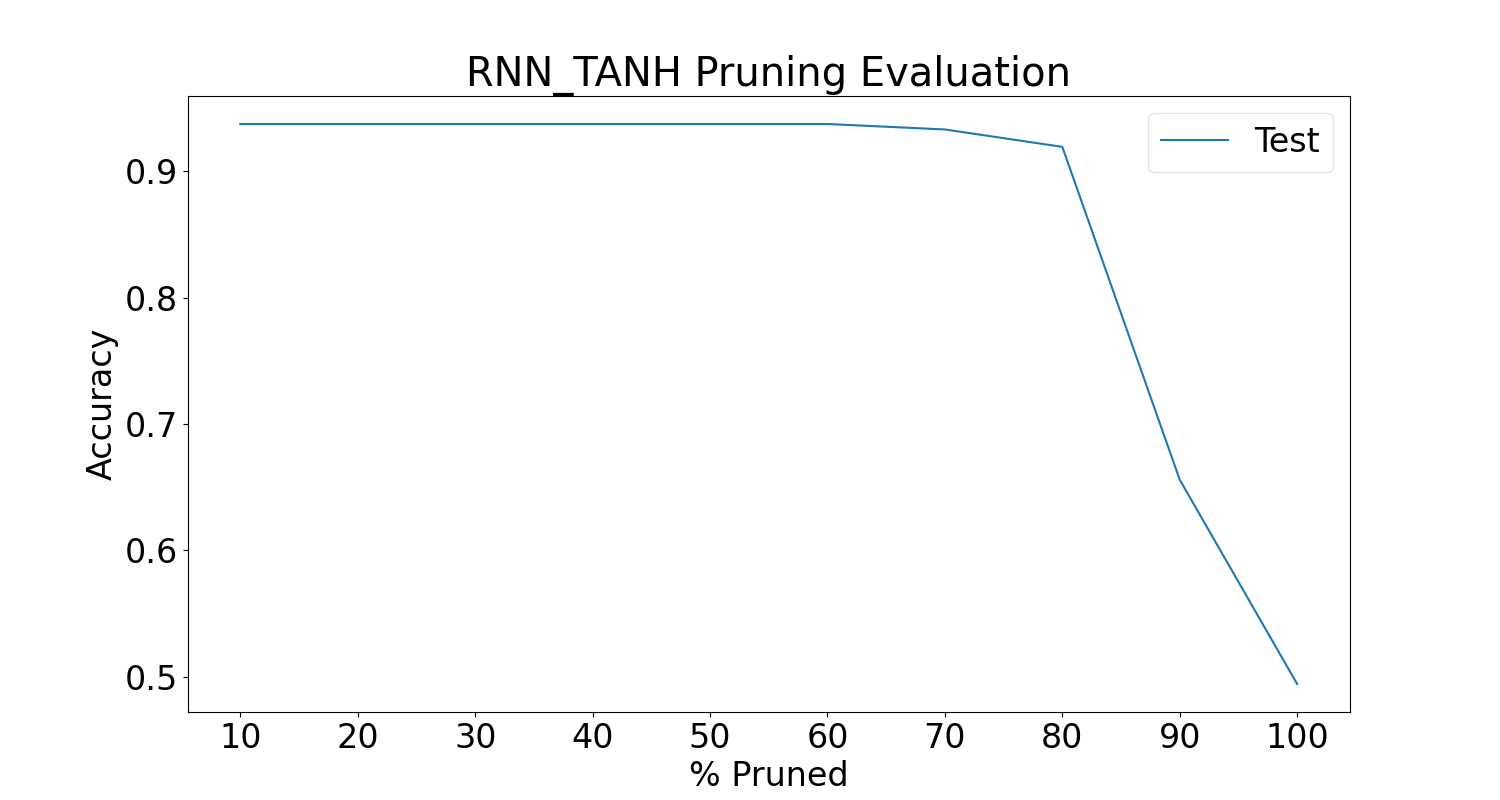}
	\caption[RNN\_Tanh base model performance after pruning]%
	{Base model performance of RNN\_Tanh after pruning. The pruning starts from $10\%$ and ends at $100\%$ with an increment of $10$ after each pruning round.}
	\label{fig:rnn_tanh_prune}
\end{figure}

First, we look at performance of RNN with Tanh nonlinearity after applying $10\%$ to $100\%$ pruning. As we can see in figure \ref{fig:rnn_tanh_prune}, the model performs consistently at above 90\% accuracy at 80\% pruning, meaning the model retains near original performance with 80\% fewer parameters. The model observes a sharp decline in performance at 90\% pruning and goes to baseline accuracy at 100\% pruning.

Next, we observe the number of epochs the model requires to regain the accuracy at 80\%, 90\%, and 100\% pruning:

\begin{figure}[h]
	\centering
	\includegraphics[width=0.8\linewidth]{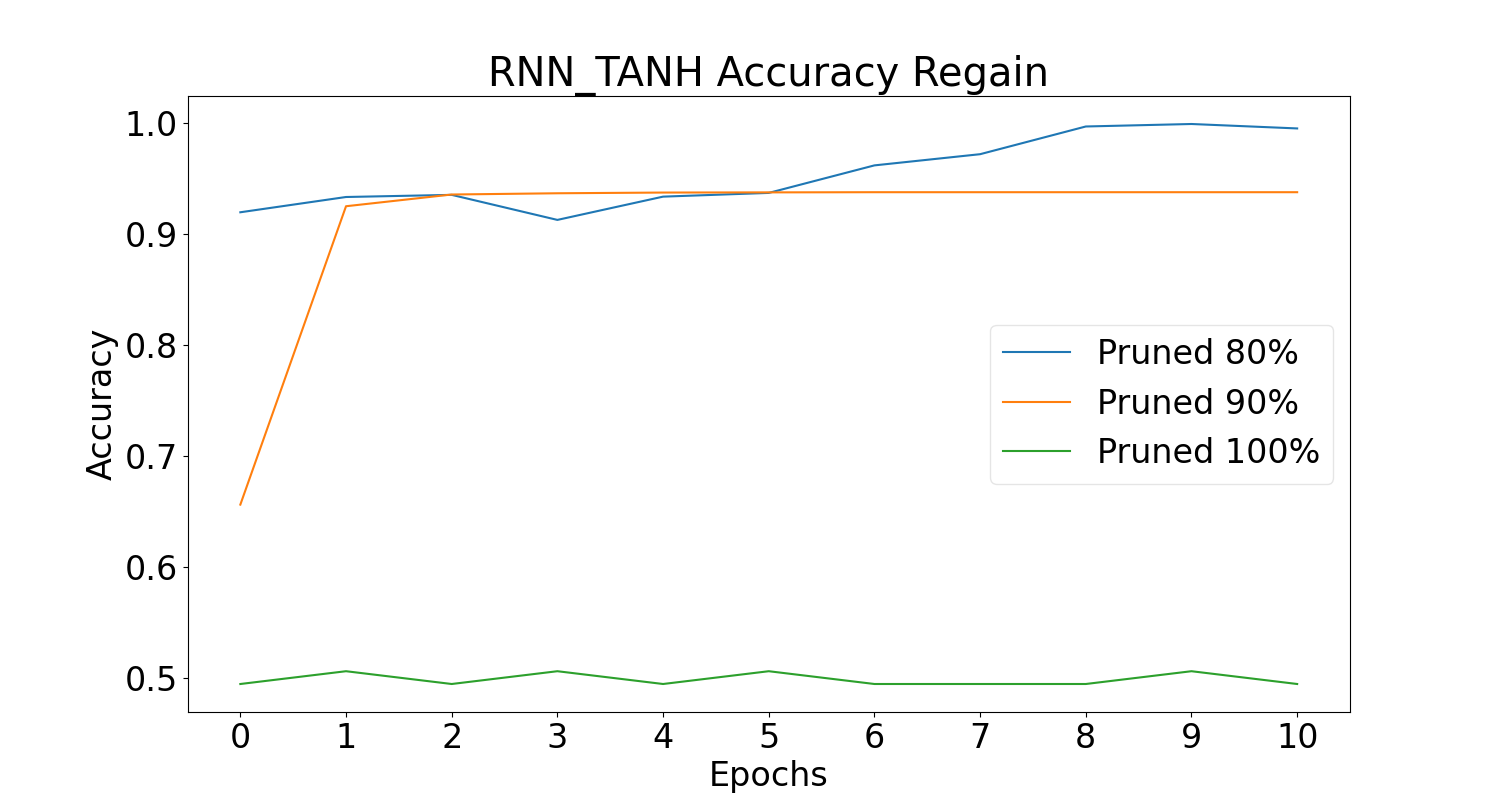}
	\caption[RNN\_Tanh base model performance regain after pruning]%
	{The number of epochs required to regain the accuracy of RNN\_Tanh model after applying 80\%, 90\%, and 100\% pruning.}
	\label{fig:rnn_tanh_prune_regain}
\end{figure}

As we can see in the above figure, after pruning 80\% and 90\% of input-to-hidden and hidden-to-hidden weights, the model regains the original above 90\% performance only after one epoch, while the model never recovers at 100\% pruning.

\begin{figure}[h]
	\centering
	\includegraphics[width=0.8\linewidth]{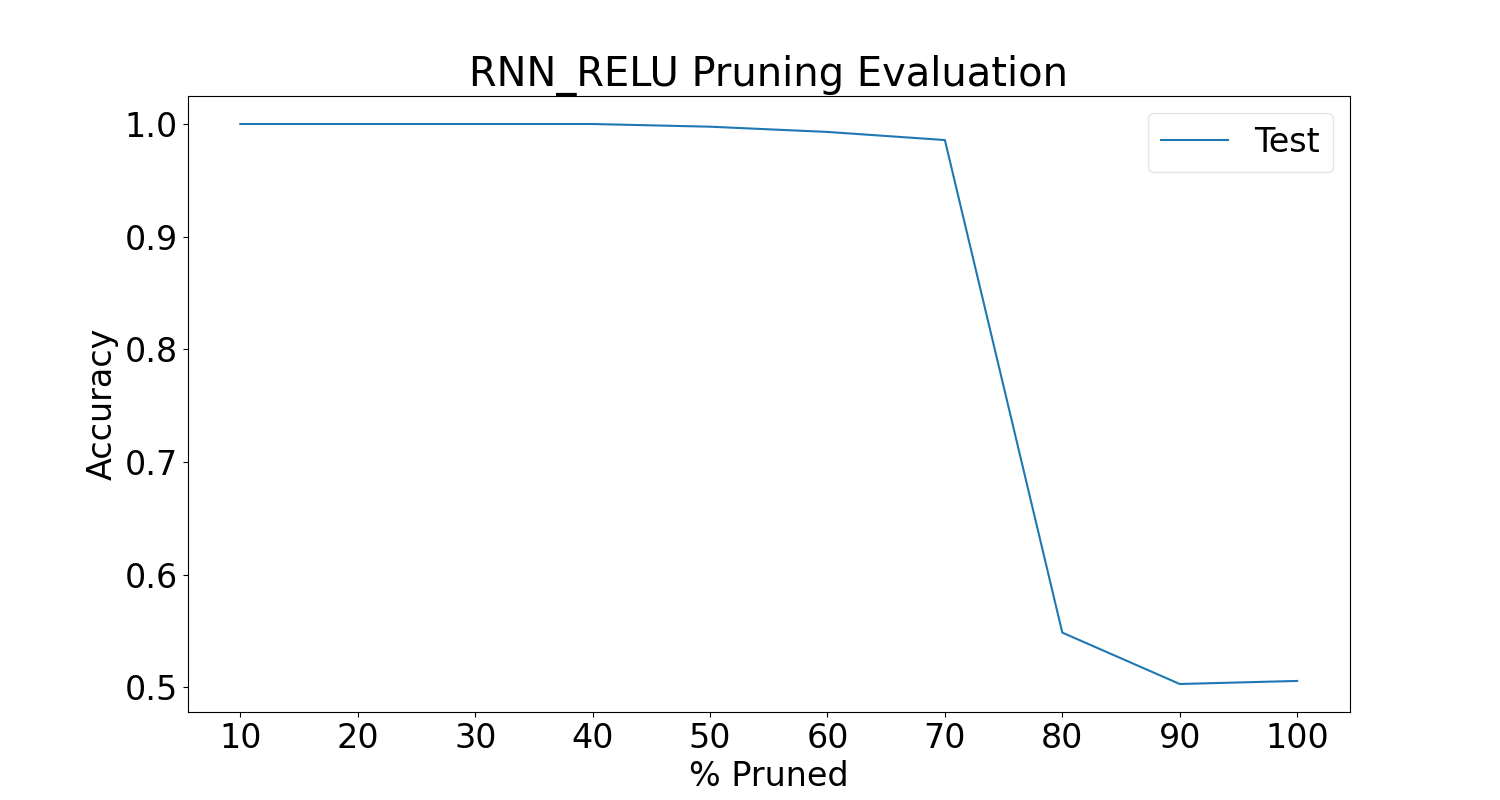}
	\caption[RNN\_ReLU base model performance after pruning]%
	{Base model performance of RNN\_ReLU after pruning. The pruning starts from $10\%$ and ends at $100\%$ with an increment of $10$ after each pruning round.}
	\label{fig:rnn_relu_prune}
\end{figure}

Next, we look at performance of RNN with ReLU nonlinearity after applying $10\%$ to $100\%$ pruning. As we can see in figure \ref{fig:rnn_relu_prune}, the model performs consistently at above 90\% accuracy at 70\% pruning, meaning the model retains near original performance with 70\% fewer parameters. The model observes a sharp decline in performance at 80\% pruning and goes to baseline accuracy at 100\% pruning.

Next, we observe the number of epochs the model requires to regain the accuracy at 80\%, 90\%, and 100\% pruning:

\begin{figure}[h]
	\centering
	\includegraphics[width=0.8\linewidth]{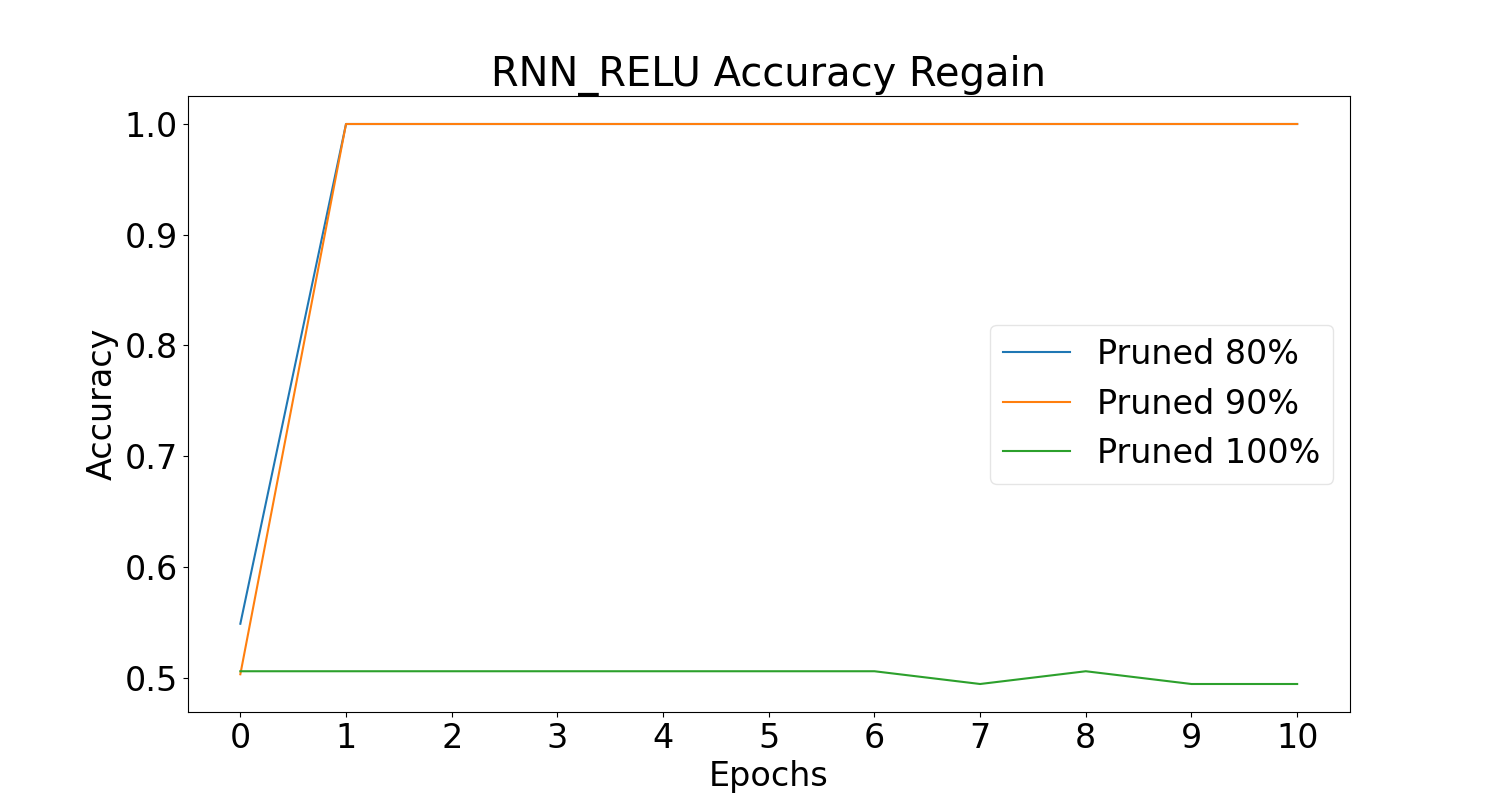}
	\caption[RNN\_ReLU base model performance regain after pruning]%
	{The number of epochs required to regain the accuracy of RNN\_ReLU model after applying 80\%, 90\%, and 100\% pruning.}
	\label{fig:rnn_relu_prune_regain}
\end{figure}

Similar to RNN\_Tanh, after pruning 80\% and 90\% of input-to-hidden and hidden-to-hidden weights, the model regains the original 100\% accuracy only after one epoch, while the model never recovers at 100\% pruning.

\begin{figure}[h]
	\centering
	\includegraphics[width=0.8\linewidth]{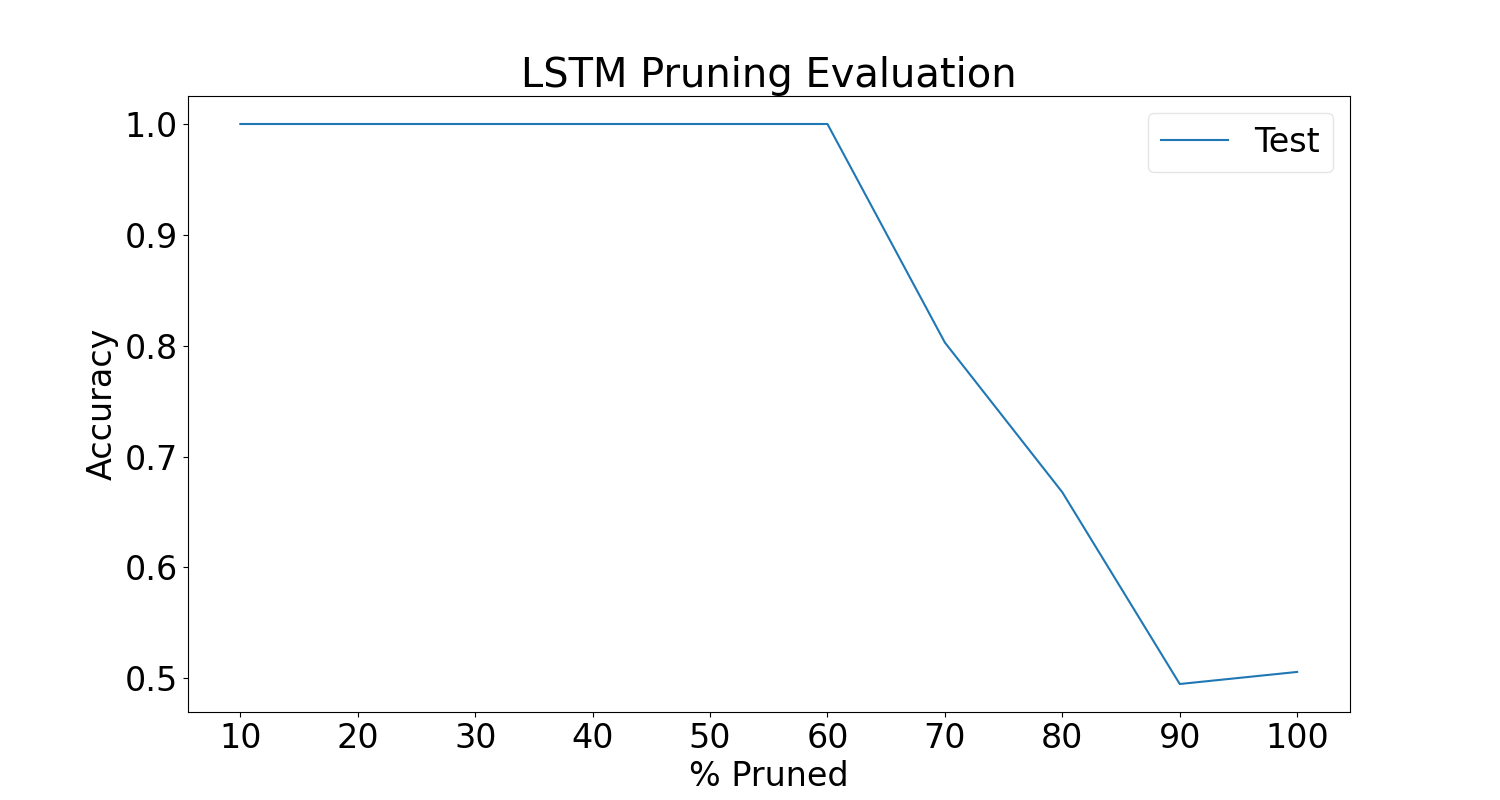}
	\caption[LSTM base model performance after pruning]%
	{Base model performance of LSTM model after pruning. The pruning starts from $10\%$ and ends at $100\%$ with an increment of $10$ after each pruning round.}
	\label{fig:lstm_prune}
\end{figure}

Next, we look at performance of LSTM model after applying $10\%$ to $100\%$ pruning. As we can see in figure \ref{fig:lstm_prune}, the model performs consistently at 100\% accuracy at 60\% pruning. The model observes a sharp decline in performance at 70\% pruning and goes to baseline accuracy at 90\% pruning.

Next, we observe the number of epochs the model requires to regain the accuracy at 70\%, 80\%, 90\%, and 100\% pruning:

\begin{figure}[h]
	\centering
	\includegraphics[width=0.8\linewidth]{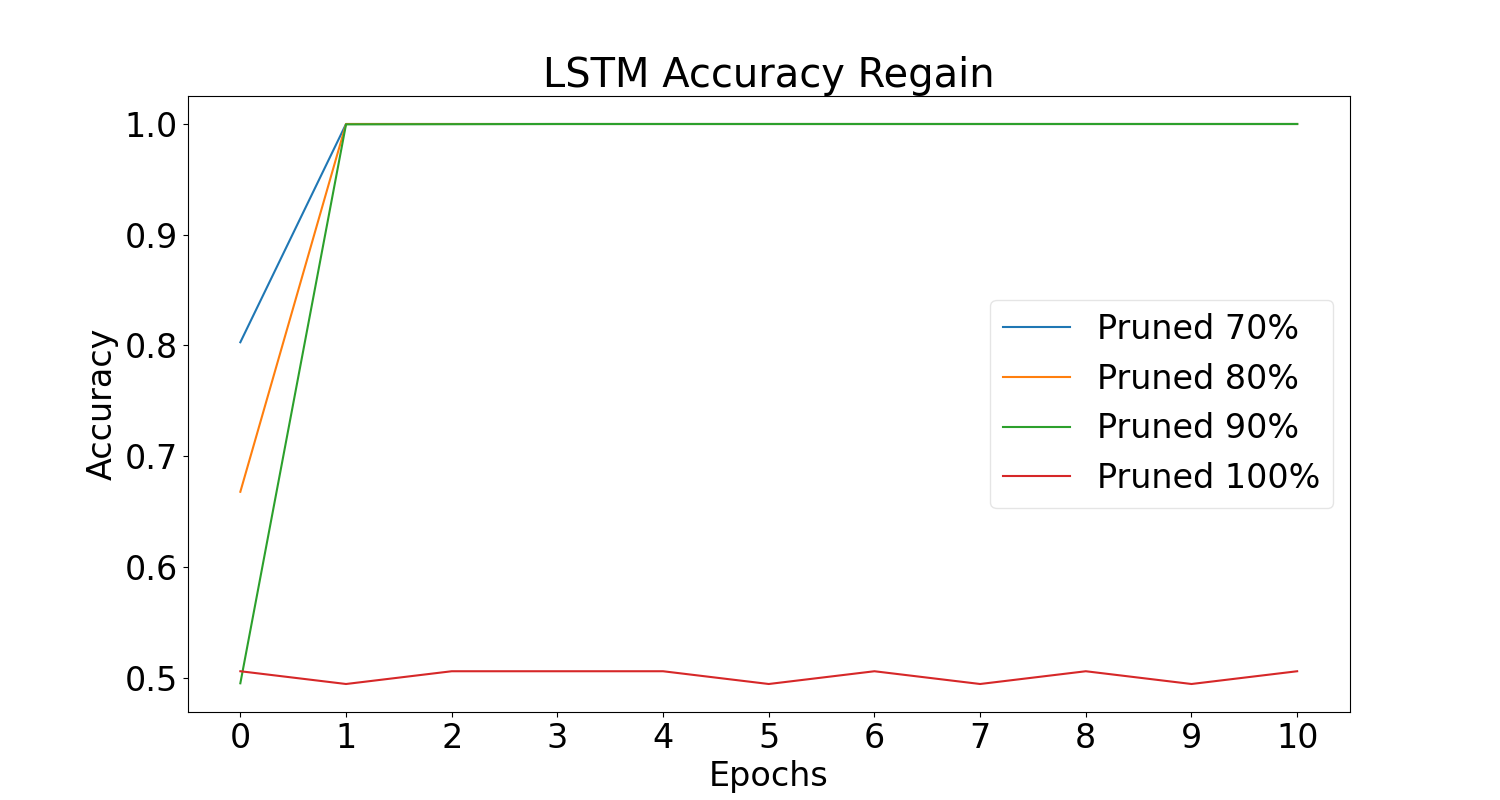}
	\caption[LSTM base model performance regain after pruning]%
	{The number of epochs required to regain the accuracy of LSTM model after applying 70\%, 80\%, 90\%, and 100\% pruning.}
	\label{fig:lstm_prune_regain}
\end{figure}

Similar to RNN\_Tanh and RNN\_ReLU, after pruning 70\%, 80\% and 90\% of input-to-hidden and hidden-to-hidden weights, the model regains the original 100\% accuracy only after one epoch, while the model never recovers at 100\% pruning.

\begin{figure}[h]
	\centering
	\includegraphics[width=0.8\linewidth]{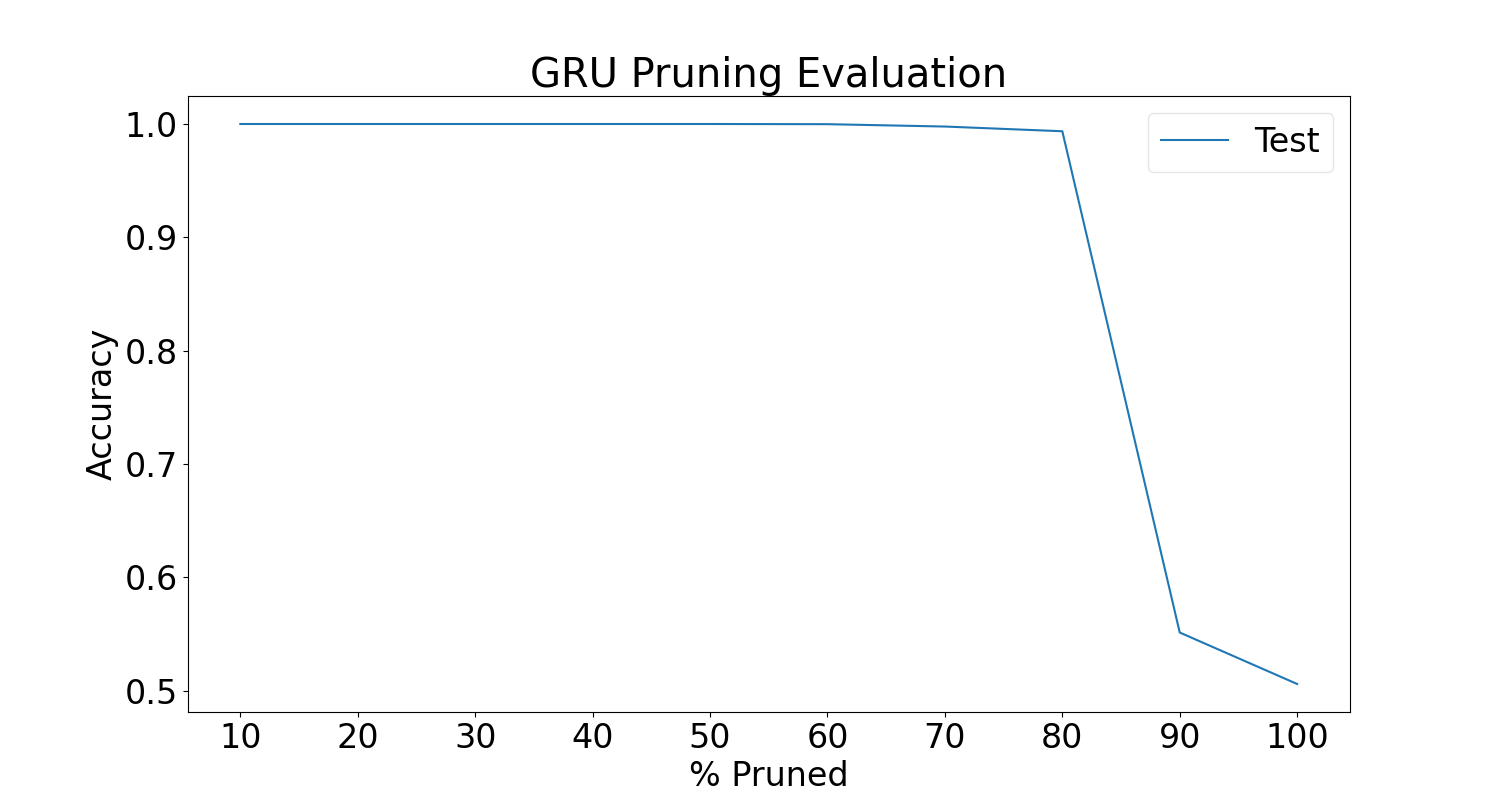}
	\caption[GRU base model performance after pruning]%
	{Base model performance of GRU model after pruning. The pruning starts from $10\%$ and ends at $100\%$ with an increment of $10$ after each pruning round.}
	\label{fig:gru_prune}
\end{figure}

Finally, we look at performance of GRU model after applying $10\%$ to $100\%$ pruning. As we can see in figure \ref{fig:gru_prune}, the model performs consistently at near 100\% accuracy at 80\% pruning. The model observes a sharp decline in performance at 90\% pruning and goes to baseline accuracy at 100\% pruning.

Next, we observe the number of epochs the model requires to regain the accuracy at 90\%, and 100\% pruning:

\begin{figure}[h]
	\centering
	\includegraphics[width=0.8\linewidth]{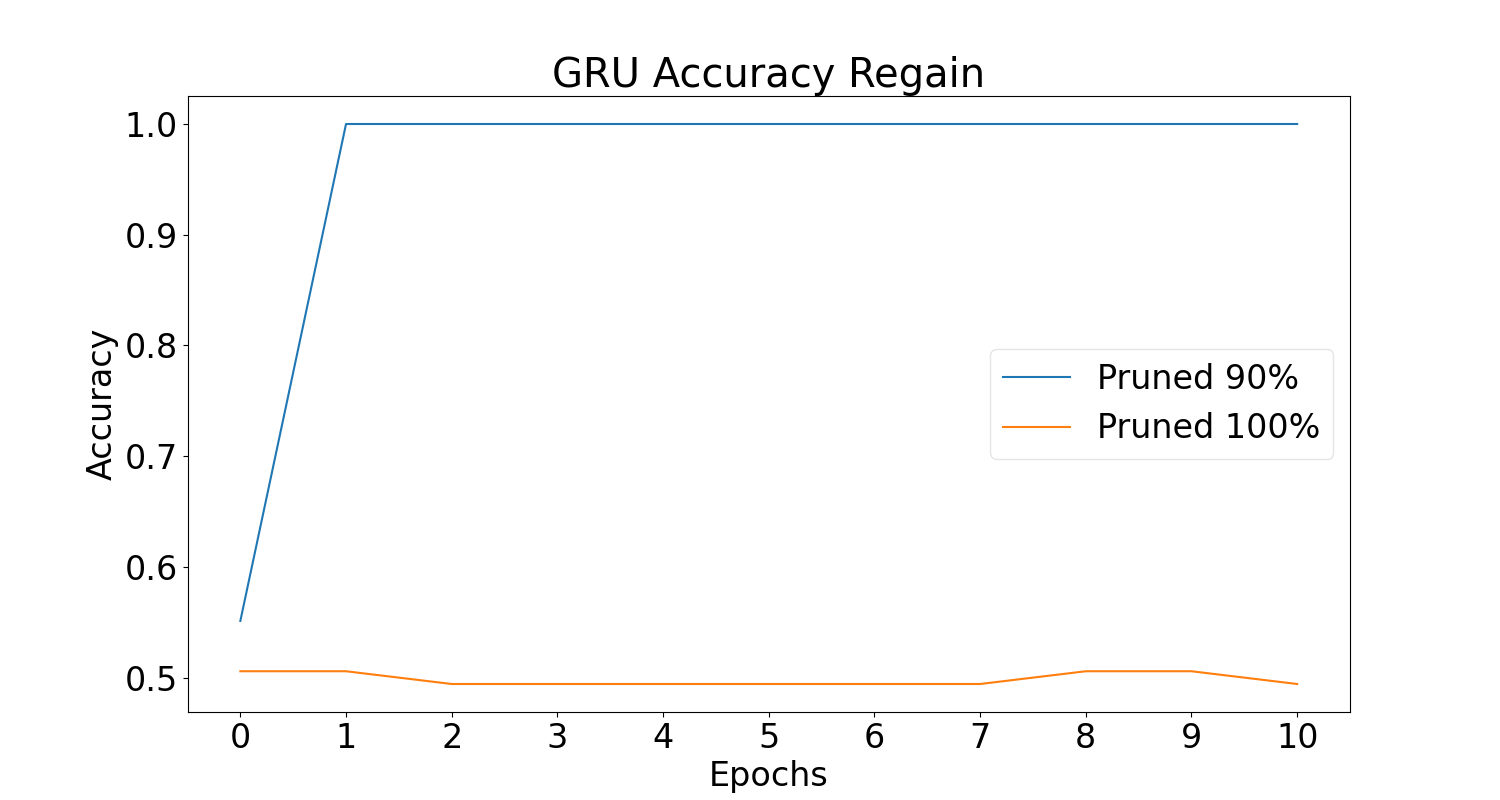}
	\caption[GRU base model performance regain after pruning]%
	{The number of epochs required to regain the accuracy of GRU model after applying 90\%, and 100\% pruning.}
	\label{fig:gru_prune_regain}
\end{figure}

Similar to RNN\_Tanh, RNN\_ReLU, and LSTM, after pruning 90\% of input-to-hidden and hidden-to-hidden weights, the model regains the original 100\% accuracy only after one epoch, while the model never recovers at 100\% pruning.

\subsection{Pruning only input-to-hidden weights}

In this section, we present the results of pruning only input-to-hidden weights in our base model.

\begin{figure}[h]
	\centering
	\includegraphics[width=0.8\linewidth]{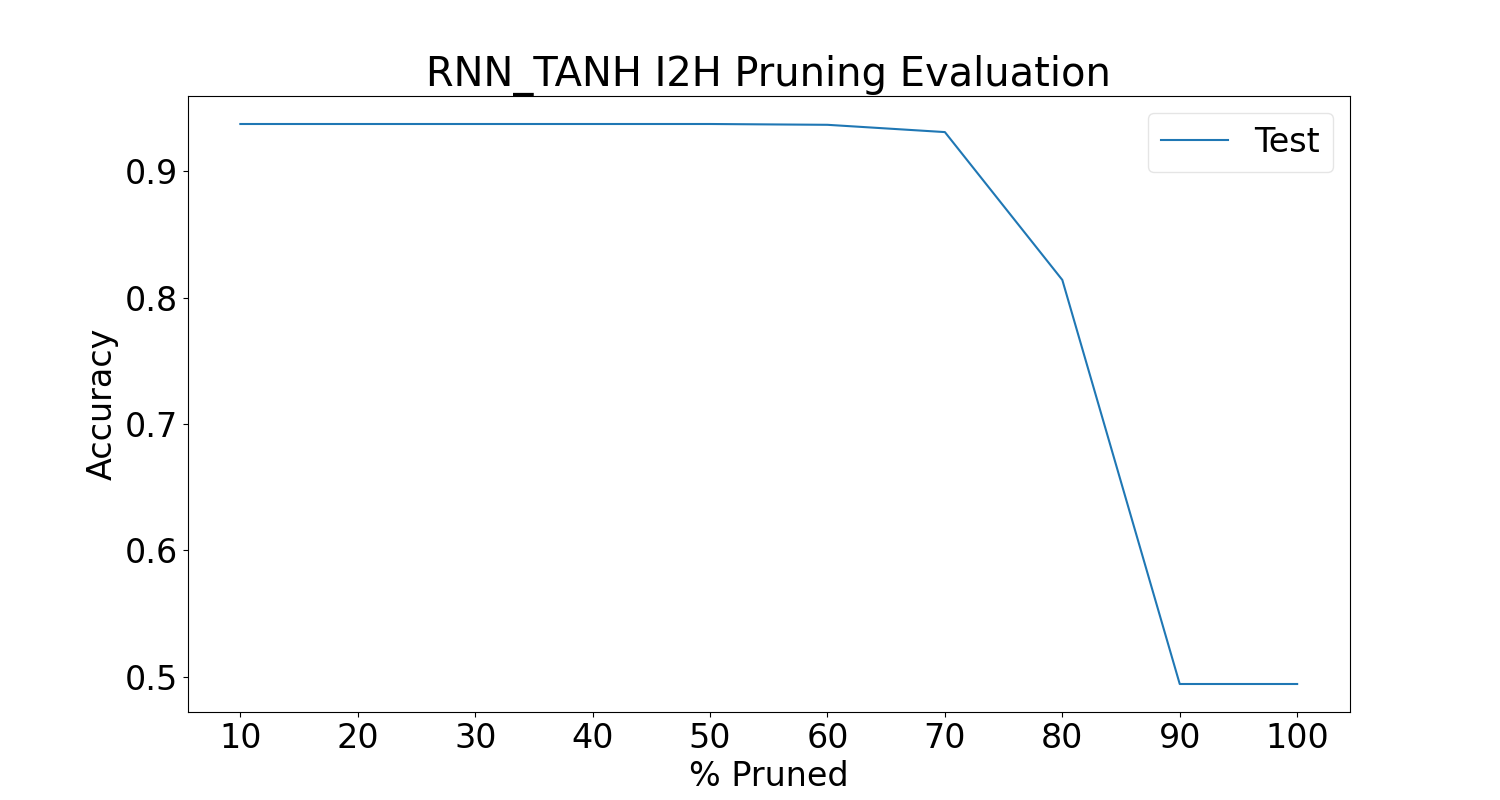}
	\caption[RNN\_Tanh base model performance after pruning i2h weights]%
	{Base model performance of RNN with Tanh nonlinearity after pruning only input-to-hidden weights. The pruning starts from $10\%$ and ends at $100\%$ with an increment of $10$ after each pruning round.}
	\label{fig:rnn_tanh_i2h_prune}
\end{figure}

First, we look at how pruning only input-to-hidden weights affect the performance of RNN with Tanh nonlinearity. As shown in figure \ref{fig:rnn_tanh_i2h_prune}, the model performs consistently at above 90\% accuracy with 70\% of input-to-hidden weights pruned. We observe a sharp drop in accuracy at 80\% pruning.

\begin{figure}[h]
	\centering
	\includegraphics[width=0.8\linewidth]{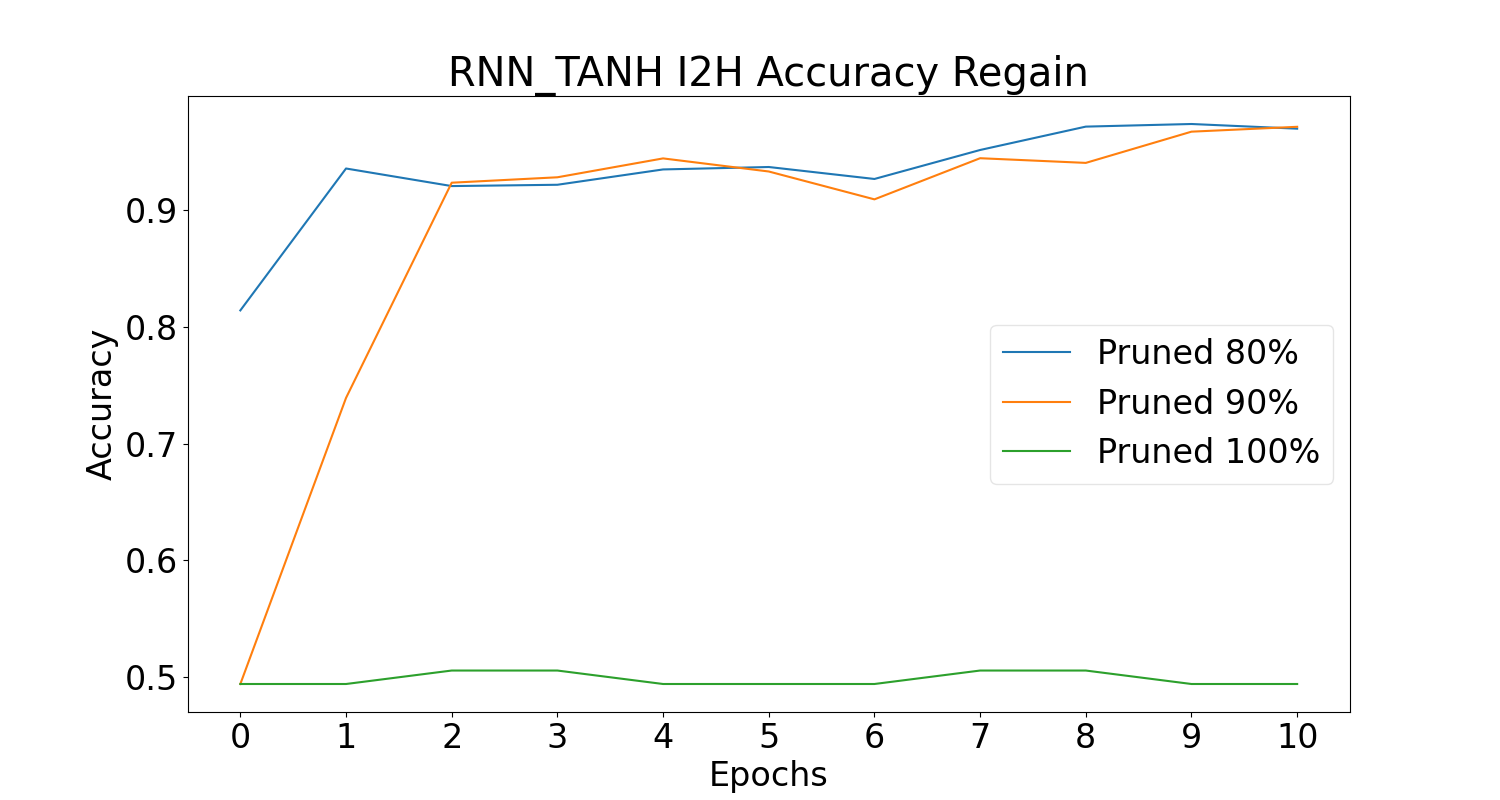}
	\caption[RNN\_Tanh base model performance regain after pruning i2h weights]%
	{The number of epochs required to regain the accuracy of RNN\_Tanh model after pruning 80\%, 90\%, and 100\% of input-to-hidden weights.}
	\label{fig:rnn_tanh_i2h_prune_regain}
\end{figure}

As shown in the above figure, the model starts performing worse from 80\% pruning and above. Figure \ref{fig:rnn_tanh_i2h_prune_regain} shows the number of epochs it takes for this pruned model to regain accuracy after pruning 80\%, 90\%, and 100\% of input-to-hidden weights.

As shown, the model regains above 90\% accuracy only after one epoch with 80\% pruning, while it takes two epochs to regain accuracy after 90\% pruning. The model never recovers after pruning 100\% of input-to-hidden weights.

\begin{figure}[H]
	\centering
	\includegraphics[width=0.8\linewidth]{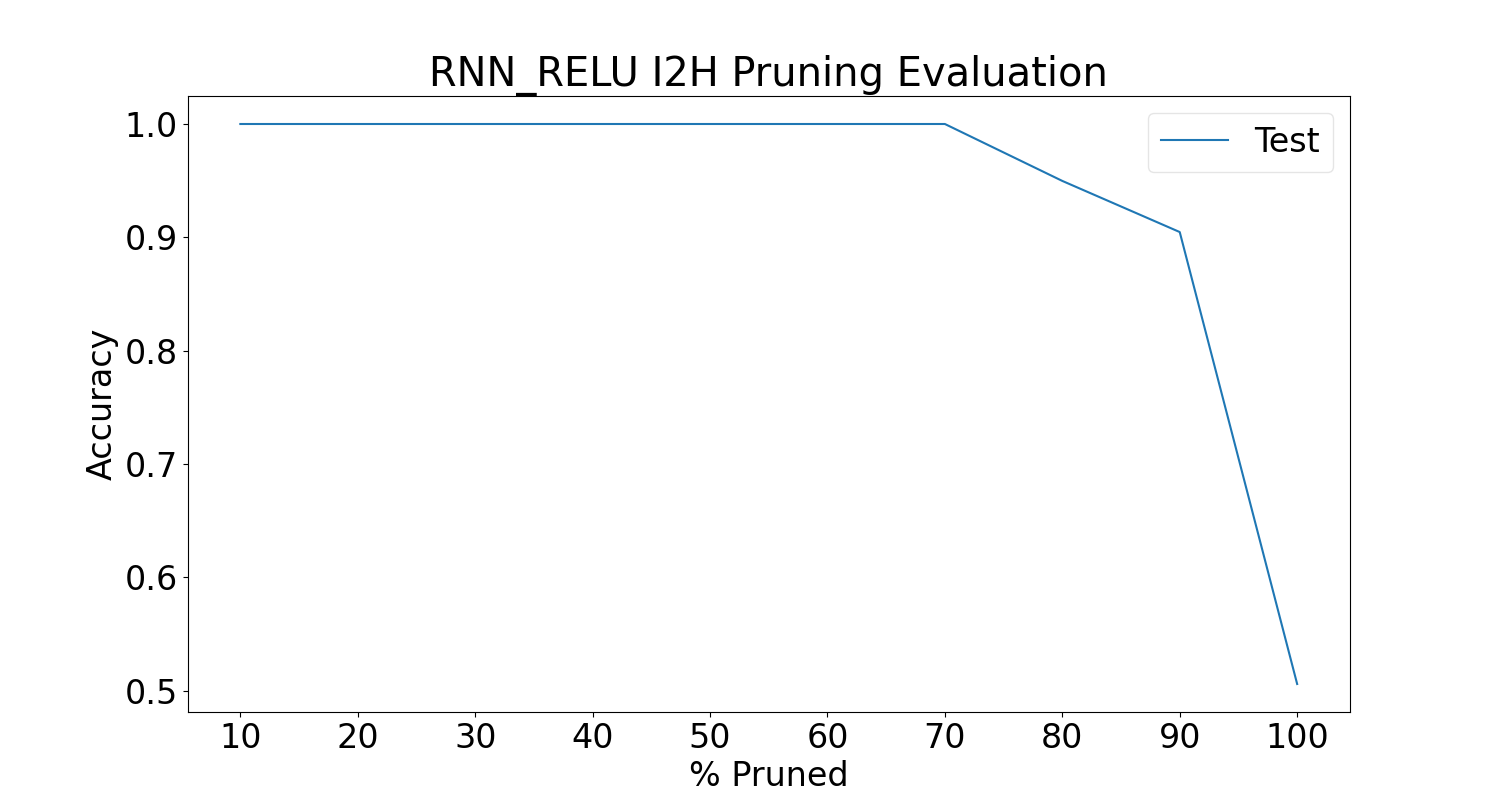}
	\caption[RNN\_ReLU base model performance after pruning i2h weights]%
	{Base model performance of RNN with ReLU nonlinearity after pruning only input-to-hidden weights. The pruning starts from $10\%$ and ends at $100\%$ with an increment of $10$ after each pruning round.}
	\label{fig:rnn_relu_i2h_prune}
\end{figure}

Next, we look at how pruning only input-to-hidden weights affect the performance of RNN with ReLU nonlinearity. As shown in figure \ref{fig:rnn_relu_i2h_prune}, the model performs consistently at above 90\% accuracy with 80\% of input-to-hidden weights pruned. We observe a sharp drop in accuracy at 90\% pruning.

\begin{figure}[h]
	\centering
	\includegraphics[width=0.8\linewidth]{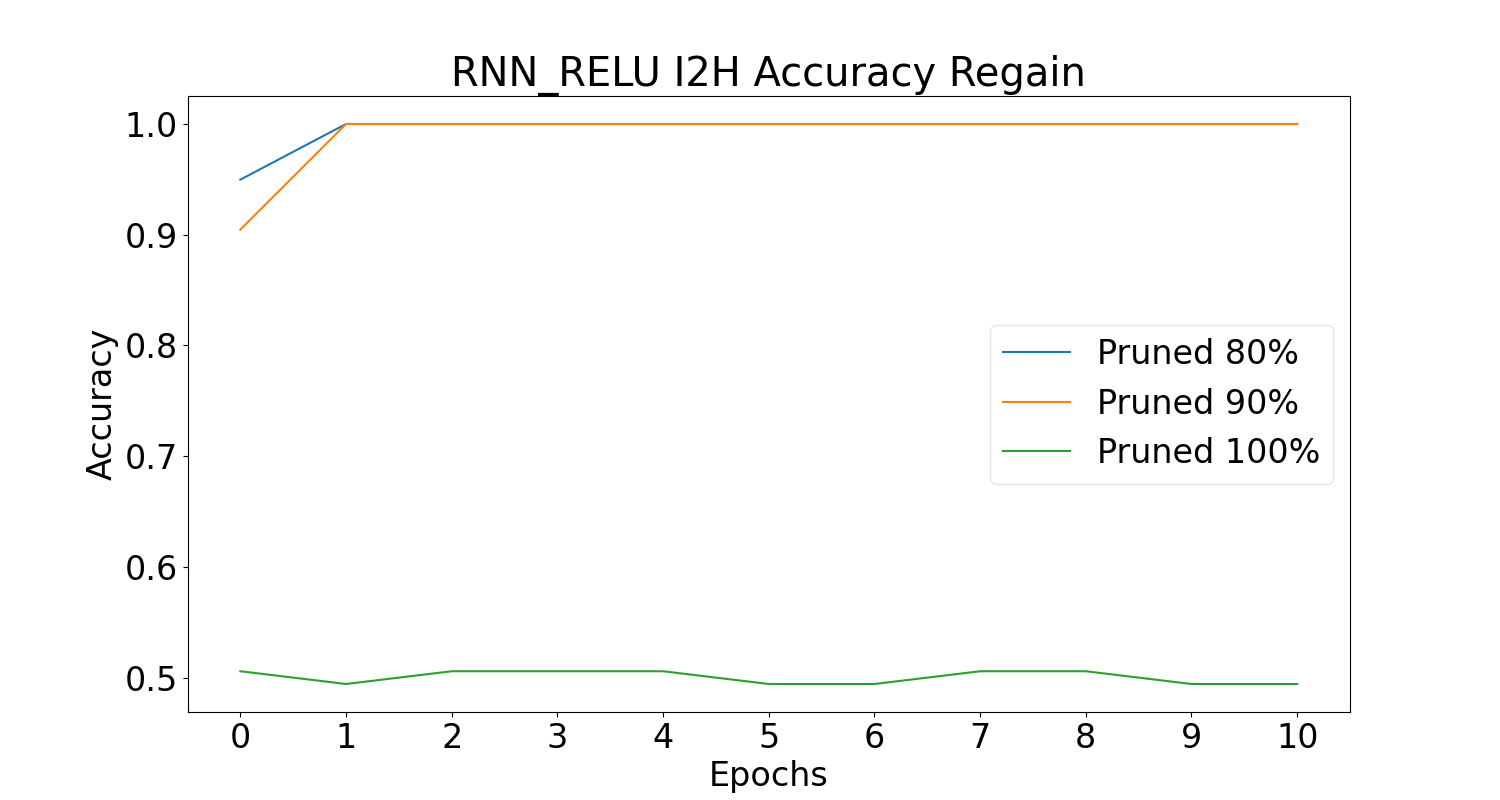}
	\caption[RNN\_ReLU base model performance regain after pruning i2h weights]%
	{The number of epochs required to regain the accuracy of RNN\_ReLU model after pruning 80\%, 90\%, and 100\% of input-to-hidden weights.}
	\label{fig:rnn_relu_i2h_prune_regain}
\end{figure}

As shown in the above figure, the model starts dropping accuracy from 80\% pruning and above. Figure \ref{fig:rnn_relu_i2h_prune_regain} shows the number of epochs it takes for this pruned model to regain accuracy after pruning 80\%, 90\%, and 100\% of input-to-hidden weights.

As shown, the model regains above 90\% accuracy only after one epoch with 80\% and 90\% pruning, while the model never recovers after pruning 100\% of input-to-hidden weights.

Next, the following graph shows the effect of pruning only input-to-hidden weights on our LSTM model's performance:

\begin{figure}[h]
	\centering
	\includegraphics[width=0.8\linewidth]{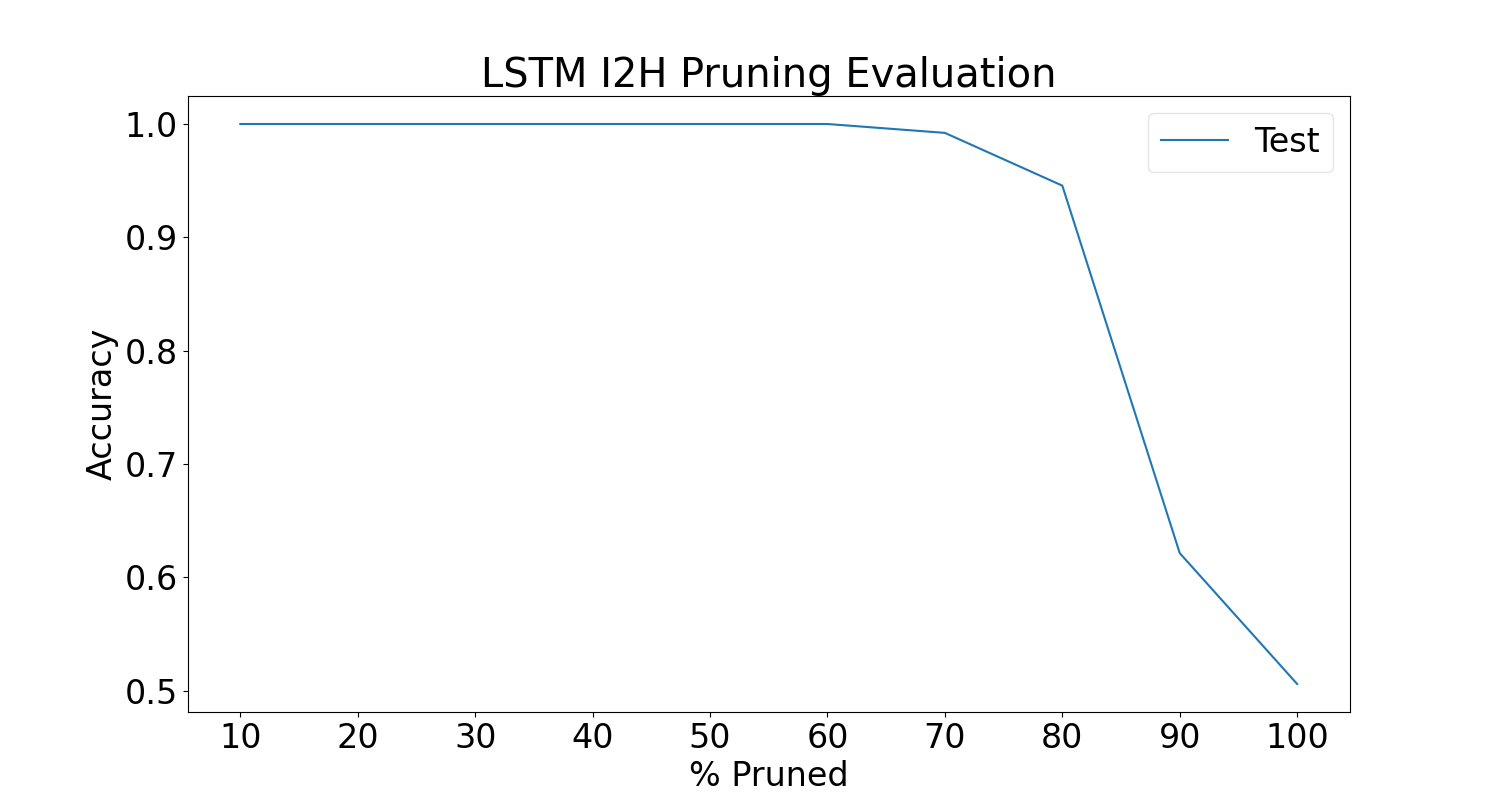}
	\caption[LSTM base model performance after pruning i2h weights]%
	{Base model performance of LSTM after pruning only input-to-hidden weights. The pruning starts from $10\%$ and ends at $100\%$ with an increment of $10$ after each pruning round.}
	\label{fig:lstm_i2h_prune}
\end{figure}

\begin{figure}[h]
	\centering
	\includegraphics[width=0.8\linewidth]{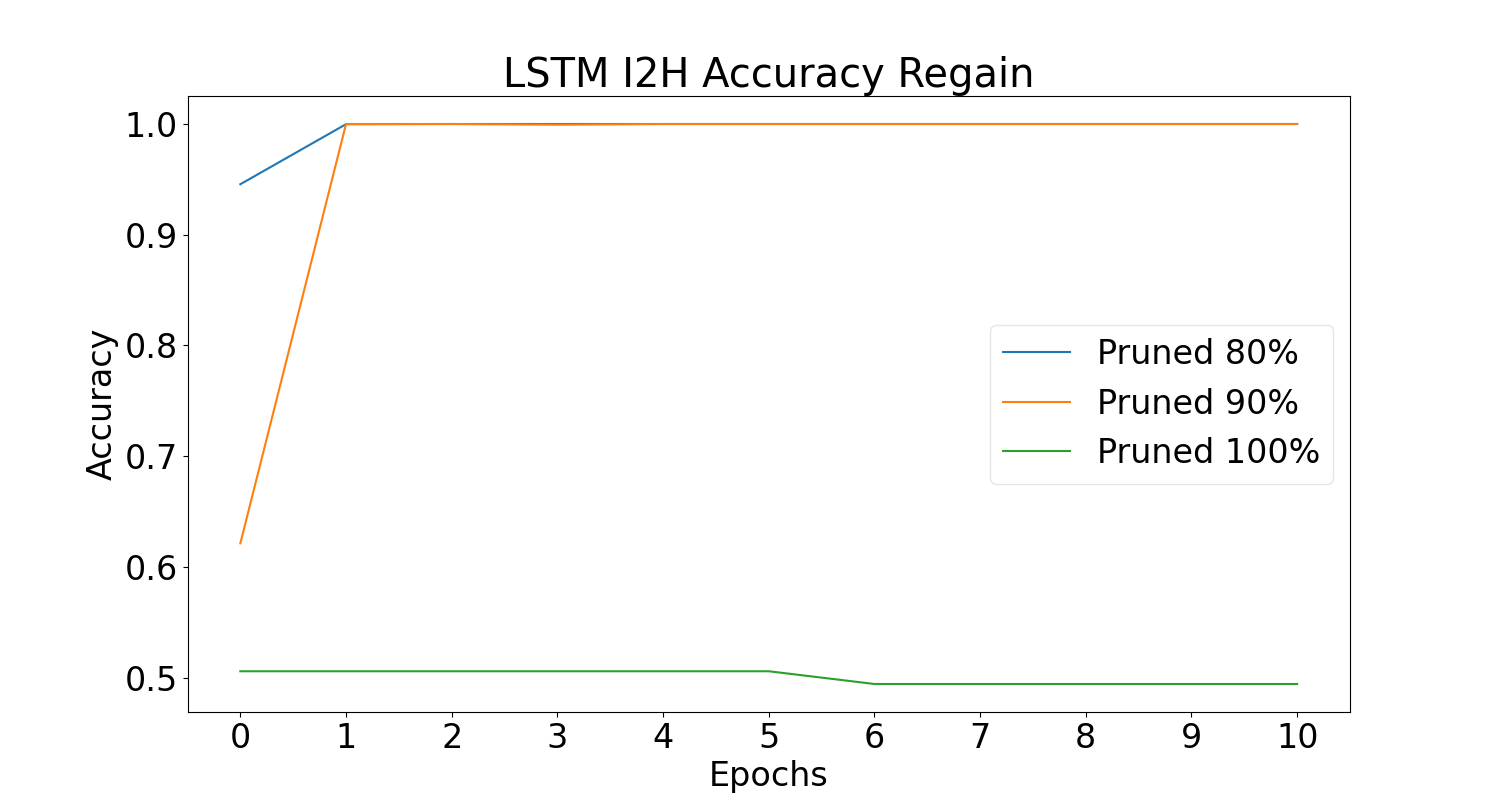}
	\caption[LSTM base model performance regain after pruning i2h weights]%
	{The number of epochs required to regain the accuracy of LSTM model after pruning 80\%, 90\%, and 100\% of input-to-hidden weights.}
	\label{fig:lstm_i2h_prune_regain}
\end{figure}

As we can see in the above graph, the model performs consistently at around 95\% accuracy with 80\% pruning. The model's performance drops sharply at 90\% pruning and returns baseline accuracy at 100\% pruning. Figure \ref{fig:lstm_i2h_prune_regain} shows the number of epochs it takes for the LSTM model to recover after pruning 80\%, 90\%, and 100\% of its input-to-hidden weights.

Finally, the following graph shows the effect of pruning only input-to-hidden weights on our GRU model's performance:

\begin{figure}[h]
	\centering
	\includegraphics[width=0.8\linewidth]{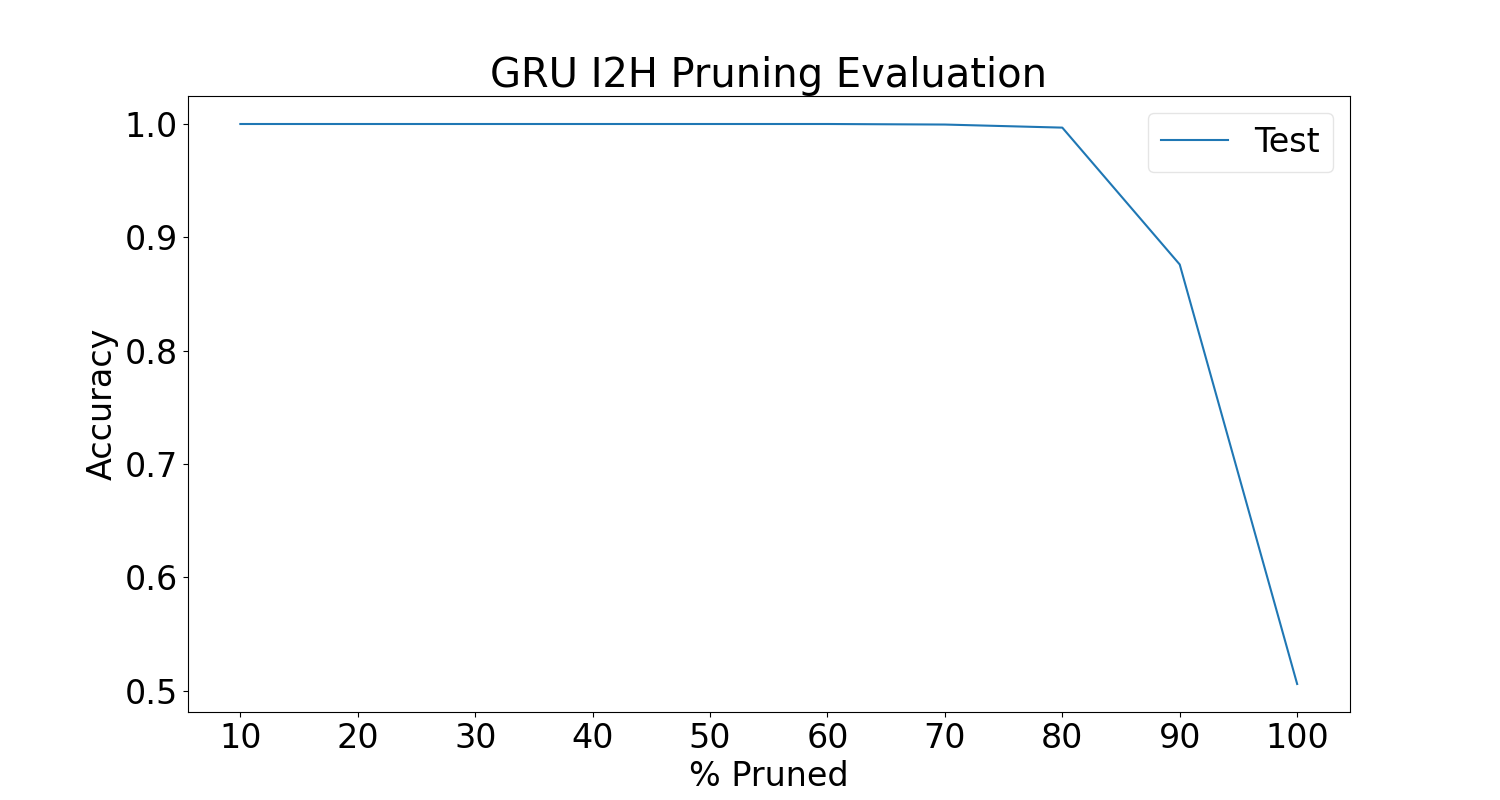}
	\caption[GRU base model performance after pruning i2h weights]%
	{Base model performance of GRU after pruning only input-to-hidden weights. The pruning starts from $10\%$ and ends at $100\%$ with an increment of $10$ after each pruning round.}
	\label{fig:gru_i2h_prune}
\end{figure}

\begin{figure}[h]
	\centering
	\includegraphics[width=0.8\linewidth]{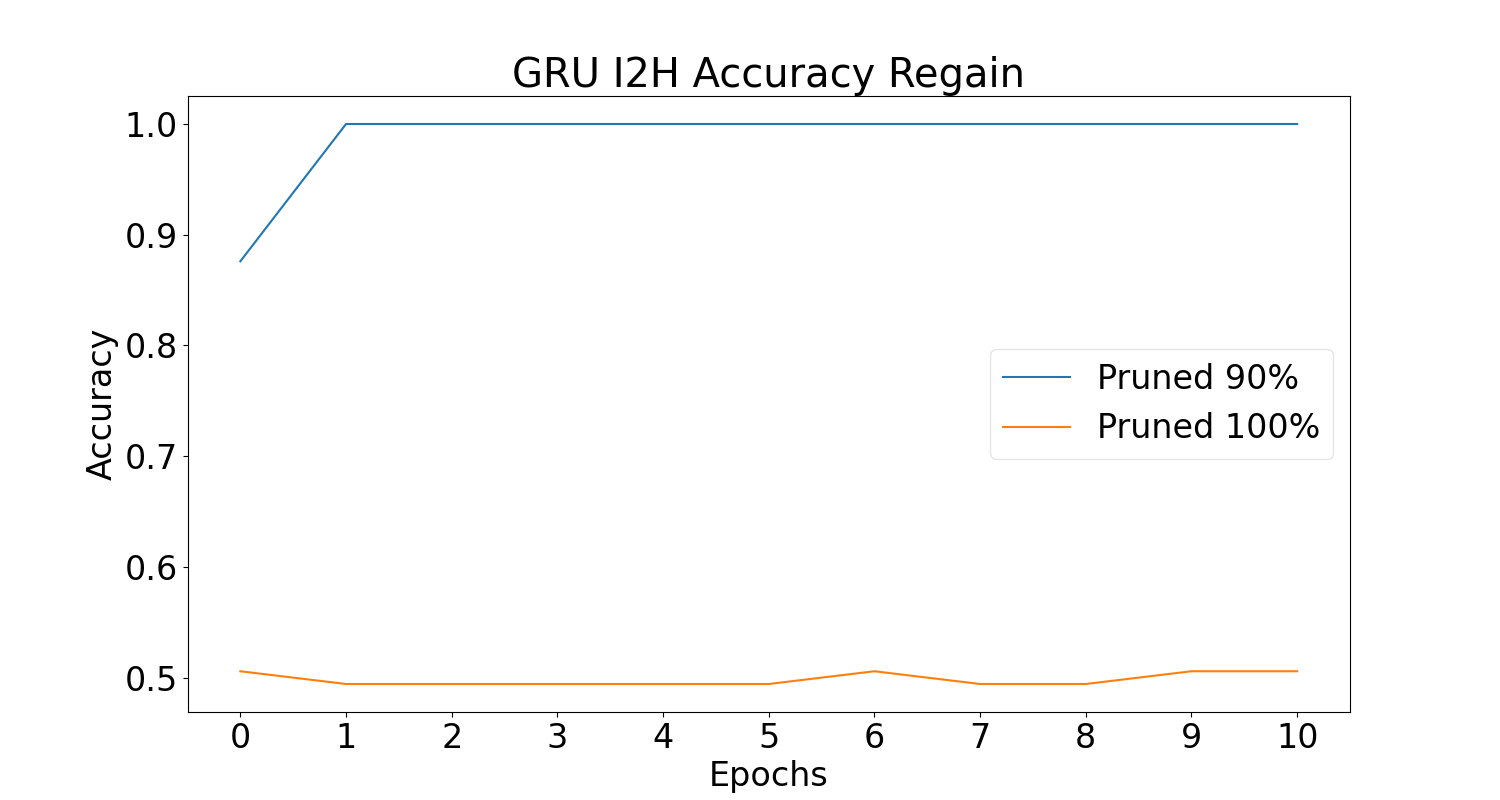}
	\caption[GRU base model performance regain after pruning i2h weights]%
	{The number of epochs required to regain the accuracy of GRU model after pruning 90\%, and 100\% of input-to-hidden weights.}
	\label{fig:gru_i2h_prune_regain}
\end{figure}

As we can see in the above graph, the model performs consistently at above 95\% accuracy with 80\% pruning. The model's performance drops sharply at 90\% pruning and returns baseline accuracy at 100\% pruning. Figure \ref{fig:gru_i2h_prune_regain} shows the number of epochs it takes for the GRU model to recover after pruning 90\%, and 100\% of its input-to-hidden weights.

As we can see in the above graph, it only takes one epoch for the GRU model to recover from pruning 90\% of its input-to-hidden weights. The model never recovers from pruning 100\% of its input-to-hidden weights.

\subsection{Pruning only hidden-to-hidden weights}

In this section, we present the plots visualizing the effects of pruning only hidden-to-hidden weights on each RNN variant's performance.

\begin{figure}[h]
	\centering
	\includegraphics[width=0.8\linewidth]{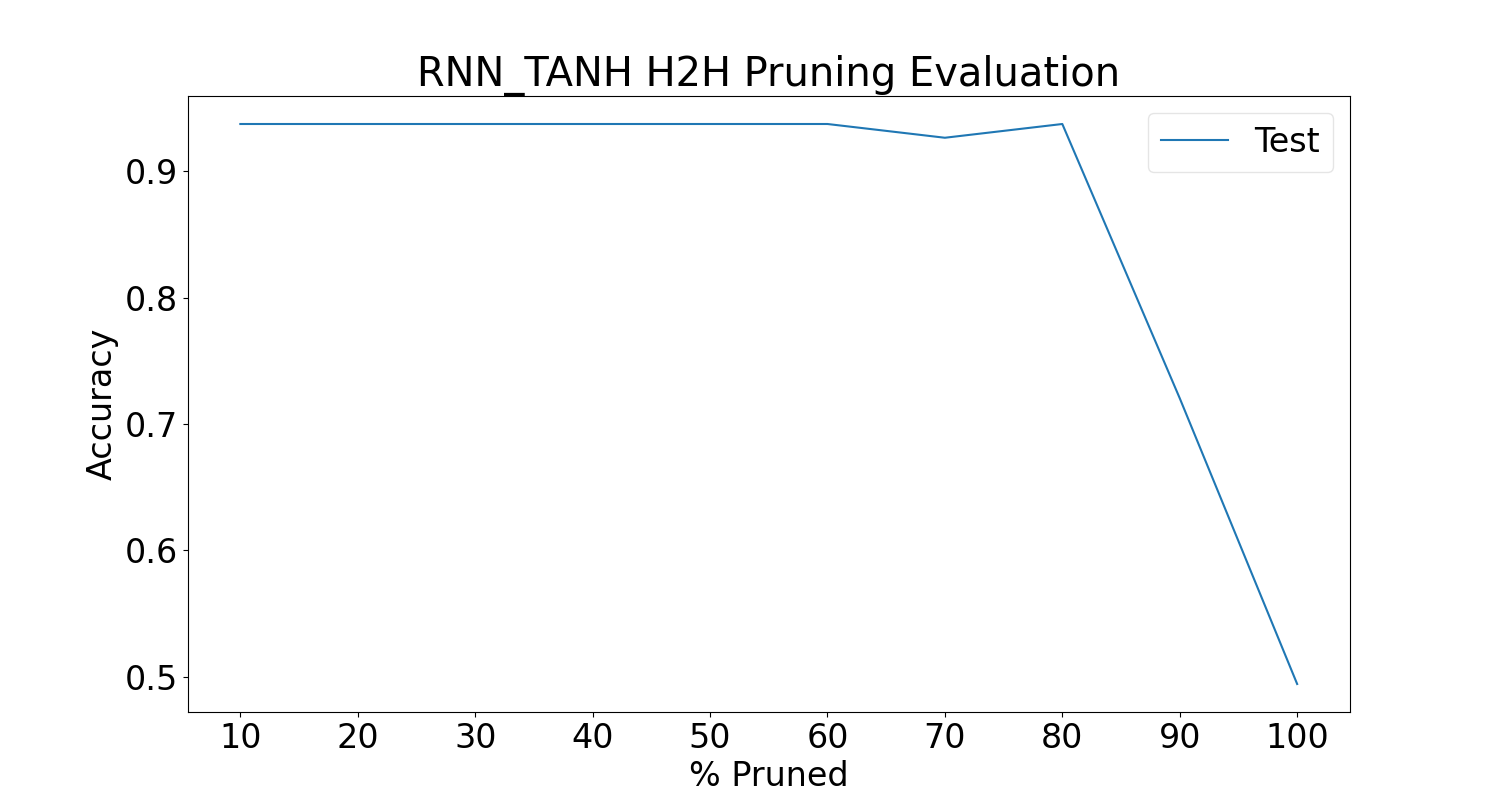}
	\caption[RNN\_Tanh base model performance after pruning h2h weights]%
	{Base model performance of RNN with Tanh nonlinearity after pruning only hidden-to-hidden weights. The pruning starts from $10\%$ and ends at $100\%$ with an increment of $10$ after each pruning round.}
	\label{fig:rnn_tanh_h2h_prune}
\end{figure}

Beginning with RNN with Tanh nonlinearity, as visualized in figure \ref{fig:rnn_tanh_h2h_prune}, the model performs consistently at above 90\% at 80\% of hidden-to-hidden weights pruned. We observe a sharp drop in accuracy at 90\% pruning, while the model returns baseline accuracy at 100\% pruning.

\begin{figure}[H]
	\centering
	\includegraphics[width=0.8\linewidth]{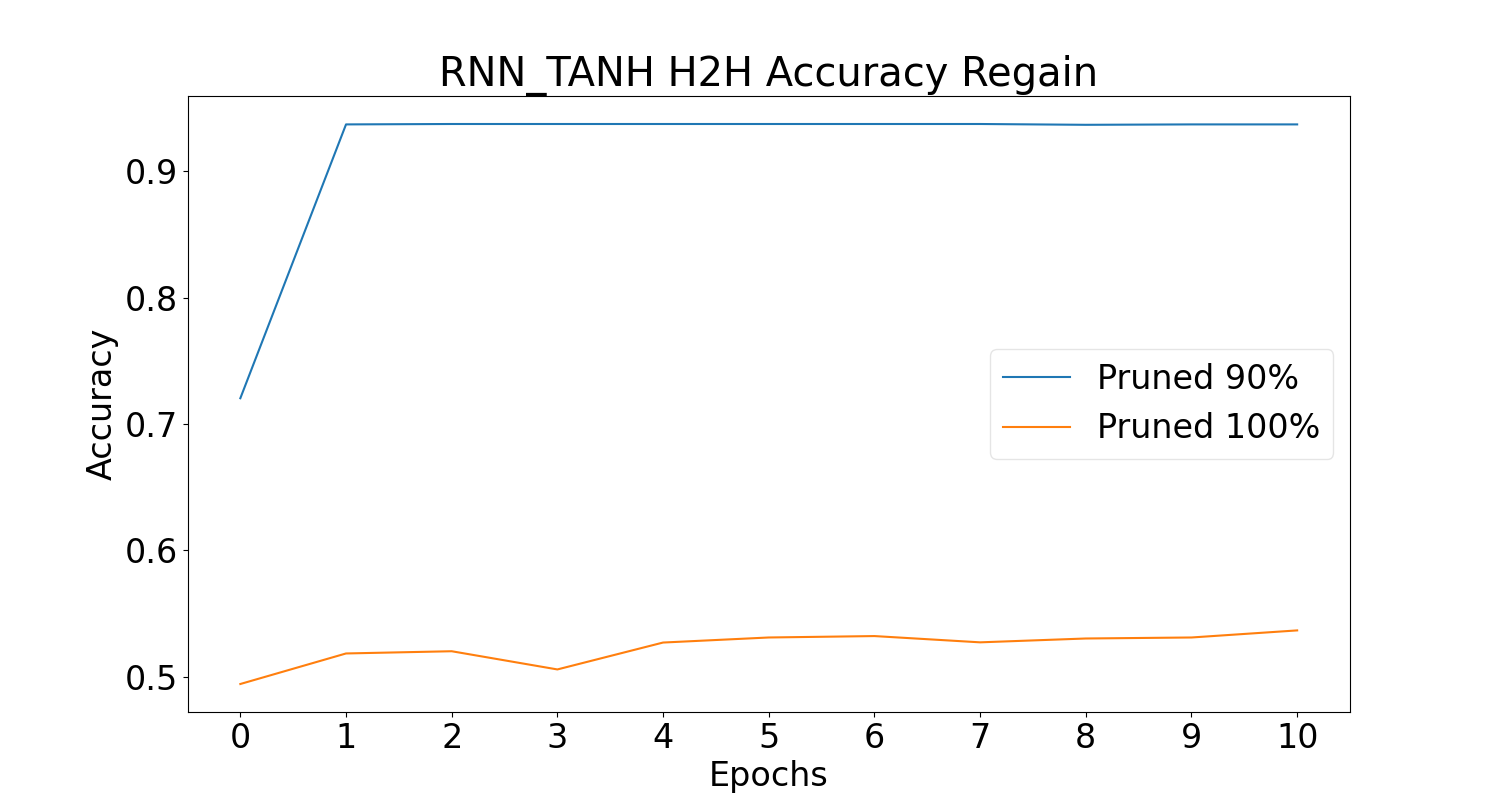}
	\caption[RNN\_Tanh base model performance regain after pruning h2h weights]%
	{The number of epochs required to regain the accuracy of RNN\_Tanh model after pruning 80\%, 90\%, and 100\% of hidden-to-hidden weights.}
	\label{fig:rnn_tanh_h2h_prune_regain}
\end{figure}

Since performance drops only at 90\% and above of pruning, we perform an accuracy regain experiment for only 90\% and 100\% pruning of hidden-to-hidden weights as shown in figure \ref{fig:rnn_tanh_h2h_prune_regain}.

As shown in the above graph, it only takes one epoch for the model to recover from 90\% of hidden-to-hidden weights pruned, while the model never recovers after pruning 100\% of hidden-to-hidden weights.

Next, we look at the effect of pruning hidden-to-hidden weights on the performance of RNN with ReLU nonlinearity. As shown in the below figure, the model performs consistently at above 95\% accuracy at 70\% pruning and starts dropping the performance from 80\% of pruning.

\begin{figure}[h]
	\centering
	\includegraphics[width=0.8\linewidth]{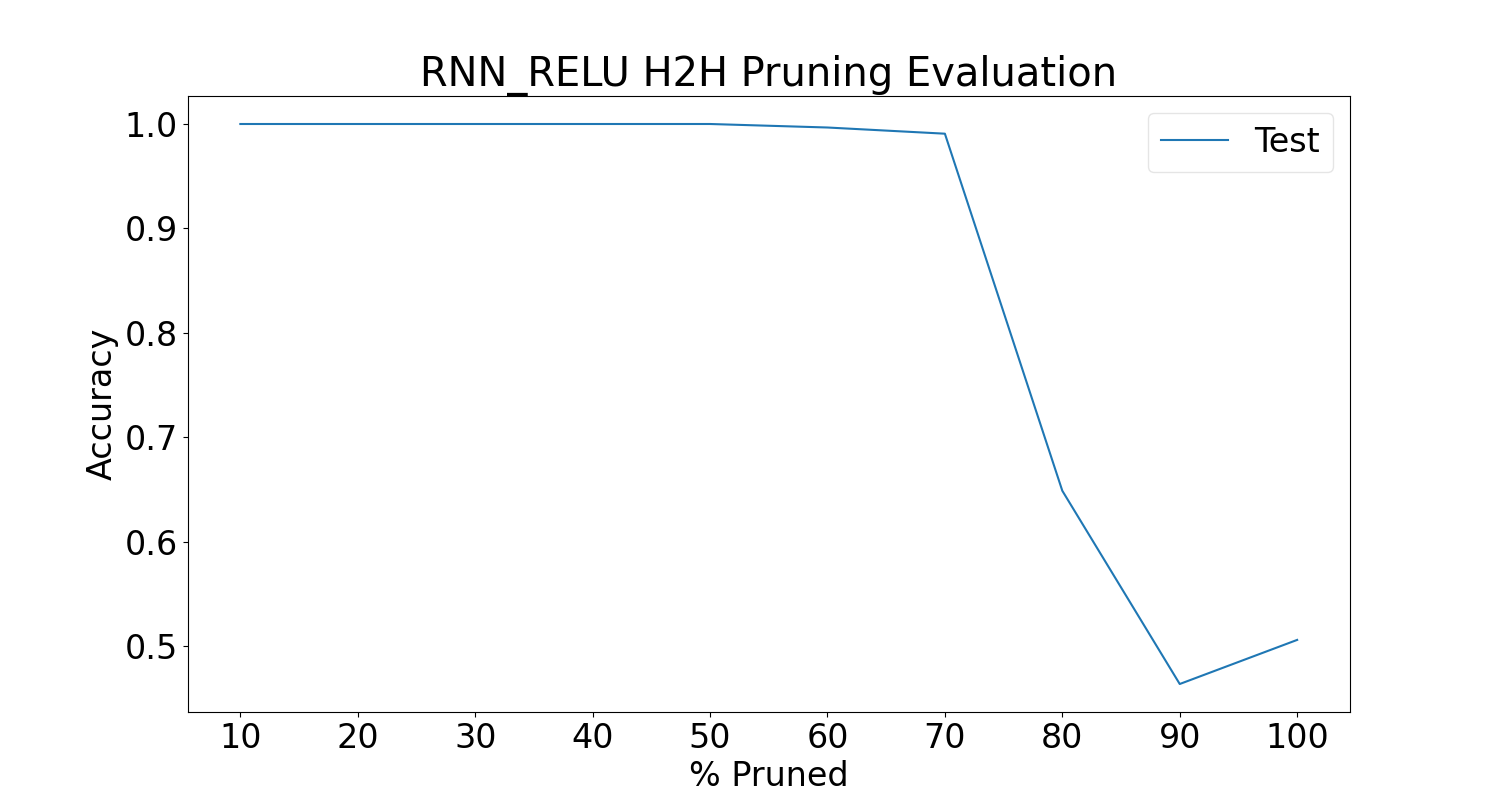}
	\caption[RNN\_ReLU base model performance after pruning h2h weights]%
	{Base model performance of RNN with ReLU nonlinearity after pruning only hidden-to-hidden weights. The pruning starts from $10\%$ and ends at $100\%$ with an increment of $10$ after each pruning round.}
	\label{fig:rnn_relu_h2h_prune}
\end{figure}

The following graph shows the required number of epochs for the RNN with ReLU nonlinearity to recover from 80\%, 90\%, and 100\% pruning.

\begin{figure}[H]
	\centering
	\includegraphics[width=0.8\linewidth]{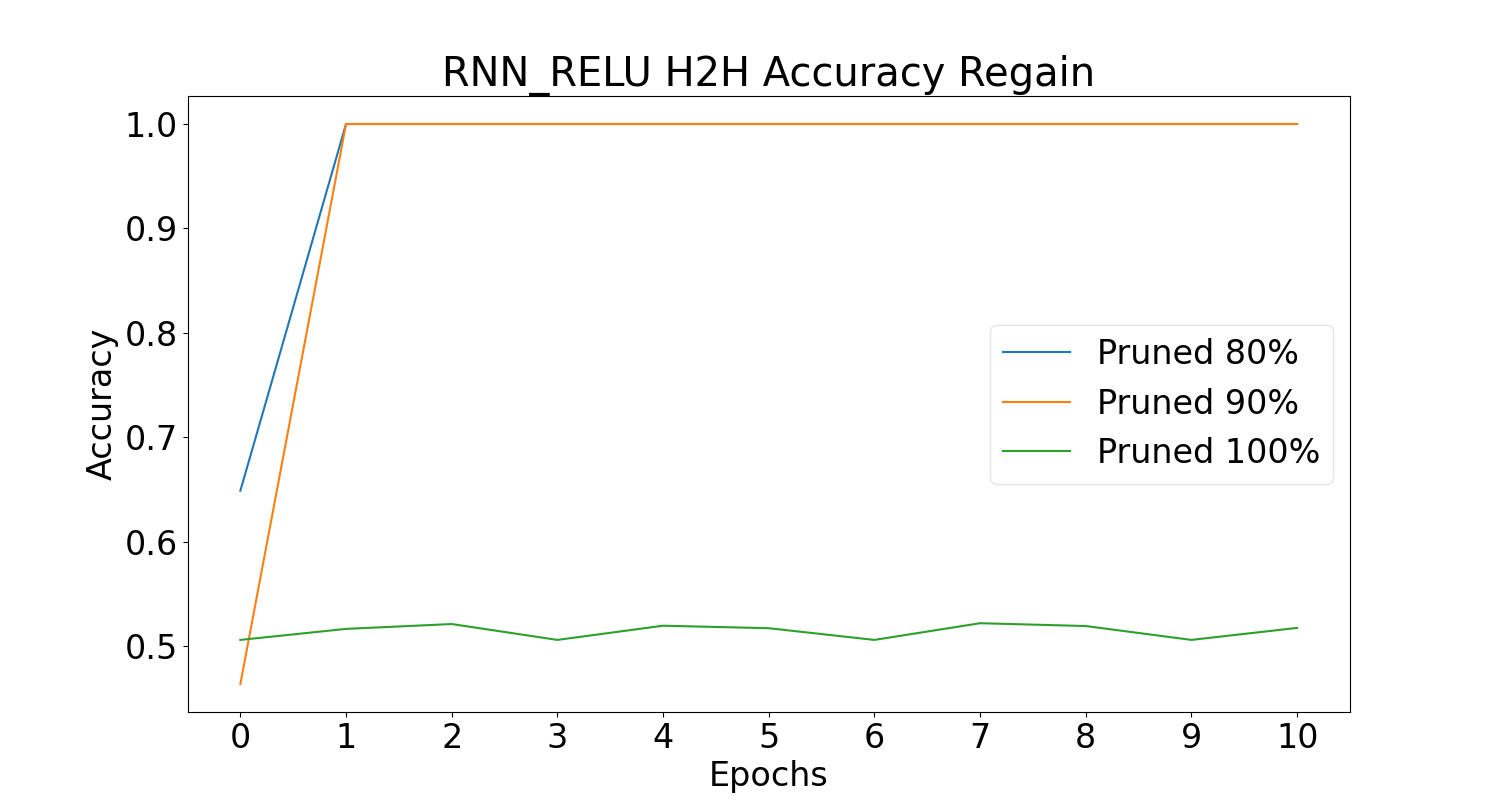}
	\caption[RNN\_ReLU base model performance regain after pruning h2h weights]%
	{The number of epochs required to regain the accuracy of RNN\_ReLU model after pruning 80\%, 90\%, and 100\% of hidden-to-hidden weights.}
	\label{fig:rnn_relu_h2h_prune_regain}
\end{figure}

As shown in the above figure, the model only requires one epoch to recover from 80\% and 90\% of hidden-to-hidden weights pruning. The model, as before, never recovers from 100\% pruning.

\begin{figure}[h]
	\centering
	\includegraphics[width=0.8\linewidth]{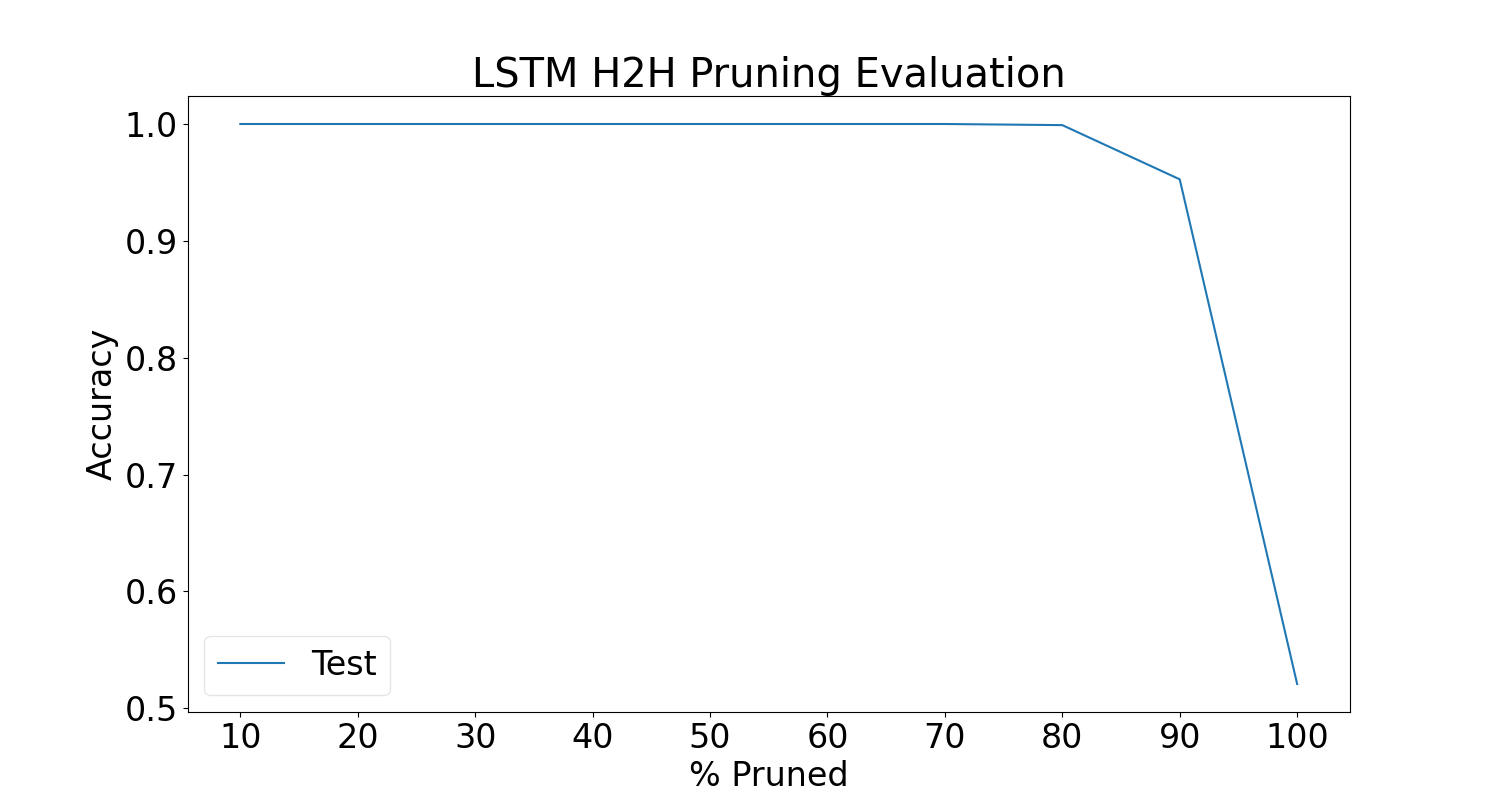}
	\caption[LSTM base model performance after pruning h2h weights]%
	{Base model performance of LSTM after pruning only hidden-to-hidden weights. The pruning starts from $10\%$ and ends at $100\%$ with an increment of $10$ after each pruning round.}
	\label{fig:lstm_h2h_prune}
\end{figure}

Next, we look at the effect of pruning hidden-to-hidden weights on the performance of LSTM, where, as shown in figure \ref{fig:lstm_h2h_prune}, the model performs consistently at around 95\% accuracy at 90\% pruning and drops the performance at 100\% of pruning.

The following graph shows the required number of epochs for the LSTM to recover from 100\% pruning.

\begin{figure}[h]
	\centering
	\includegraphics[width=0.8\linewidth]{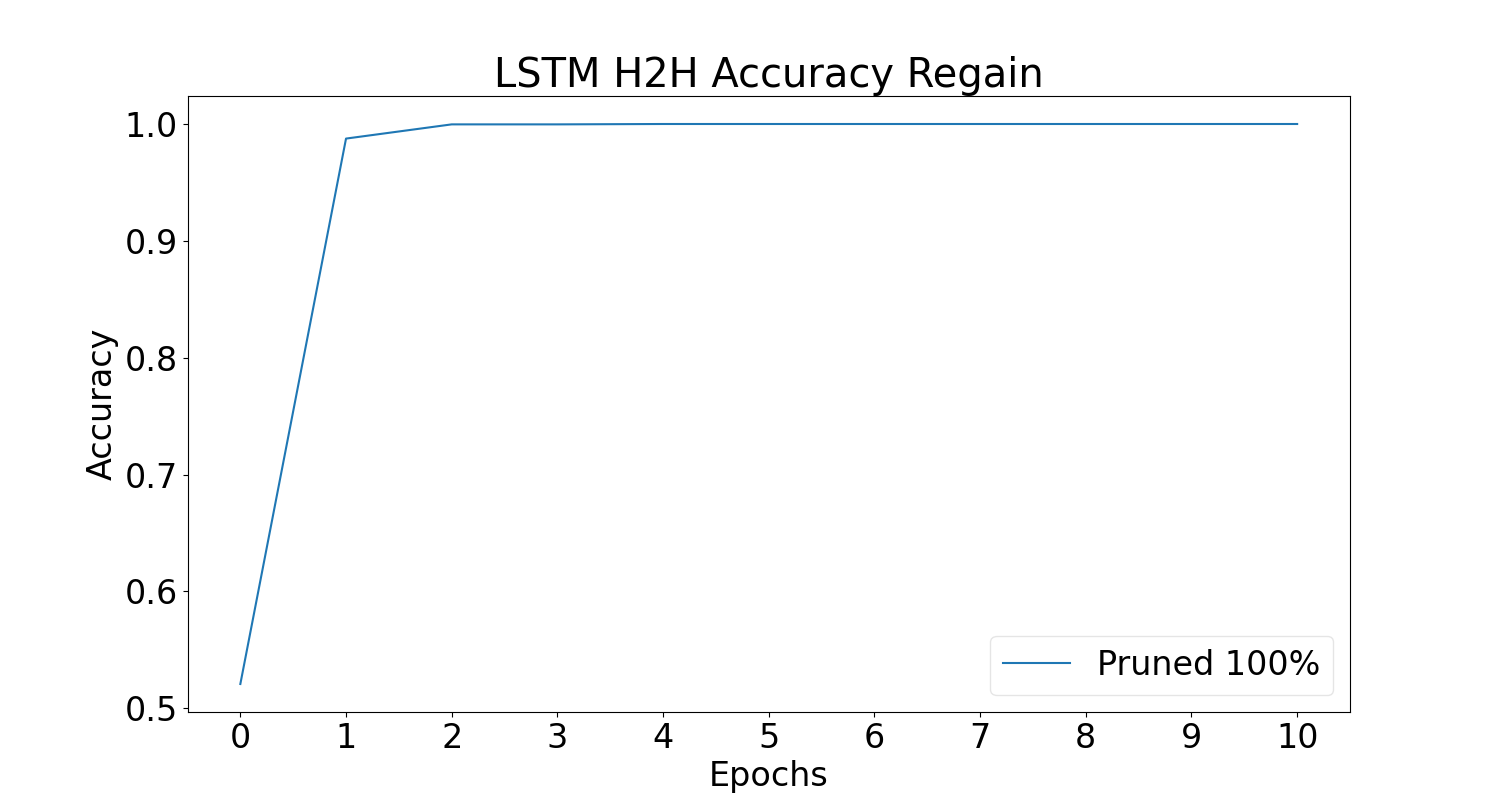}
	\caption[LSTM base model performance regain after pruning h2h weights]%
	{The number of epochs required to regain the accuracy of LSTM model after pruning 100\% of hidden-to-hidden weights.}
	\label{fig:lstm_h2h_prune_regain}
\end{figure}

As shown in the above figure, even at 100\% pruning of hidden-to-hidden weights, the model still recovers in just one epoch. The reason behind this recovery is the existence of a cell state, as explained in section \ref{section:lstm}.

Finally, similar to LSTM, GRU also performs consistently at around 95\% at 90\% of hidden-to-hidden weights pruning, as shown in the following figure:

\begin{figure}[h]
	\centering
	\includegraphics[width=0.8\linewidth]{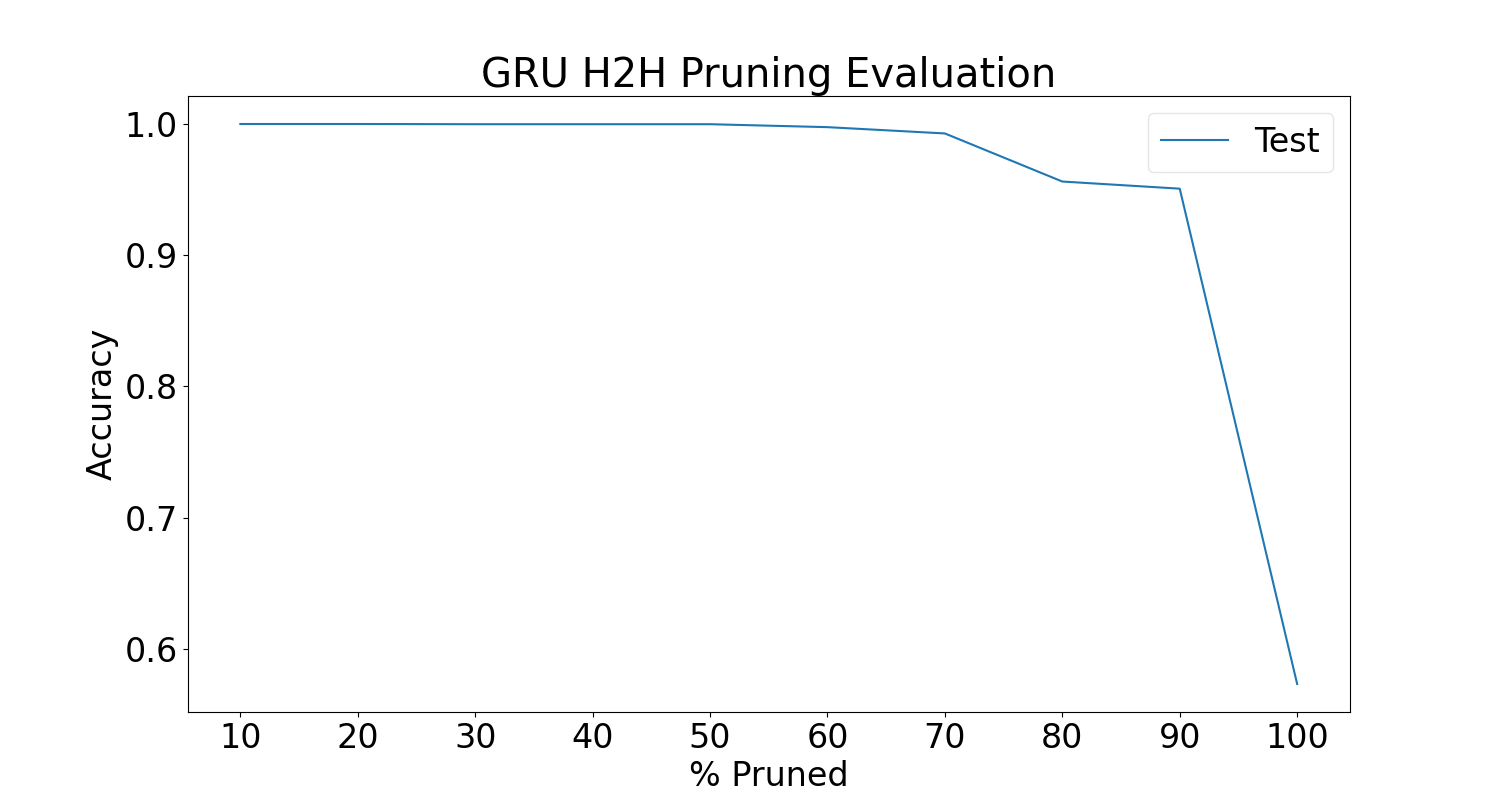}
	\caption[GRU base model performance after pruning h2h weights]%
	{Base model performance of GRU after pruning only hidden-to-hidden weights. The pruning starts from $10\%$ and ends at $100\%$ with an increment of $10$ after each pruning round.}
	\label{fig:gru_h2h_prune}
\end{figure}

The following graph shows the required number of epochs for the GRU to recover from 100\% pruning.

\begin{figure}[h]
	\centering
	\includegraphics[width=0.8\linewidth]{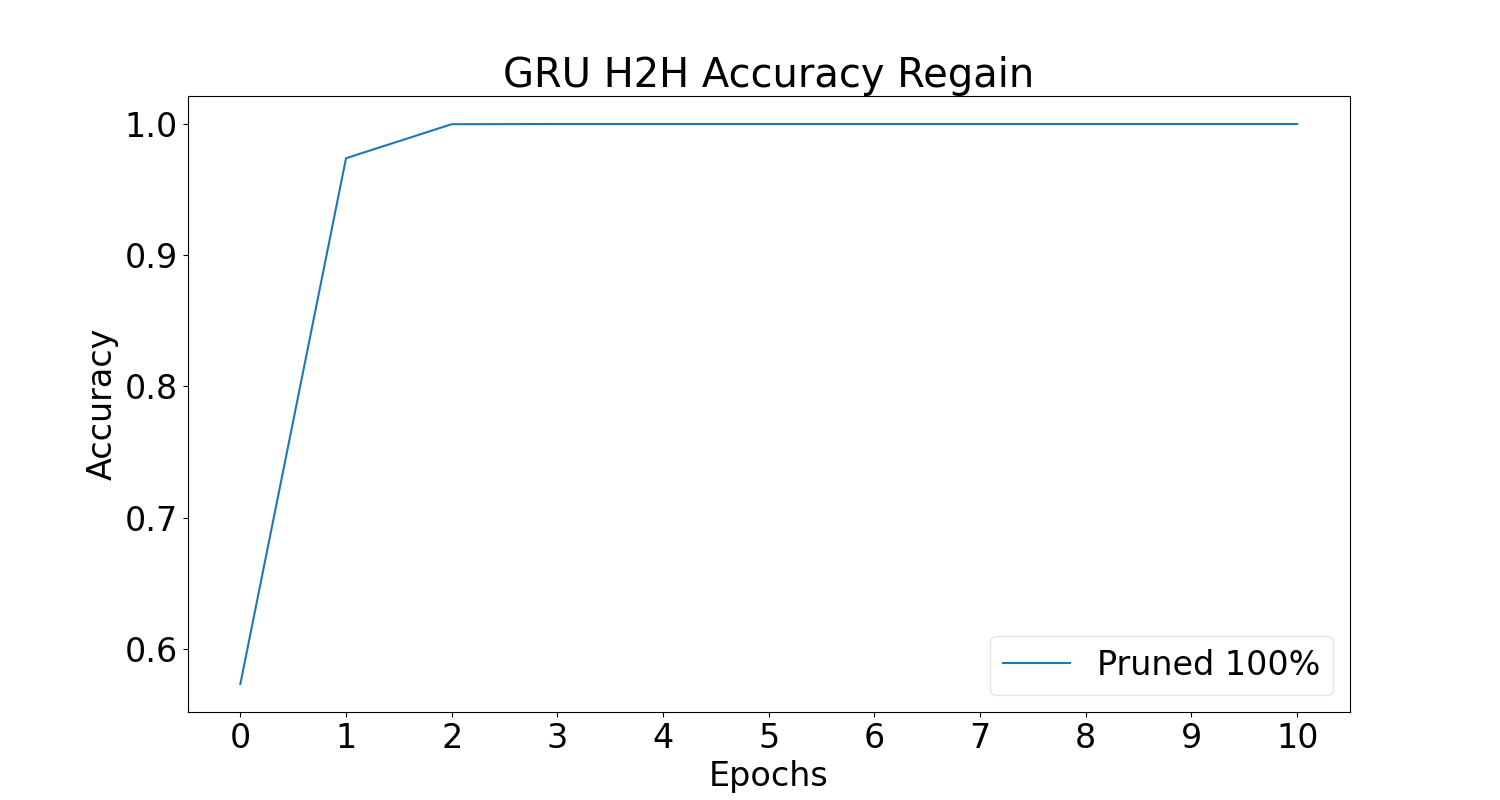}
	\caption[GRU base model performance regain after pruning h2h weights]%
	{The number of epochs required to regain the accuracy of GRU model after pruning 100\% of hidden-to-hidden weights.}
	\label{fig:gru_h2h_prune_regain}
\end{figure}

Like the LSTM model, even at 100\% pruning of hidden-to-hidden weights, the model still recovers in just one epoch, and after two epochs, the model returns 100\% accuracy.

This concludes the results of the pruning experiment. In the next section, we present the results of our randomly structured recurrent networks.


\section{Performance of Randomly Structured Recurrent Networks}\label{section:r_st_rn}

Our second experiment was to construct a random structure recurrent network using a random graph and train it on the Reber grammar dataset. During this training, we record different graph properties of the base random graph and try to correlate it with the test performance. In this section, we present the results of this experiment.

\subsection{Randomly Structured RNN with Tanh nonlinearity}

We begin with the plot that compares the accuracy of the Watts–Strogatz model and Barabási–Albert model based random structure RNN\_Tanh.

\begin{figure}[h]
	\centering
	\includegraphics[width=0.45\linewidth]{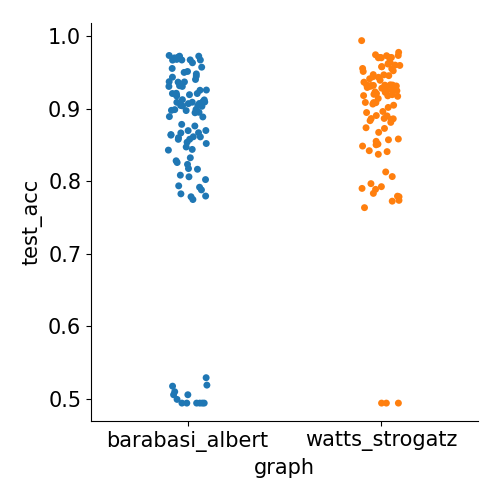}
	\caption[Performance of the WS and BA based random structure RNN\_Tanh]%
	{Plot comparing the performance of the Watts–Strogatz and Barabási–Albert model based random structure RNN\_Tanh.}
	\label{fig:tanh_acc_comp}
\end{figure}

Next, the following table shows the Pearson correlation between test accuracy and different graph and recurrent network properties. The corresponding important plots are later shown in appendix \ref{app:rs_tanh}.

\begin{table}[h]
	\centering
	\begin{tabular}{|l|c|}
	    \hline
		\textbf{Property} & \textbf{Correlation with $test\_acc$}\\
		\hline
		layers & 0.25\\
		nodes & \textbf{0.40}\\
		edges & 0.38\\
		source\_nodes & 0.35\\
		diameter & -0.23\\
		density & 0.29\\
		average\_shortest\_path\_length & -0.27\\
		eccentricity\_var & -0.22\\
		degree\_var & -0.28\\
		closeness\_var & \textbf{-0.46}\\
		nodes\_betweenness\_var & \textbf{-0.49}\\
		edge\_betweenness\_var & -0.34\\
		\hline
	\end{tabular}
	\caption[Pearson correlation between test accuracy of RNN\_Tanh and different graph and recurrent network properties]{Pearson correlation between test accuracy of RNN\_Tanh and different graph and recurrent network properties}
	\label{tab:tanh_corr}
\end{table}

\subsection{Randomly Structured RNN with ReLU nonlinearity}

As with RNN\_Tanh, we begin with the plot that compares the accuracy of the Watts–Strogatz model and Barabási–Albert model based random structure RNN\_ReLU.

\begin{figure}[H]
	\centering
	\includegraphics[width=0.45\linewidth]{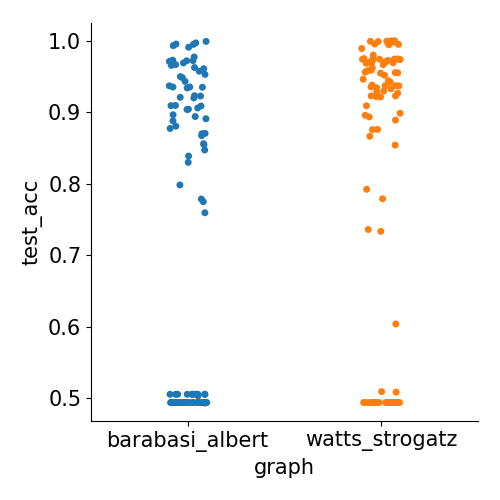}
	\caption[Performance of the WS and BA based random structure RNN\_ReLU]%
	{Plot comparing the performance of the Watts–Strogatz and Barabási–Albert model based random structure RNN\_ReLU.}
	\label{fig:relu_acc_comp}
\end{figure}

Next, the following table shows the Pearson correlation between test accuracy and different graph and recurrent network properties. The corresponding important plots are later shown in appendix \ref{app:rs_relu}.

\begin{table}[h]
	\centering
	\begin{tabular}{|l|c|}
	    \hline
		\textbf{Property} & \textbf{Correlation with $test\_acc$}\\
		\hline
		layers & 0.30\\
		nodes & \textbf{0.44}\\
		edges & \textbf{0.43}\\
		source\_nodes & \textbf{0.47}\\
		diameter & -0.27\\
		density & 0.15\\
		average\_shortest\_path\_length & -0.25\\
		eccentricity\_var & -0.24\\
		degree\_var & -0.26\\
		closeness\_var & -0.39\\
		nodes\_betweenness\_var & \textbf{-0.41}\\
		edge\_betweenness\_var & -0.30\\
		\hline
	\end{tabular}
	\caption[Pearson correlation between test accuracy of RNN\_ReLU and different graph and recurrent network properties]{Pearson correlation between test accuracy of RNN\_ReLU and different graph and recurrent network properties}
	\label{tab:relu_corr}
\end{table}

\subsection{Randomly Structured LSTM}

As before, we begin with the plot that compares the accuracy of the Watts–Strogatz model and Barabási–Albert model based random structure LSTM.

\begin{figure}[H]
	\centering
	\includegraphics[width=0.45\linewidth]{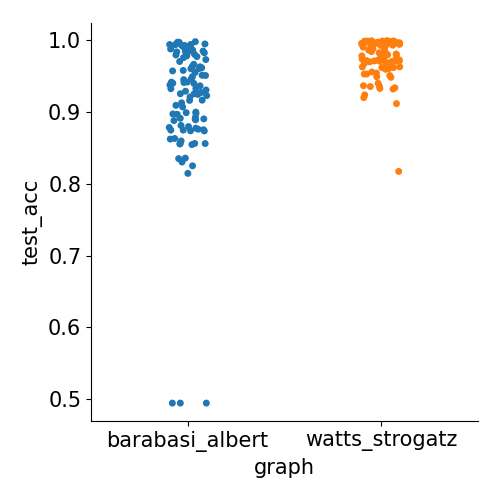}
	\caption[Performance of the WS and BA based random structure LSTM]%
	{Plot comparing the performance of the Watts–Strogatz and Barabási–Albert model based random structure LSTM.}
	\label{fig:lstm_acc_comp}
\end{figure}

Next, the following table shows the Pearson correlation between test accuracy and different graph and recurrent network properties. The corresponding important plots are later shown in appendix \ref{app:rs_lstm}.

\begin{table}[h]
	\centering
	\begin{tabular}{|l|c|}
	    \hline
		\textbf{Property} & \textbf{Correlation with $test\_acc$}\\
		\hline
		layers & 0.28\\
		nodes & \textbf{0.44}\\
		edges & \textbf{0.42}\\
		source\_nodes & \textbf{0.57}\\
		diameter & -0.32\\
		density & 0.29\\
		average\_shortest\_path\_length & -0.36\\
		eccentricity\_var & -0.30\\
		degree\_var & -0.39\\
		closeness\_var & \textbf{-0.51}\\
		nodes\_betweenness\_var & \textbf{-0.56}\\
		edge\_betweenness\_var & \textbf{-0.44}\\
		\hline
	\end{tabular}
	\caption[Pearson correlation between test accuracy of LSTM and different graph and recurrent network properties]{Pearson correlation between test accuracy of LSTM and different graph and recurrent network properties}
	\label{tab:lstm_corr}
\end{table}

\subsection{Randomly Structured GRU}

As with LSTM, we begin with the plot that compares the accuracy of the Watts–Strogatz model and Barabási–Albert model based random structure GRU.

\begin{figure}[H]
	\centering
	\includegraphics[width=0.45\linewidth]{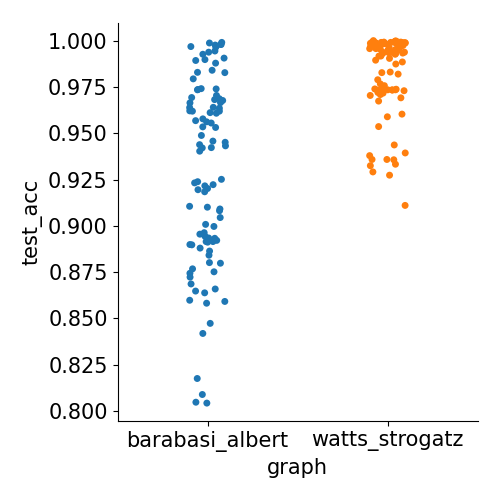}
	\caption[Performance of the WS and BA based random structure GRU]%
	{Plot comparing the performance of the Watts–Strogatz and Barabási–Albert model based random structure GRU.}
	\label{fig:gru_acc_comp}
\end{figure}

Next, the following table shows the Pearson correlation between test accuracy and different graph and recurrent network properties. The corresponding important plots are later shown in appendix \ref{app:rs_gru}.

\begin{table}[h]
	\centering
	\begin{tabular}{|l|c|}
	    \hline
		\textbf{Property} & \textbf{Correlation with $test\_acc$}\\
		\hline
		layers & 0.34\\
		nodes & \textbf{0.49}\\
		edges & \textbf{0.49}\\
		source\_nodes & \textbf{0.74}\\
		diameter & -0.20\\
		density & 0.34\\
		average\_shortest\_path\_length & -0.23\\
		eccentricity\_var & -0.21\\
		degree\_var & \textbf{-0.58}\\
		closeness\_var & \textbf{-0.67}\\
		nodes\_betweenness\_var & \textbf{-0.52}\\
		edge\_betweenness\_var & -0.26\\
		\hline
	\end{tabular}
	\caption[Pearson correlation between test accuracy of GRU and different graph and recurrent network properties]{Pearson correlation between test accuracy of GRU and different graph and recurrent network properties}
	\label{tab:gru_corr}
\end{table}

Positive Pearson correlation values in tables \ref{tab:tanh_corr}, \ref{tab:relu_corr}, \ref{tab:lstm_corr}, and \ref{tab:gru_corr} indicate that an increase in the property variable increases test accuracy. Similarly, a negative Pearson correlation in the same tables indicates that a decrease in the property variable decreases the test accuracy. A higher positive or negative correlation indicates a strong relationship between a graph property and test corresponding test accuracy.

\section{Performance prediction}

While training randomly structured recurrent networks, we create a small dataset by storing all the graph and recurrent network properties and their corresponding test accuracy. We train three regressor algorithms on this dataset, and based on the resulting R-squared value, we determine if it is possible to predict a randomly structured recurrent network's performance.

For each RNN variant, the following table shows the R-squared values from each regressor algorithm we used:

\begin{table}[h]
	\centering
	\begin{tabular}{|c||c|c|c|}
	    \hline
		\textbf{RNN Variant} & \textbf{Bayesian Ridge} & \textbf{Random Forest} & \textbf{AdaBoost}\\
		\hline
		RNN\_Tanh & 0.47919 & 0.43163 & 0.35698\\
		RNN\_ReLU & 0.36075 & 0.61504 & 0.53469\\
		LSTM & 0.37206 & 0.57933 & 0.59514\\
		GRU & 0.67224 & 0.87635 & 0.78313\\
		\hline
	\end{tabular}
	\caption[R-squared values from each regressor algorithm, for each RNN variant]{R-squared values from each regressor algorithm for each RNN variant help determine our data's closeness to the fitted regression line. All four RNN variants presented in this table are randomly structured.}
	\label{tab:r-squared}
\end{table}

In simple words, the R-squared value helps us to understand how close our data is to the fitted regression line. A higher R-square value usually means that our data is a better fit for the model.

In our case, RNN\_Tanh has below 0.5 R-squared value for all three regressors, while RNN\_ReLU has a good R-squared value with Random Forest regressor (0.61). In contrast to RNN\_ReLU, LSTM has a better R-squared value with AdaBoost regressor, while GRU has excellent R-squared values for all three regressor algorithms.

Based on these results, we further experimented with Random Forest regressor and data from randomly structured RNN\_ReLU and randomly structured GRU to extract feature importance scores of different properties under different circumstances. Results from this experiment are shown in appendix \ref{app:tables}.
\chapter{Discussion}\label{chap:discussion}

In this section, we discuss our findings from each experiment. Beginning from base model performance, we review the results of Sparse RNNs.

\section{Base model performance}

For our experiment purpose, we developed custom recurrent models using PyTorch, such that we can easily modify the weights based on our requirements. Each of our recurrent models, i.e., RNN with Tanh nonlinearity, RNN with ReLU nonlinearity, LSTM, and GRU, performs consistently with over 90\% accuracy after training for only 50 epochs, as shown in section \ref{section:base_perf}.

We used the same models to perform pruning experiments, a technique to generate sparsity in recurrent networks.

\section{Pruning recurrent networks}

This experiment helped answer the first three pruning-related research questions given in section \ref{section:research_questions}. 

Our pruning experiment was divided into three separate sub-experiments: pruning input-to-hidden and hidden-to-hidden weights simultaneously, pruning only input-to-hidden weights and pruning only hidden-to-hidden weights. For each sub-experiment, we also find the number of epochs required to regain the original performance.

While pruning both types of weights simultaneously, we found we can safely prune 80\% of RNN\_Tanh, 70\% of RNN\_ReLU, 60\% of LSTM, and 80\% of GRU. Afterward, we retrained these pruned models to find that for each RNN variant, we require only one epoch to regain the original performance, while these models never recover after pruning 100\% of weights.

While pruning only input-to-hidden weights, we found we can safely prune 70\% of RNN\_Tanh, 70\% of RNN\_ReLU, 70\% of LSTM, and 80\% of GRU. Afterward, we retrained these pruned models to find that for each RNN variant, we mostly require only one epoch to regain the original performance and just two epochs in the case of RNN\_Tanh with 80\% pruning. These models never recover after pruning 100\% of the weights.

While pruning only hidden-to-hidden weights, we found we can safely prune 80\% of RNN\_Tanh, 70\% of RNN\_ReLU, 90\% of LSTM, and 90\% of GRU. Afterward, we retrained these pruned models to find that for each RNN variant, we require only one epoch to regain the original performance. RNN\_Tanh and RNN\_ReLU models never recover after pruning 100\% of the weights, while LSTM and GRU still regain the original performance even with 100\% pruning of hidden-to-hidden weights.

\section{Randomly structured recurrent networks}

This experiment helped answer the remaining two research questions given in section \ref{section:research_questions}.

The resulting Pearson correlation values from training randomly structured recurrent networks help identify important graph and recurrent network properties for each RNN variant.

Based on this correlation, we found closeness\_var, nodes\_betweenness\_var and the number of nodes to be essential properties for randomly structured RNN\_Tanh. For randomly structured RNN\_ReLU, the essential properties are the number of nodes, the number of edges, the number of source nodes, and nodes\_betweenness\_var.

In the case of randomly structured LSTM, we found six essential properties, i.e., the number of nodes, the number of edges, the number of source nodes, closeness\_var, nodes\_betweenness\_var, and edge\_betweenness\_var. Similarly, we found six essential properties for randomly structured GRU, namely, the number of nodes, the number of edges, the number of source nodes, degree\_var, closeness\_var, and nodes\_betweenness\_var.

Next, we trained three regressor algorithms to find if we can use graph properties of a base random graph for performance prediction based on how well our data fit a model. Based on the results, we found that RNN\_Tanh has below 0.5 R-squared value for each regressor, meaning our data is a weak fit to all three regression models.

In the case of RNN\_ReLU,  we have a 0.61 R-squared value with Random Forest regressor, meaning our data is slightly moderate fit to this particular regressor. For LSTM, our model is again slightly moderate fit to AdaBoost regressor with ~0.60 R-squared value.

GRU has the best R-squared value of 0.81 with Random Forest regressor, indicating our data from randomly structured GRU is a strong fit to the Random Forest regressor model.
\chapter{Conclusion}\label{chap:conclusion}

The main objective of our thesis, as described in chapter \ref{chap:introduction}, was to investigate the effects of sparsity in recurrent neural networks. In chapter \ref{chap:experiments}, we thoroughly explained the methodology followed, with results presented in chapter \ref{chap:results}. In this chapter, we briefly summarize our outcomes.

We followed two different methods to induce sparsity in recurrent networks. One method introduced sparsity by pruning a certain percent of input-to-hidden and hidden-to-hidden weights, and another method is to generate sparse structures based on random graphs.

Based on the results of pruning experiments, we conclude that it is possible to reduce the weight complexity of different RNN variants by more than 60\%. Our experiments also confirmed that it is possible to regain the accuracy of a pruned model with only one epoch in most cases. One big difference between RNN\_Tanh, RNN\_ReLU, and LSTM, GRU is that LSTM and GRU can regain the original performance even with 100\% hidden-to-hidden weight pruning.

Random structure experiments helped to identify the essential graph properties of a base random graph. In all four RNN variants we experimented with, two mutual graph properties are the number of nodes and node betweenness. The two most mutual and important graph properties are the number of edges and the number of source nodes.

The performance prediction experiment results identified that data from RNN\_Tanh has a weak fit with all three regressors used. Data from RNN\_ReLU and GRU has moderate and strong fit Random Forest regressor, respectively. Finally,  LSTM has a moderate fit with AdaBoost regressor.

\begin{appendices}
\chapter{Code}

\begin{code}
\captionof{listing}{Calculating threshold for pruning}
\label{code:threshold}
\begin{lstlisting}
weights = []
for param, data in self.named_parameters():
    if 'bias' not in param and key in param:
        weights += list(data.cpu().data.abs().numpy().flatten())
threshold = np.percentile(np.array(weights), percent)
\end{lstlisting}
\end{code}

\begin{code}
\captionof{listing}{Calculating binary mask}
\label{code:mask}
\begin{lstlisting}
masks = {}
for l, layer in enumerate(self.recurrent_layers):
    masks[l] = []
    for param, data in layer.named_parameters():
        if 'bias' not in param and key in param:
            mask = torch.ones(data.shape,
                              dtype=torch.bool,
                              device=data.device)
            mask[torch.where(abs(data) < threshold)] = False
            masks[l].append(mask)
\end{lstlisting}
\end{code}

\chapter{Tables}\label{app:tables}

\begin{table}[h]
	\centering
	\begin{tabular}{|l|c|c|c|c|}
	    \hline
		 & \shortstack{\textbf{All}\\\textbf{ properties}} & \shortstack{\textbf{Only nodes}\\\textbf{ and edges}} & \shortstack{\textbf{Without nodes}\\\textbf{ and edges}} & \shortstack{\textbf{Only}\\\textbf{ variances}}\\
		\hline
		Random Forest $R^{2}$ & 0.61 & 0.71 & -0.05 & 0.06\\
		\hline\hline
		\textbf{Property} & \multicolumn{4}{|l|}{\textbf{Feature importance}}\\
		\hline
		layers & 0.04 & & 0.06 & \\
        nodes & 0.02 & 0.23 & & \\
        edges & 0.01 & 0.29 & & \\
        source\_nodes & 0.08 & 0.29 & & \\
        sink\_nodes & 0.11 & 0.19 & & \\
        diameter & 0.0 & & 0.0 & \\
        density & 0.03 & & 0.03 & \\
        average\_shortest\_path\_length & 0.03 & & 0.03 & \\
        eccentricity\_mean & 0.03 & & 0.04 & \\
        eccentricity\_var & 0.04 & & 0.06 & 0.21 \\
        eccentricity\_std & 0.05 & & 0.09 & \\
        degree\_mean & 0.03 & & 0.03 & \\
        degree\_var & 0.03 & & 0.06 & 0.17 \\
        degree\_std & 0.03 & & 0.04 & \\
        closeness\_mean & 0.04 & & 0.05 & \\
        closeness\_var & 0.07 & & 0.09 & 0.22 \\
        closeness\_std & 0.07 & & 0.08 & \\
        nodes\_betweenness\_mean & 0.05 & & 0.08 & \\
        nodes\_betweenness\_var & 0.08 & & 0.08 & 0.24 \\
        nodes\_betweenness\_std & 0.07 & & 0.07 & \\
        edge\_betweenness\_mean & 0.05 & & 0.06 & \\
        edge\_betweenness\_var & 0.02 & & 0.03 & 0.16 \\
        edge\_betweenness\_std & 0.02 & & 0.02 & \\
		\hline
	\end{tabular}
	\caption[RNN\_ReLU - Feature importance scores for the Random Forest regressor under different circumstances]{Structural properties of the randomly structured RNN\_ReLU's base random graphs and their feature importance scores for the Random Forest regressor under different circumstances.}
	\label{tab:relu_fis}
\end{table}

\begin{table}[h]
	\centering
	\begin{tabular}{|l|c|c|c|c|}
	    \hline
		 & \shortstack{\textbf{All}\\\textbf{ properties}} & \shortstack{\textbf{Only nodes}\\\textbf{ and edges}} & \shortstack{\textbf{Without nodes}\\\textbf{ and edges}} & \shortstack{\textbf{Only}\\\textbf{ variances}}\\
		\hline
		Random Forest $R^{2}$ & 0.87 & 0.75 & 0.78 & 0.76\\
		\hline\hline
		\textbf{Property} & \multicolumn{4}{|l|}{\textbf{Feature importance}}\\
		\hline
		layers & 0.01 & & 0.04 & \\
        nodes & 0.01 & 0.11 & & \\
        edges & 0.01 & 0.12 & & \\
        source\_nodes & 0.67 & 0.71 & & \\
        sink\_nodes & 0.01 & 0.06 & & \\
        diameter & 0.0 & & 0.01 & \\
        density & 0.01 & & 0.02 & \\
        average\_shortest\_path\_length & 0.01 & & 0.02 & \\
        eccentricity\_mean & 0.01 & & 0.02 & \\
        eccentricity\_var & 0.02 & & 0.03 & 0.17 \\
        eccentricity\_std & 0.02 & & 0.03 & \\
        degree\_mean & 0.01 & & 0.03 & \\
        degree\_var & 0.01 & & 0.03 & 0.15 \\
        degree\_std & 0.01 & & 0.03 & \\
        closeness\_mean & 0.01 & & 0.03 & \\
        closeness\_var & 0.02 & & 0.11 & 0.36 \\
        closeness\_std & 0.01 & & 0.16 & \\
        nodes\_betweenness\_mean & 0.04 & & 0.11 & \\
        nodes\_betweenness\_var & 0.01 & & 0.15 & 0.25 \\
        nodes\_betweenness\_std & 0.02 & & 0.14 & \\
        edge\_betweenness\_mean & 0.03 & & 0.03 & \\
        edge\_betweenness\_var & 0.01 & & 0.02 & 0.08 \\
        edge\_betweenness\_std & 0.01 & & 0.02 & \\
		\hline
	\end{tabular}
	\caption[GRU - Feature importance scores for the Random Forest regressor under different circumstances]{Structural properties of the randomly structured GRU's base random graphs and their feature importance scores for the Random Forest regressor under different circumstances.}
	\label{tab:gru_fis}
\end{table}


\chapter{Plots}

\section{Randomly structured RNN\_Tanh}\label{app:rs_tanh}

\begin{figure}[H]
    \centering
    \begin{subfigure}{0.4\textwidth}
        \includegraphics[width=\linewidth]{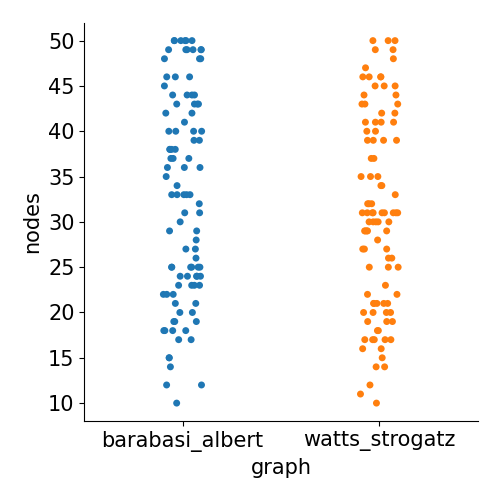}
        \caption{} \label{fig:tanh_graph_nodes}
    \end{subfigure}%
    \hfill
    \begin{subfigure}{0.4\textwidth}
        \includegraphics[width=\linewidth]{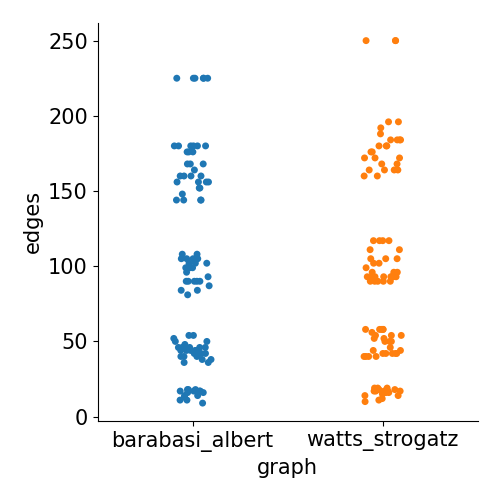}
        \caption{} \label{fig:tanh_graph_edges}
    \end{subfigure}%
  
    \bigskip
    \begin{subfigure}{0.4\textwidth}
        \includegraphics[width=\linewidth]{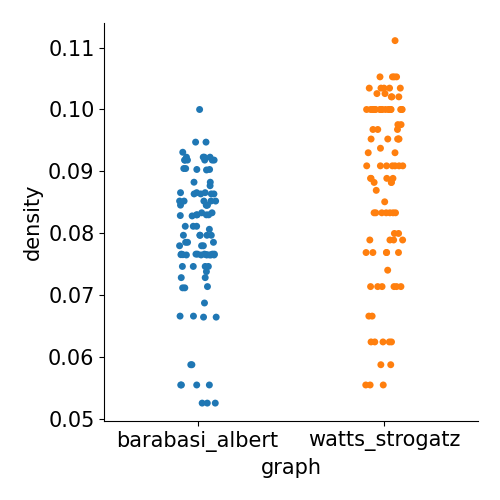}
        \caption{} \label{fig:tanh_graph_density}
    \end{subfigure}
    \hfill
    \begin{subfigure}{0.4\textwidth}
        \includegraphics[width=\linewidth]{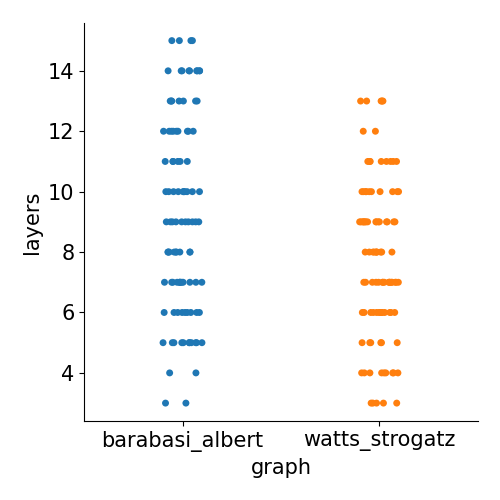}
        \caption{} \label{fig:tanh_graph_layers}
    \end{subfigure}

\caption[Comparison of basic graph properties and number of layers in WS and BA based RNN\_Tanh models]{Comparison of basic graph properties and number of layers in Watts–Strogatz and Barabási–Albert based RNN\_Tanh models} \label{fig:tanh_graphs}
\end{figure}

\begin{figure}[H]
    \centering
    \begin{subfigure}{0.45\textwidth}
        \includegraphics[width=\linewidth]{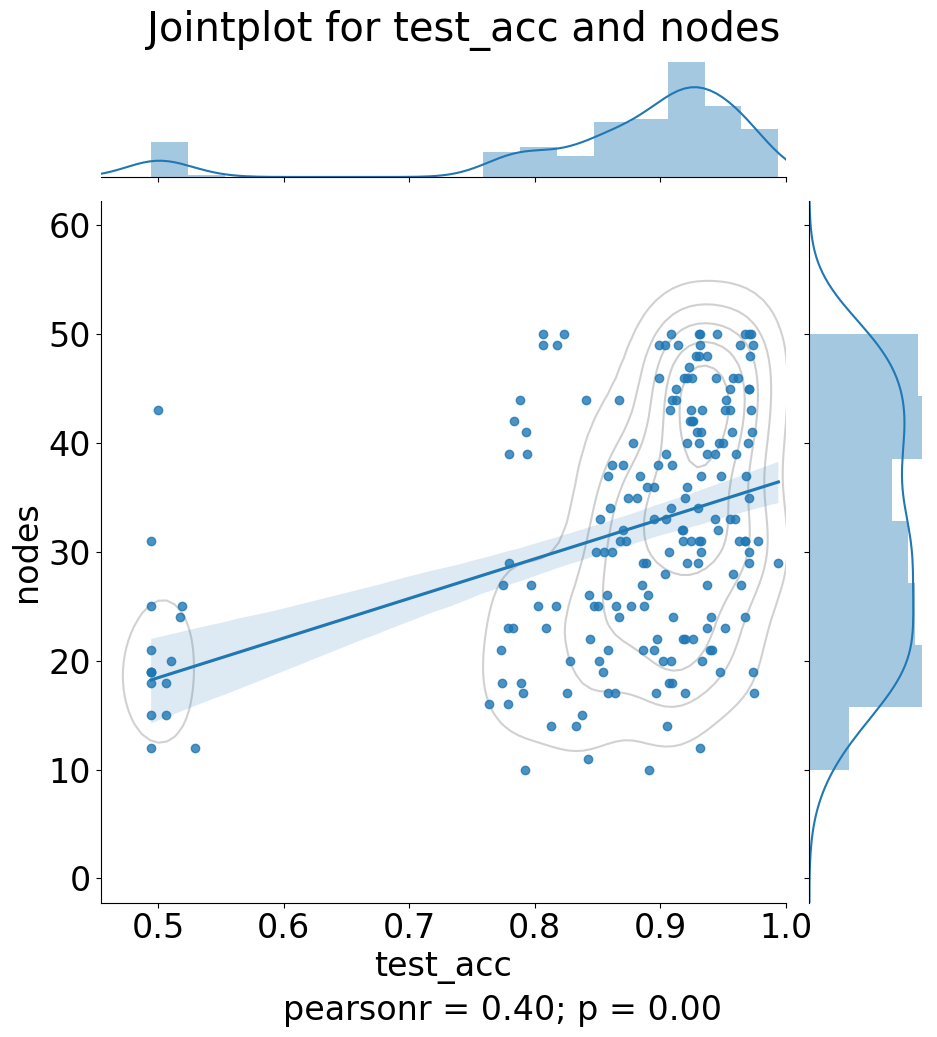}
        \caption{Correlation between test\_acc and the number of nodes} \label{fig:jp_tanh_node}
    \end{subfigure}%
    \hfill
    \begin{subfigure}{0.45\textwidth}
        \includegraphics[width=\linewidth]{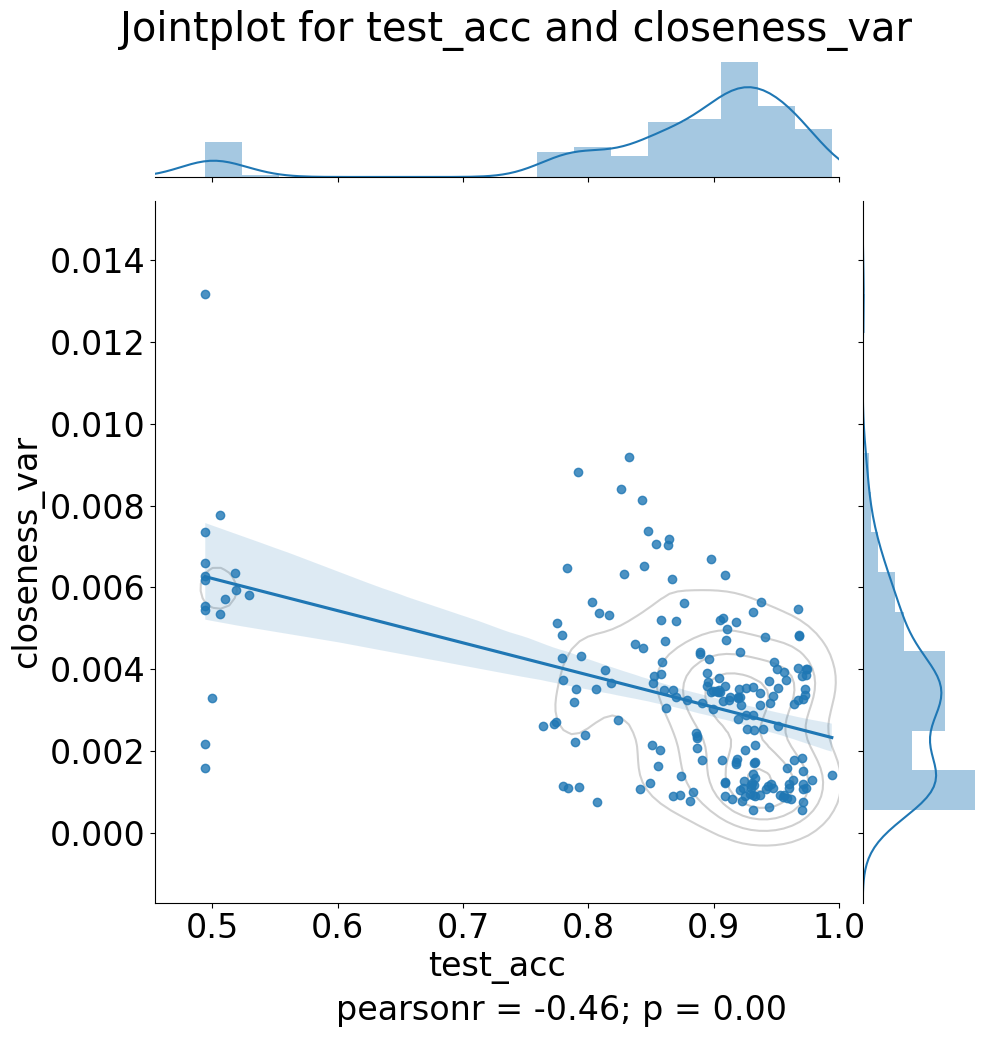}
        \caption{Correlation between test\_acc and closeness in base random graph} \label{fig:jp_tanh_close}
    \end{subfigure}%
  
    \bigskip
    \begin{subfigure}{0.45\textwidth}
        \includegraphics[width=\linewidth]{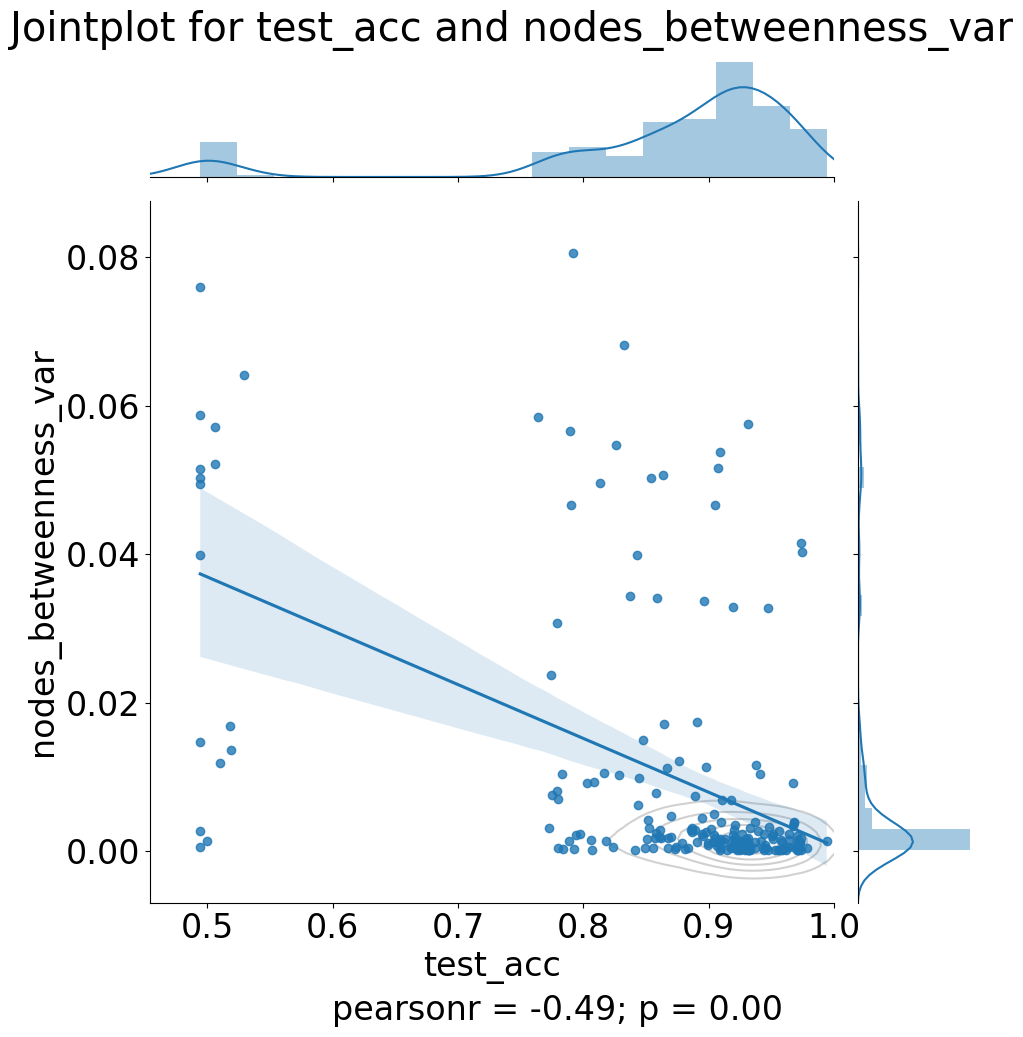}
        \caption{Correlation between test\_acc and node betweenness in base random graph} \label{fig:jp_tanh_between}
    \end{subfigure}
    \hfill
    \begin{subfigure}{0.45\textwidth}
        \includegraphics[width=\linewidth]{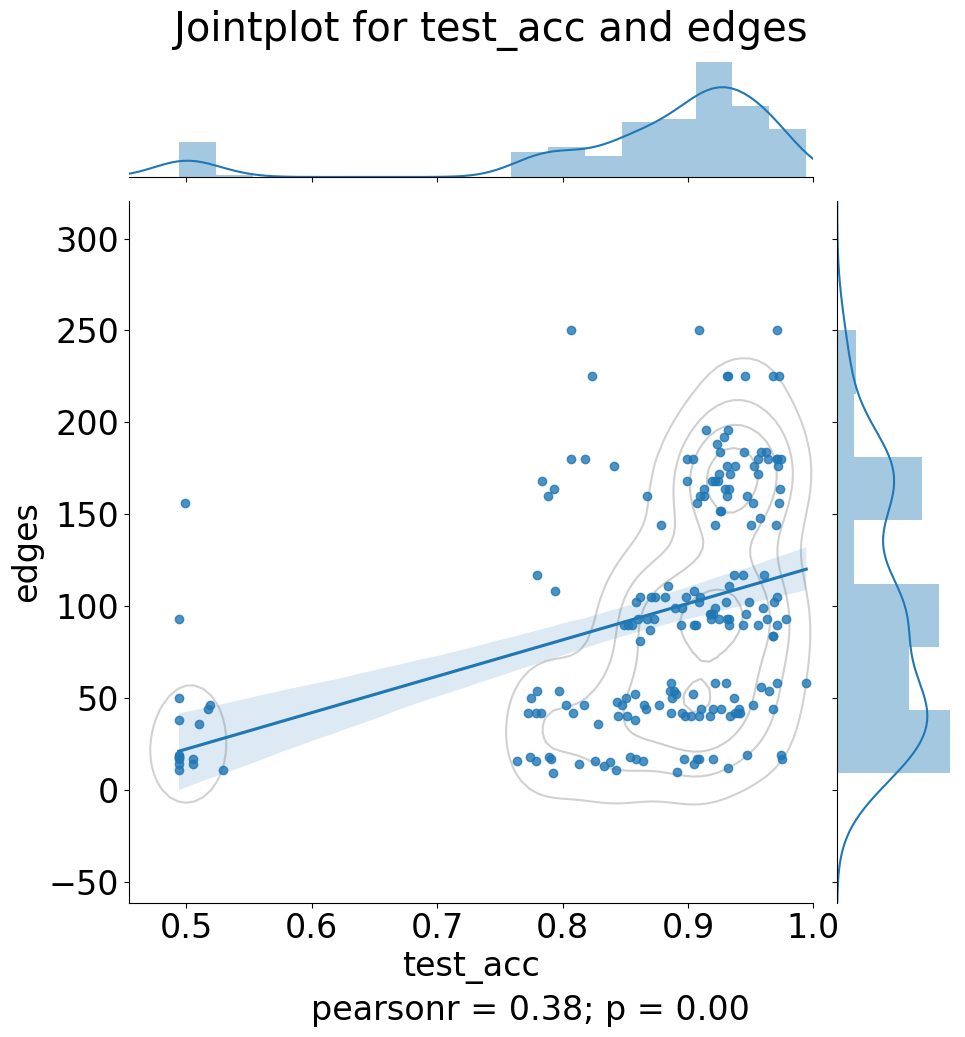}
        \caption{Correlation between test\_acc and the number of edges} \label{fig:jp_tanh_edge}
    \end{subfigure}%

\caption[Correlation between test accuracy of RNN\_Tanh and its different graph and recurrent network properties - 1]{Correlation between test accuracy of RNN\_Tanh and its different graph and recurrent network properties} \label{fig:tanh_correlation}
\end{figure}

\newpage
\section{Randomly structured RNN\_ReLU}\label{app:rs_relu}

\begin{figure}[H]
    \centering
    \begin{subfigure}{0.45\textwidth}
        \includegraphics[width=\linewidth]{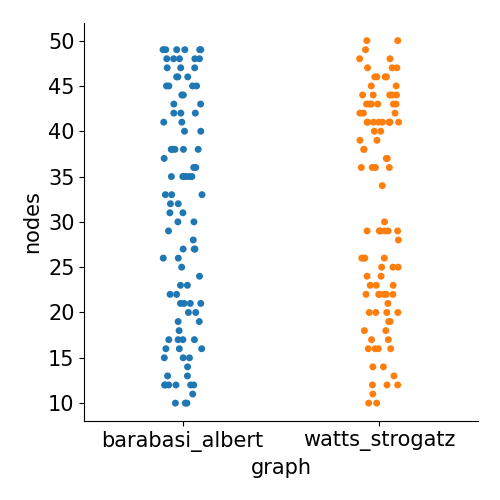}
        \caption{} \label{fig:relu_graph_nodes}
    \end{subfigure}%
    \hfill
    \begin{subfigure}{0.45\textwidth}
        \includegraphics[width=\linewidth]{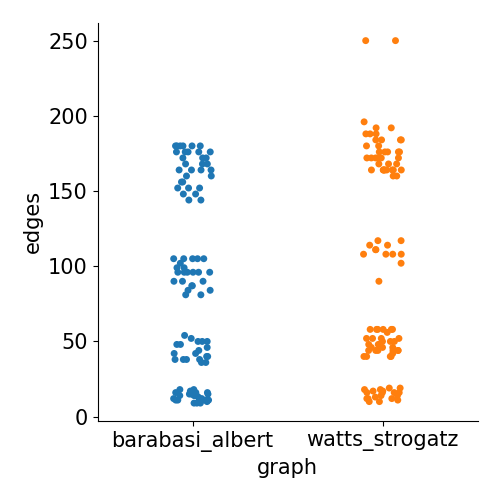}
        \caption{} \label{fig:relu_graph_edges}
    \end{subfigure}%
  
    \bigskip
    \begin{subfigure}{0.45\textwidth}
        \includegraphics[width=\linewidth]{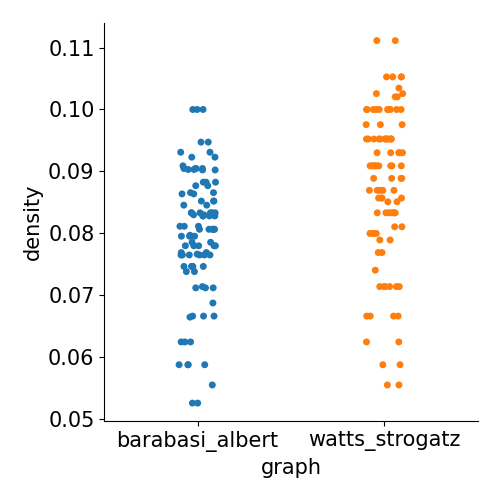}
        \caption{} \label{fig:relu_graph_density}
    \end{subfigure}
    \hfill
    \begin{subfigure}{0.45\textwidth}
        \includegraphics[width=\linewidth]{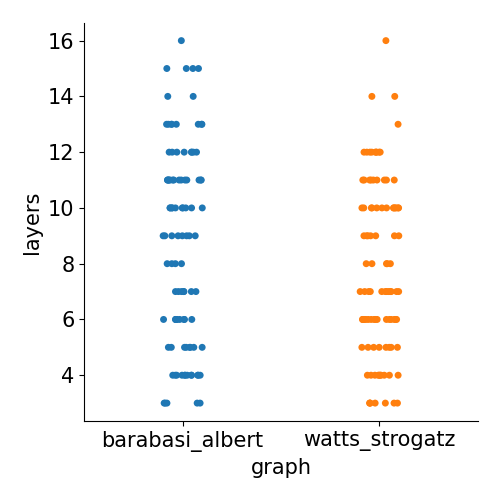}
        \caption{} \label{fig:relu_graph_layers}
    \end{subfigure}

\caption[Comparison of basic graph properties and number of layers in WS and BA based RNN\_ReLU models]{Comparison of basic graph properties and number of layers in Watts–Strogatz and Barabási–Albert based RNN\_ReLU models} \label{fig:relu_graphs}
\end{figure}

\begin{figure}[H]
    \centering
    \begin{subfigure}{0.45\textwidth}
        \includegraphics[width=\linewidth]{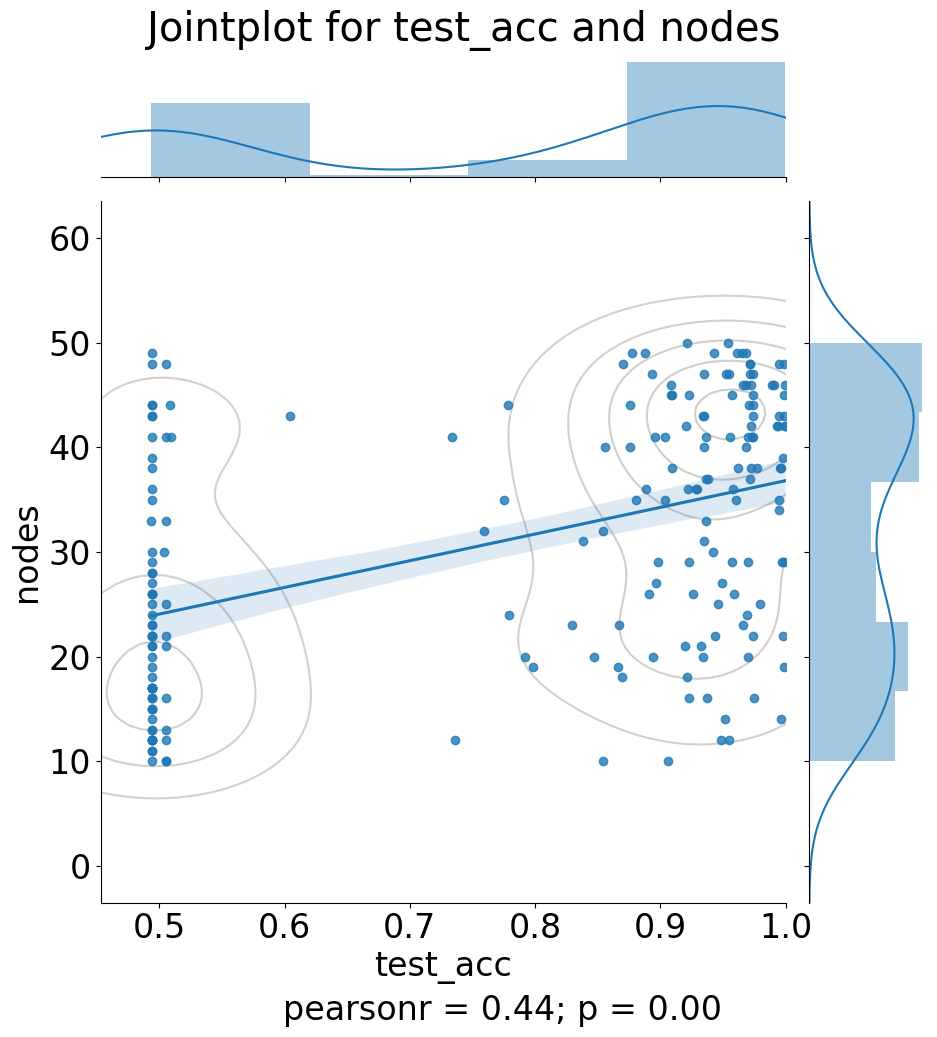}
        \caption{Correlation between test\_acc and the number of nodes} \label{fig:jp_relu_node}
    \end{subfigure}%
    \hfill
    \begin{subfigure}{0.45\textwidth}
        \includegraphics[width=\linewidth]{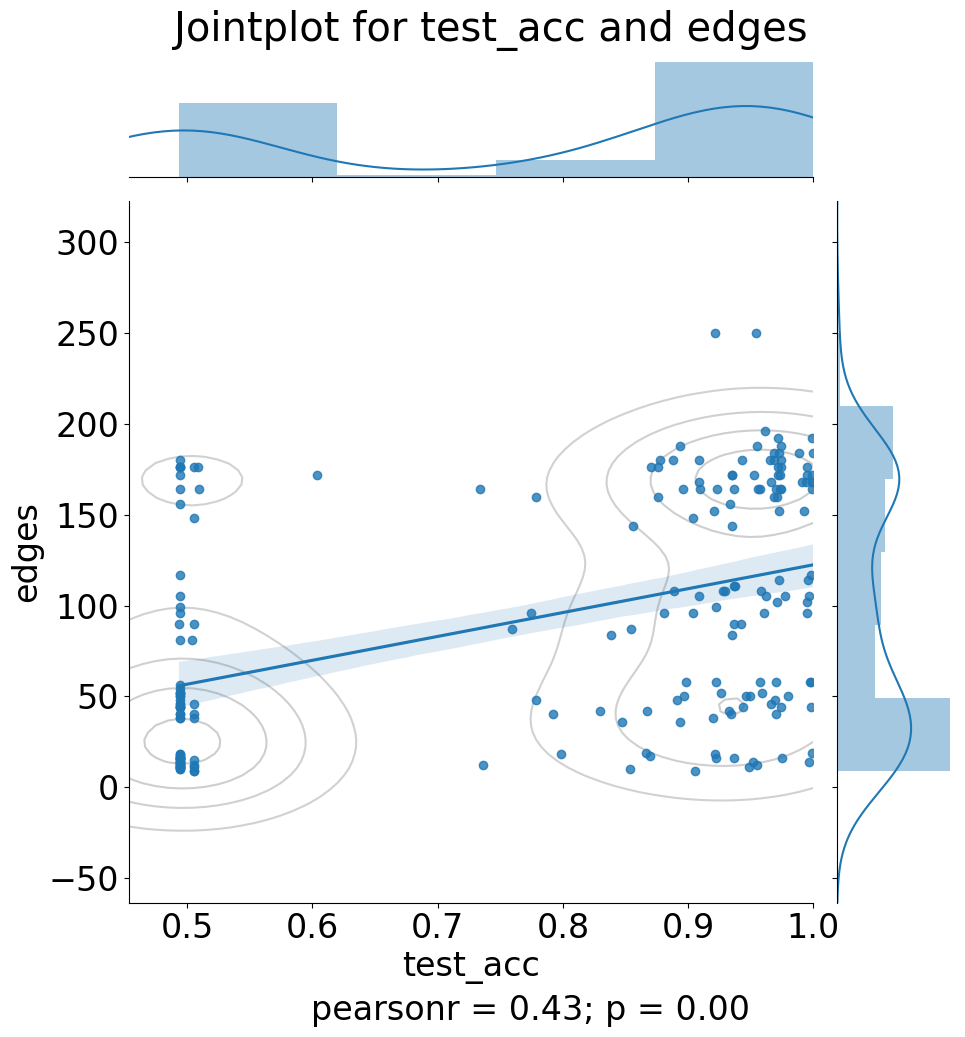}
        \caption{Correlation between test\_acc and the number of edges} \label{fig:jp_relu_edge}
    \end{subfigure}%
  
    \bigskip
    \begin{subfigure}{0.45\textwidth}
        \includegraphics[width=\linewidth]{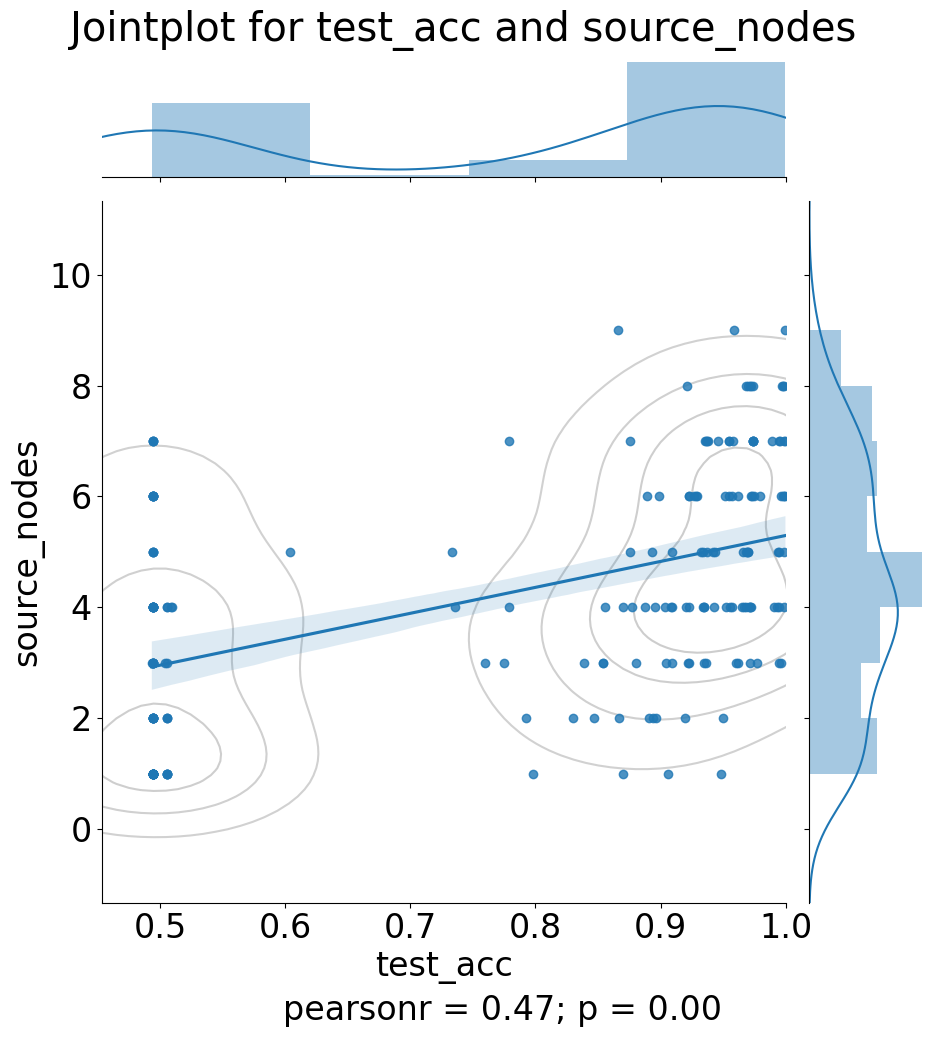}
        \caption{Correlation between test\_acc and the number of source nodes} \label{fig:jp_relu_source}
    \end{subfigure}
    \hfill
    \begin{subfigure}{0.45\textwidth}
        \includegraphics[width=\linewidth]{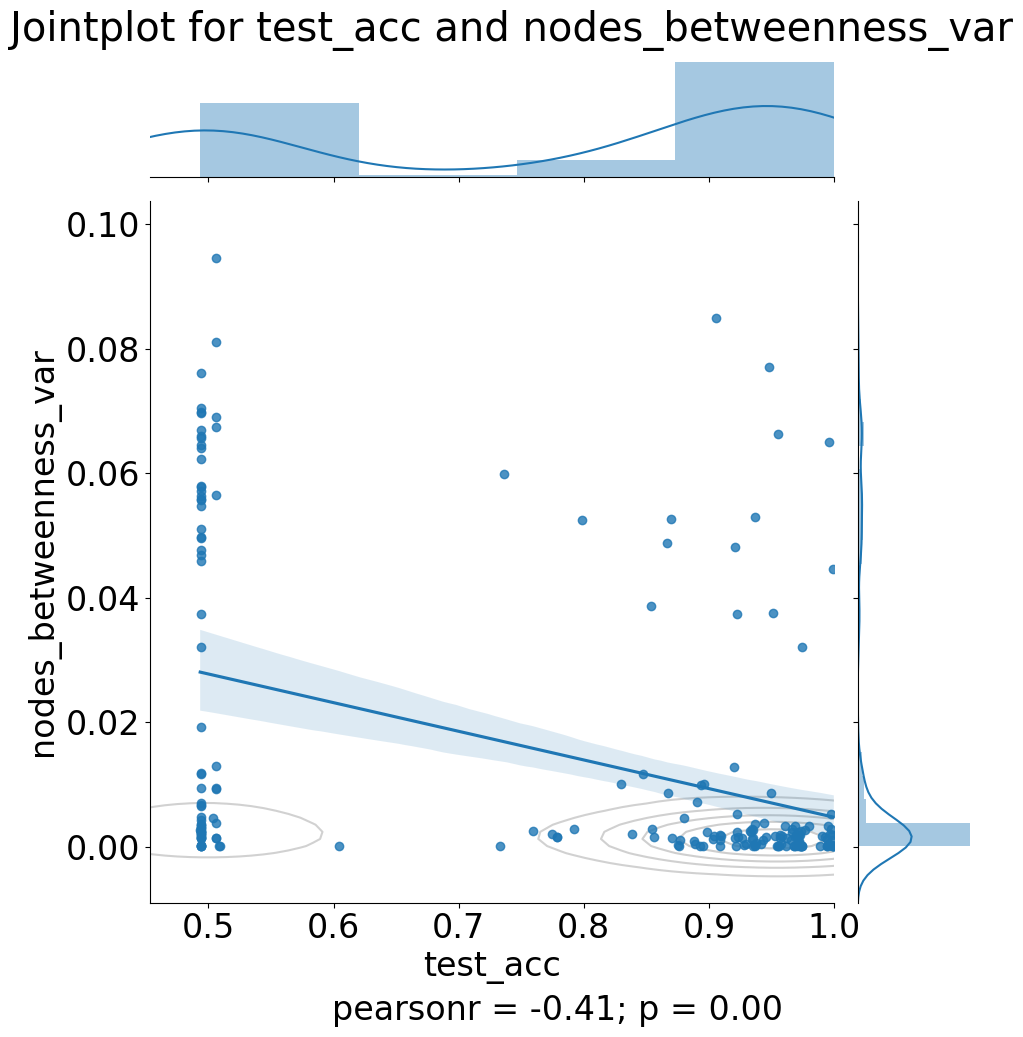}
        \caption{Correlation between test\_acc and node betweenness in base random graph} \label{fig:jp_relu_between}
    \end{subfigure}

\caption[Correlation between test accuracy of RNN\_ReLU and its different graph and recurrent network properties - 1]{Correlation between test accuracy of RNN\_ReLU and its different graph and recurrent network properties} \label{fig:relu_correlation}
\end{figure}

\newpage
\section{Randomly structured LSTM}\label{app:rs_lstm}

\begin{figure}[H]
    \centering
    \begin{subfigure}{0.45\textwidth}
        \includegraphics[width=\linewidth]{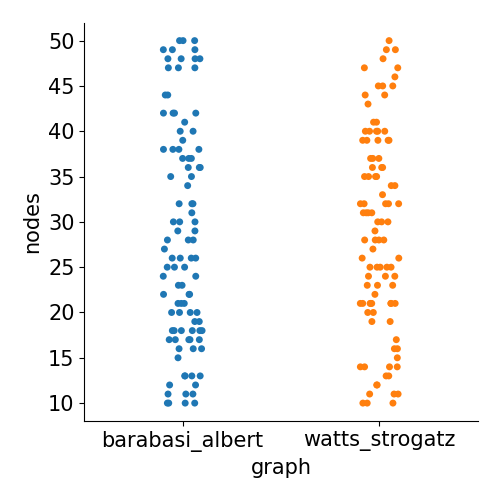}
        \caption{} \label{fig:lstm_graph_nodes}
    \end{subfigure}%
    \hfill
    \begin{subfigure}{0.45\textwidth}
        \includegraphics[width=\linewidth]{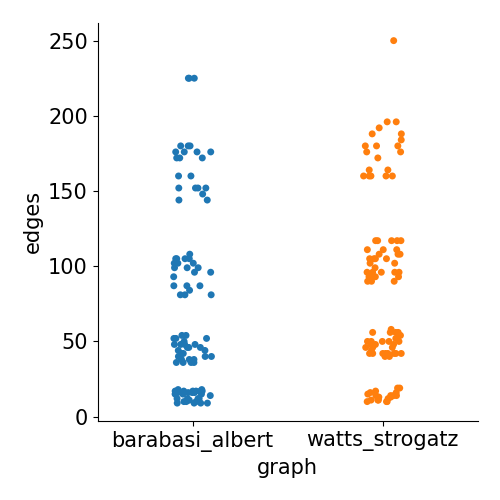}
        \caption{} \label{fig:lstm_graph_edges}
    \end{subfigure}%
  
    \bigskip
    \begin{subfigure}{0.45\textwidth}
        \includegraphics[width=\linewidth]{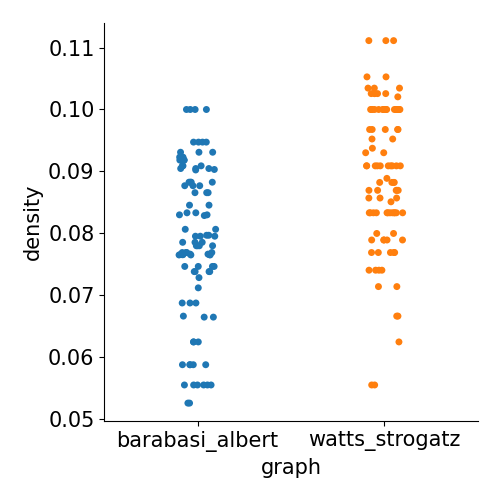}
        \caption{} \label{fig:lstm_graph_density}
    \end{subfigure}
    \hfill
    \begin{subfigure}{0.45\textwidth}
        \includegraphics[width=\linewidth]{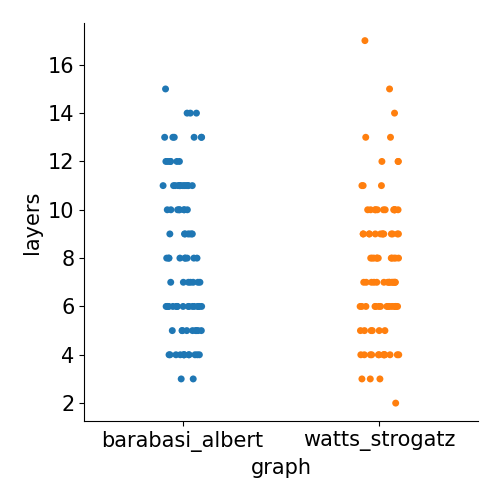}
        \caption{} \label{fig:lstm_graph_layers}
    \end{subfigure}

\caption[Comparison of basic graph properties and number of layers in WS and BA based LSTM models]{Comparison of basic graph properties and number of layers in Watts–Strogatz and Barabási–Albert based LSTM models} \label{fig:lstm_graphs}
\end{figure}

\begin{figure}[H]
    \centering
    \begin{subfigure}{0.45\textwidth}
        \includegraphics[width=\linewidth]{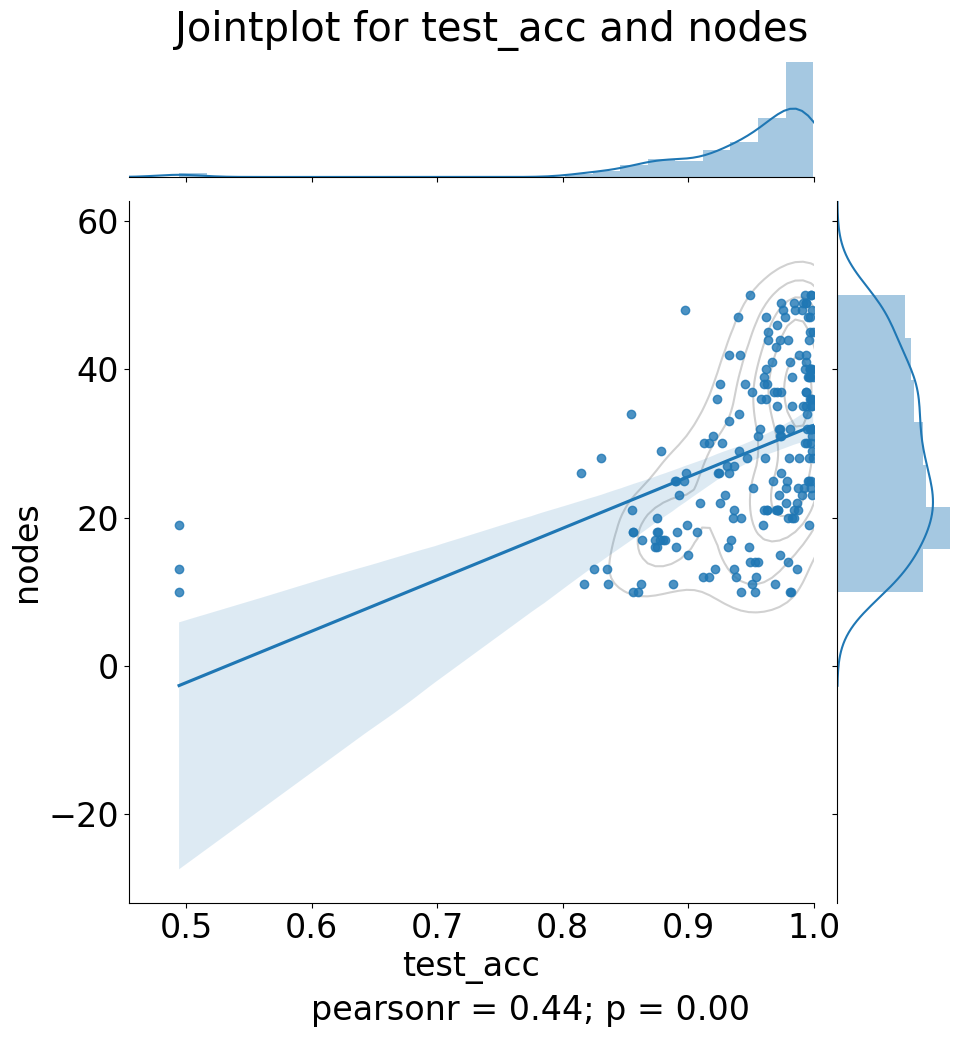}
        \caption{Correlation between test\_acc and the number of nodes} \label{fig:jp_lstm_node}
    \end{subfigure}%
    \hfill
    \begin{subfigure}{0.45\textwidth}
        \includegraphics[width=\linewidth]{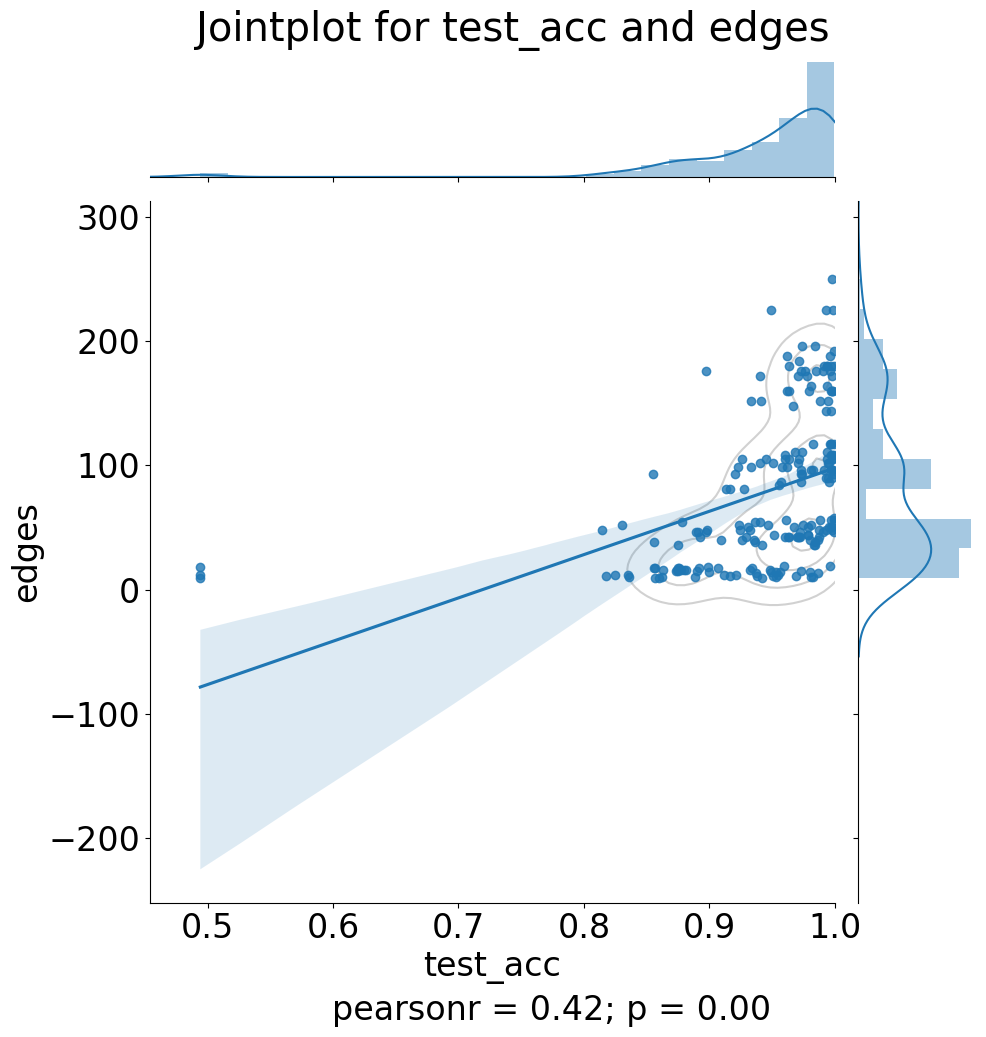}
        \caption{Correlation between test\_acc and the number of edges} \label{fig:jp_lstm_edge}
    \end{subfigure}%
  
    \bigskip
    \begin{subfigure}{0.45\textwidth}
        \includegraphics[width=\linewidth]{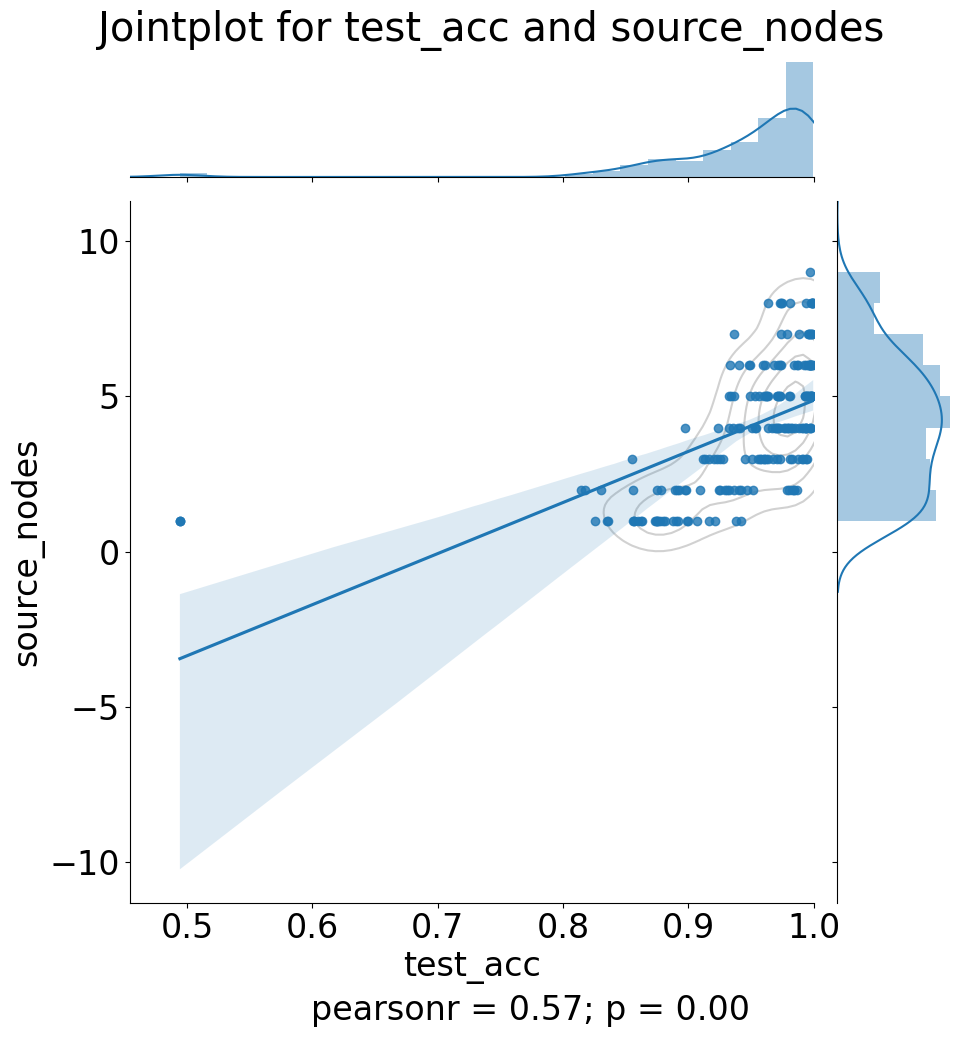}
        \caption{Correlation between test\_acc and the number of source nodes} \label{fig:jp_lstm_source}
    \end{subfigure}
    \hfill
    \begin{subfigure}{0.45\textwidth}
        \includegraphics[width=\linewidth]{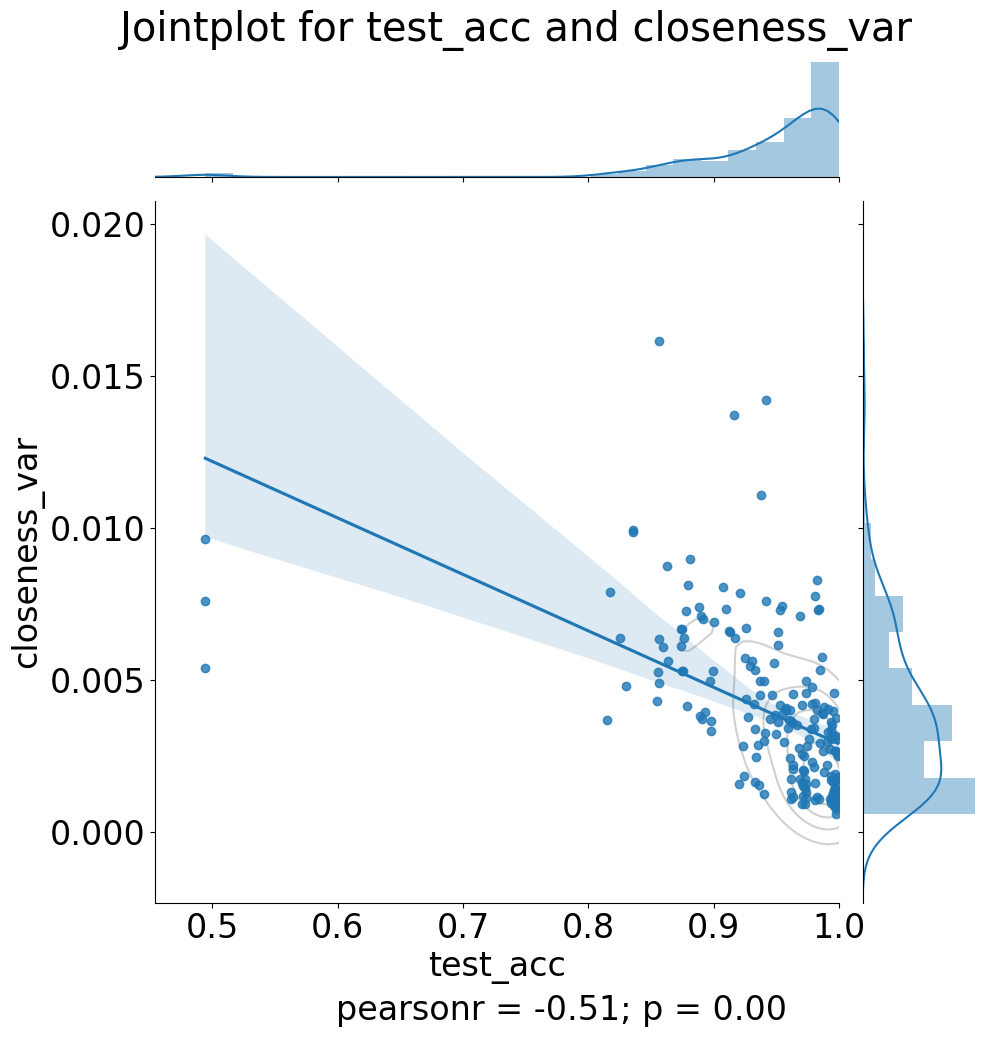}
        \caption{Correlation between test\_acc and closeness in base random graph} \label{fig:jp_lstm_close}
    \end{subfigure}

\caption[Correlation between test accuracy of LSTM and its different graph and recurrent network properties - 1]{Correlation between test accuracy of LSTM and its different graph and recurrent network properties - 1} \label{fig:lstm_correlation_1}
\end{figure}

\begin{figure}[H]
    \centering
    \begin{subfigure}{0.45\textwidth}
        \includegraphics[width=\linewidth]{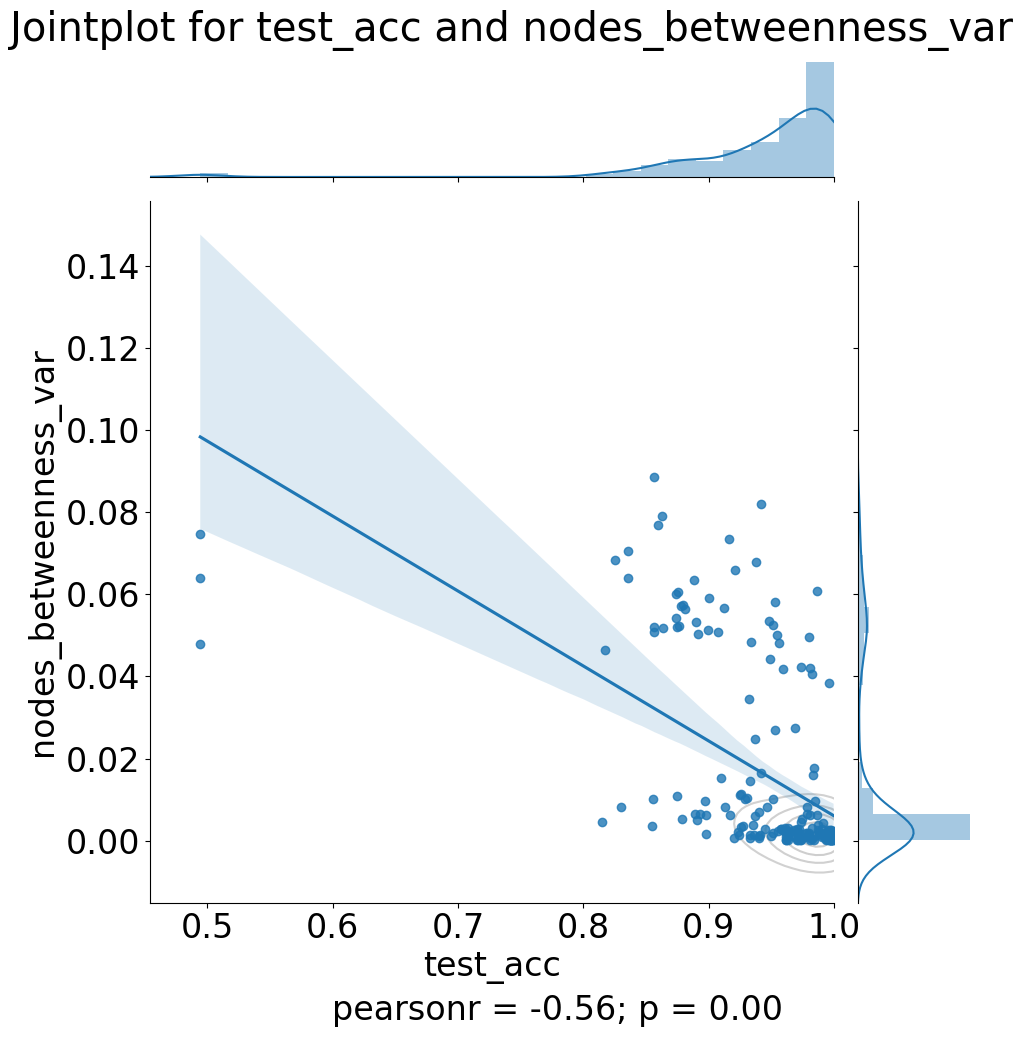}
        \caption{Correlation between test\_acc and node betweenness in base random graph} \label{fig:jp_lstm_node_bn}
    \end{subfigure}
    \hfill
    \begin{subfigure}{0.45\textwidth}
        \includegraphics[width=\linewidth]{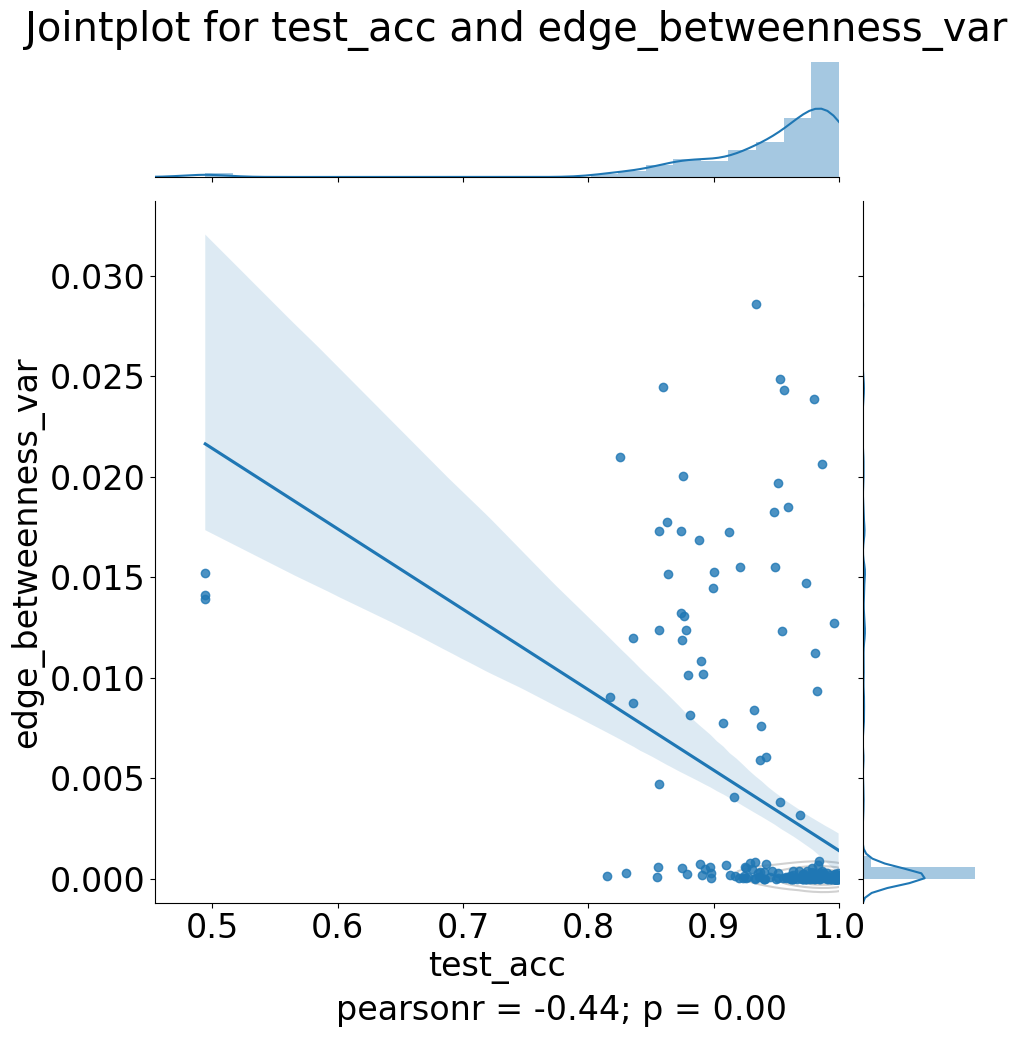}
        \caption{Correlation between test\_acc and edge betweenness in base random graph} \label{fig:jp_lstm_edge_bn}
    \end{subfigure}

\caption[Correlation between test accuracy of LSTM and its different graph and recurrent network properties - 2]{Correlation between test accuracy of LSTM and its different graph and recurrent network properties - 2} \label{fig:lstm_correlation_2}
\end{figure}

\newpage
\section{Randomly structured GRU}\label{app:rs_gru}

\begin{figure}[H]
    \centering
    \begin{subfigure}{0.45\textwidth}
        \includegraphics[width=\linewidth]{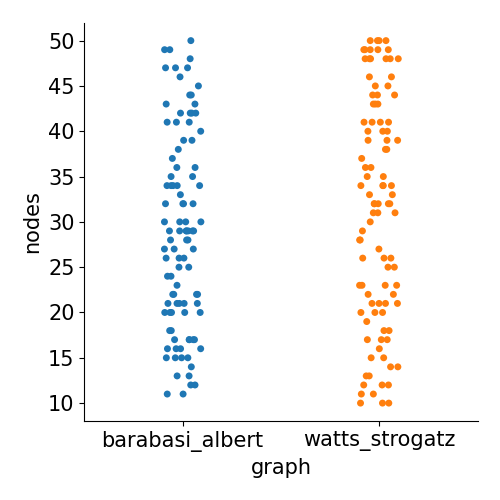}
        \caption{} \label{fig:gru_graph_nodes}
    \end{subfigure}%
    \hfill
    \begin{subfigure}{0.45\textwidth}
        \includegraphics[width=\linewidth]{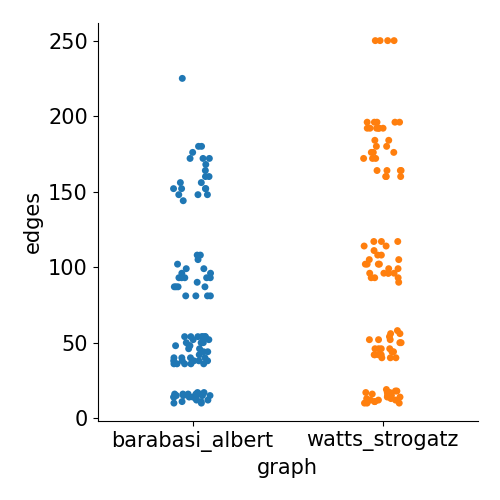}
        \caption{} \label{fig:gru_graph_edges}
    \end{subfigure}%
  
    \bigskip
    \begin{subfigure}{0.45\textwidth}
        \includegraphics[width=\linewidth]{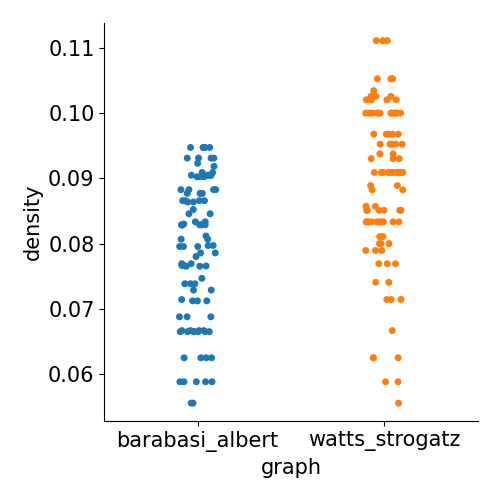}
        \caption{} \label{fig:gru_graph_density}
    \end{subfigure}
    \hfill
    \begin{subfigure}{0.45\textwidth}
        \includegraphics[width=\linewidth]{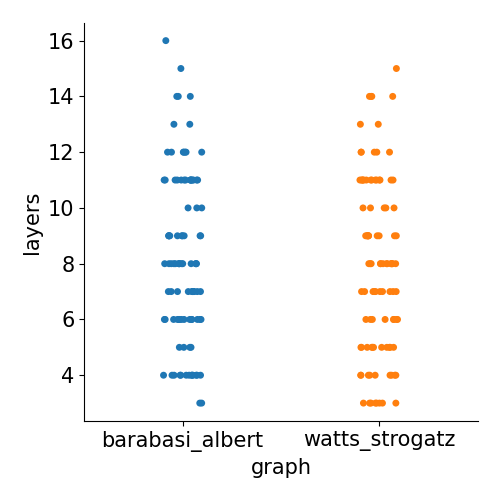}
        \caption{} \label{fig:gru_graph_layers}
    \end{subfigure}

\caption[Comparison of basic graph properties and number of layers in WS and BA based GRU models]{Comparison of basic graph properties and number of layers in Watts–Strogatz and Barabási–Albert based GRU models} \label{fig:gru_graphs}
\end{figure}

\begin{figure}[H]
    \centering
    \begin{subfigure}{0.45\textwidth}
        \includegraphics[width=\linewidth]{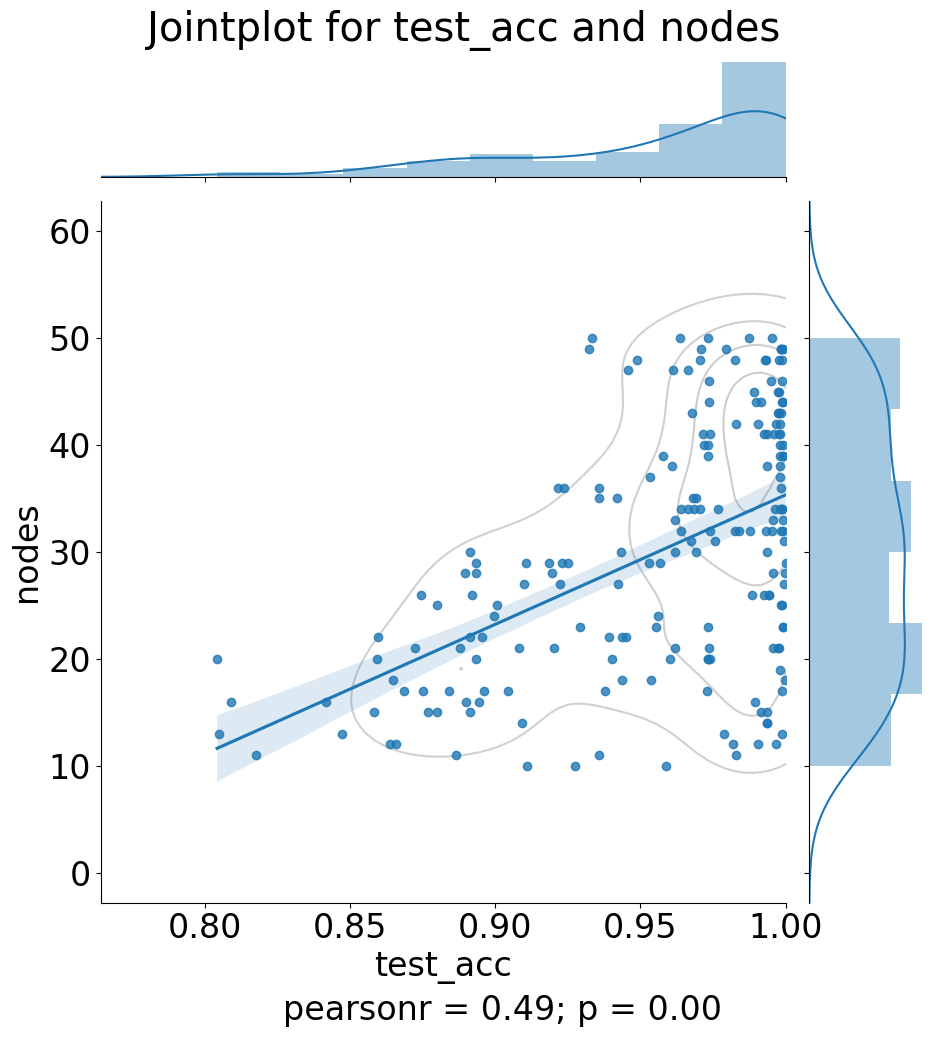}
        \caption{Correlation between test\_acc and the number of nodes} \label{fig:jp_gru_node}
    \end{subfigure}%
    \hfill
    \begin{subfigure}{0.45\textwidth}
        \includegraphics[width=\linewidth]{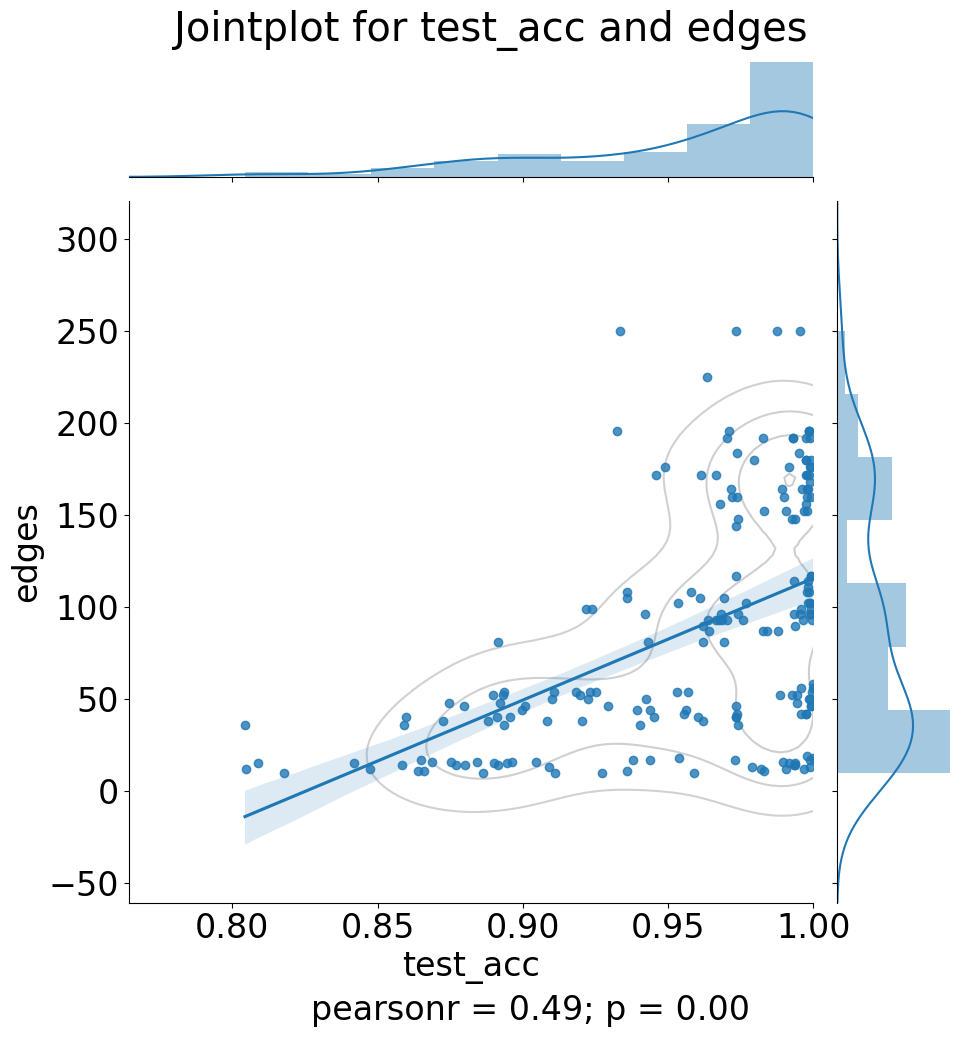}
        \caption{Correlation between test\_acc and the number of edges} \label{fig:jp_gru_edge}
    \end{subfigure}%
  
    \bigskip
    \begin{subfigure}{0.45\textwidth}
        \includegraphics[width=\linewidth]{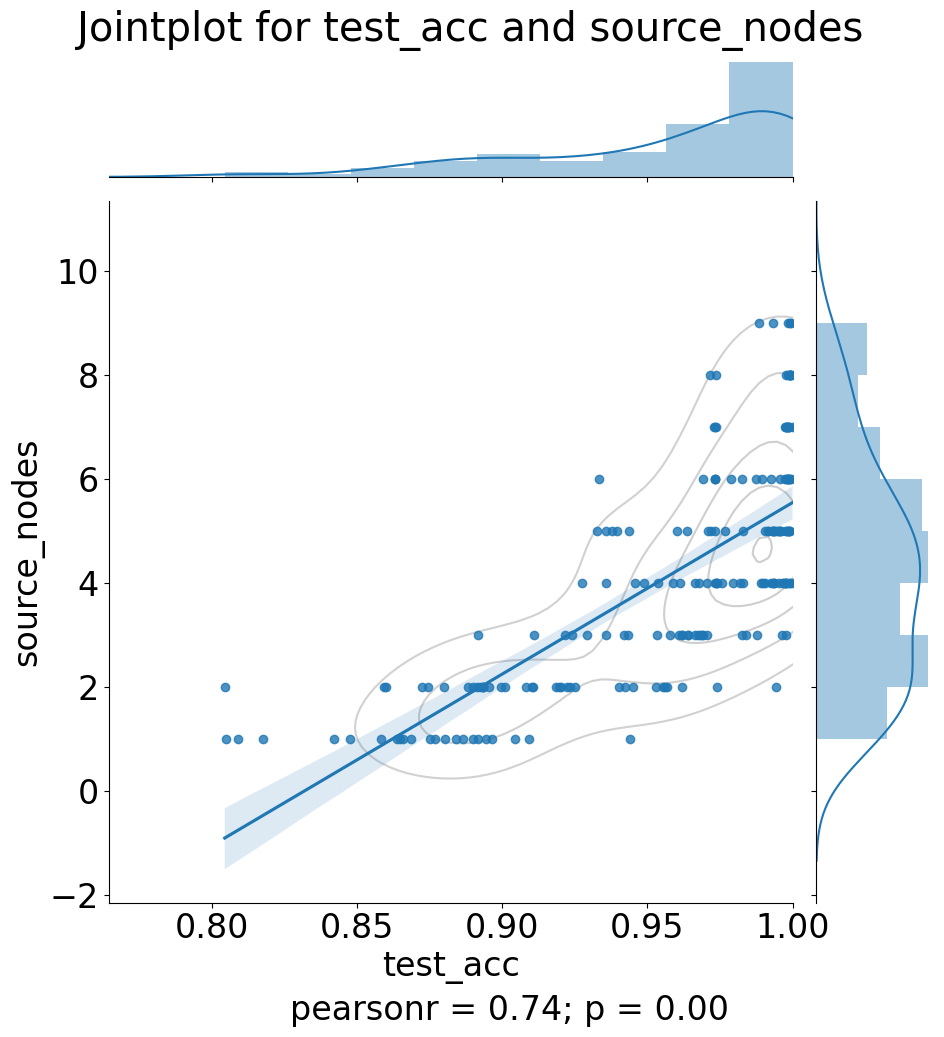}
        \caption{Correlation between test\_acc and the number of source nodes} \label{fig:jp_gru_source}
    \end{subfigure}
    \hfill
    \begin{subfigure}{0.45\textwidth}
        \includegraphics[width=\linewidth]{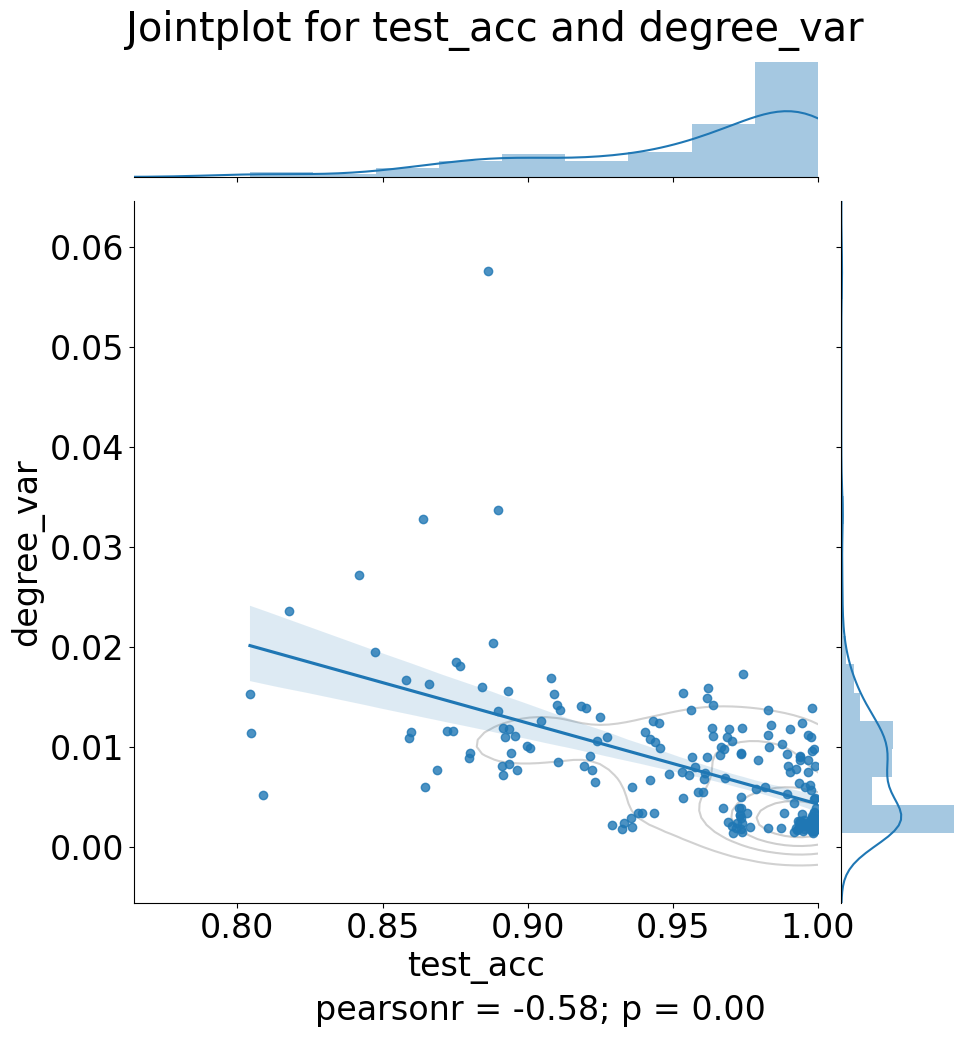}
        \caption{Correlation between test\_acc and degree of the base random graph} \label{fig:jp_gru_degree}
    \end{subfigure}

\caption[Correlation between test accuracy of GRU and its different graph and recurrent network properties - 1]{Correlation between test accuracy of GRU and its different graph and recurrent network properties - 1} \label{fig:gru_correlation_1}
\end{figure}

\begin{figure}[H]
    \centering
    \begin{subfigure}{0.45\textwidth}
        \includegraphics[width=\linewidth]{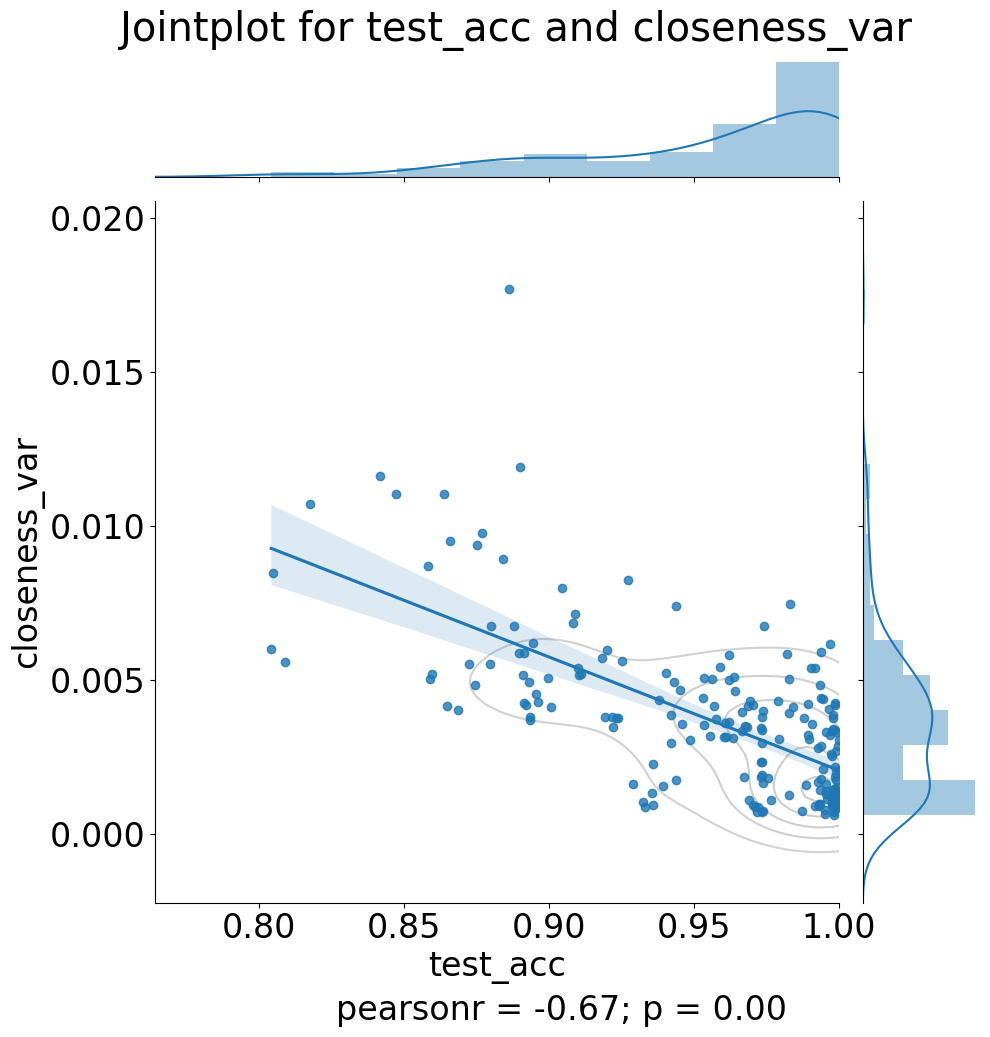}
        \caption{Correlation between test\_acc and closeness in base random graph} \label{fig:jp_gru_close}
    \end{subfigure}
    \hfill
    \begin{subfigure}{0.45\textwidth}
        \includegraphics[width=\linewidth]{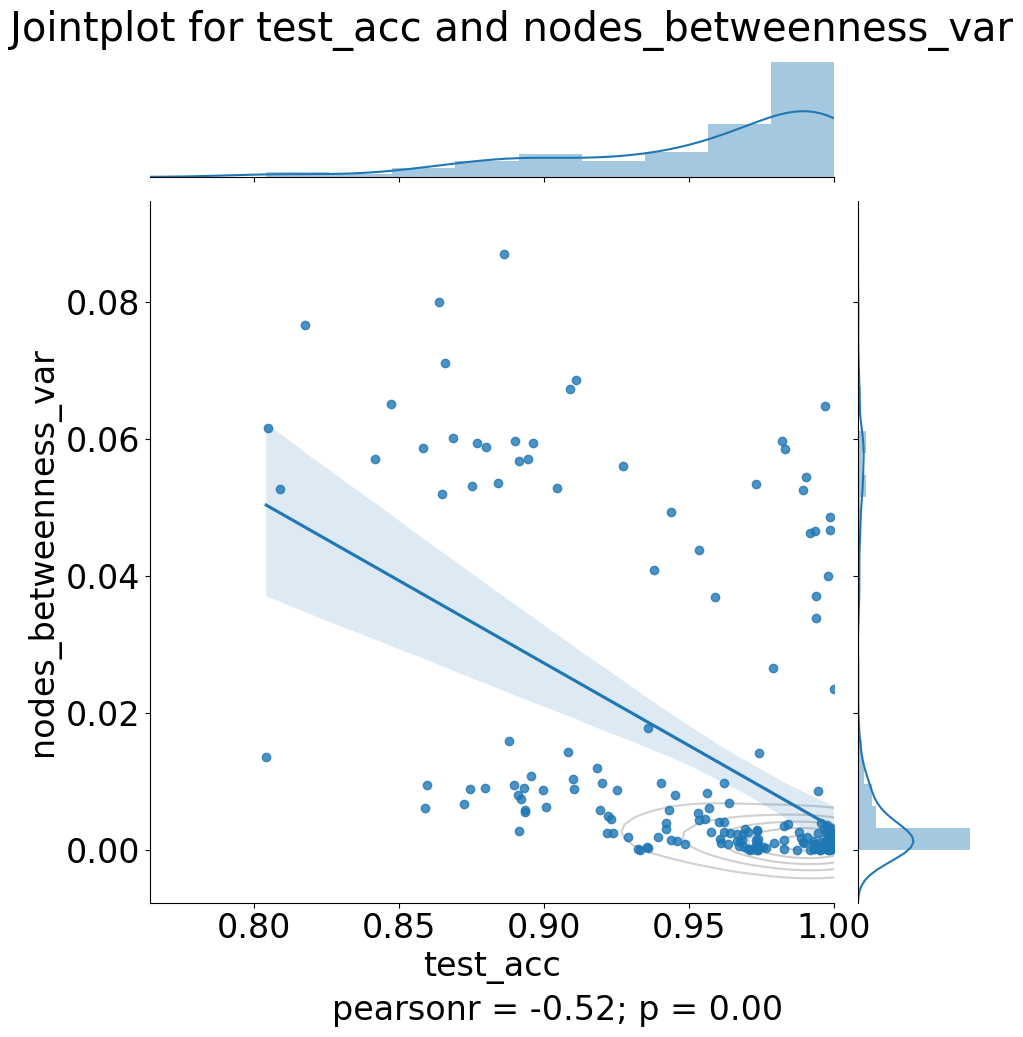}
        \caption{Correlation between test\_acc and node betweenness in base random graph} \label{fig:jp_gru_node_bn}
    \end{subfigure}

\caption[Correlation between test accuracy of GRU and its different graph and recurrent network properties - 2]{Correlation between test accuracy of GRU and its different graph and recurrent network properties - 2} \label{fig:gru_correlation_2}
\end{figure}

\end{appendices}

\backmatter

\printbibliography


\chapter{Eidesstattliche Erkl\"arung}

	Hiermit versichere ich, dass ich diese \thesisType{} selbstst\"andig und ohne Benutzung anderer als der angegebenen Quellen und Hilfsmittel angefertigt habe und alle Ausf\"uhrungen, die w\"ortlich oder sinngem\"a\ss{} übernommen wurden, als solche gekennzeichnet sind, sowie, dass ich die \thesisType ~in gleicher oder \"ahnlicher Form noch keiner anderen Pr\"ufungsbeh\"orde vorgelegt habe.

	\vspace{3cm}

	Passau, \thedate

	\vspace{2cm}

	\parbox{8cm}{
		\hrule \strut \theauthor
	}

\end{document}